\pdfoutput=1

\documentclass[11pt]{article}

\usepackage[final]{acl}

\usepackage{array}
\usepackage{xspace}

\newcolumntype{L}[1]{>{\raggedright\let\newline\\\arraybackslash\hspace{0pt}}m{#1}}
\newcolumntype{C}[1]{>{\centering\let\newline\\\arraybackslash\hspace{0pt}}m{#1}}
\newcolumntype{R}[1]{>{\raggedleft\let\newline\\\arraybackslash\hspace{0pt}}m{#1}}

\makeatletter
\def\adl@drawiv#1#2#3{%
        \hskip.5\tabcolsep
        \xleaders#3{#2.5\@tempdimb #1{1}#2.5\@tempdimb}%
                #2\z@ plus1fil minus1fil\relax
        \hskip.5\tabcolsep}
\newcommand{\cdashlinelr}[1]{%
  \noalign{\vskip\aboverulesep
           \global\let\@dashdrawstore\adl@draw
           \global\let\adl@draw\adl@drawiv}
  \cdashline{#1}
  \noalign{\global\let\adl@draw\@dashdrawstore
           \vskip\belowrulesep}}
\makeatother

\makeatletter
\DeclareRobustCommand\onedot{\futurelet\@let@token\@onedot}
\def\@onedot{\ifx\@let@token.\else.\null\fi\xspace}

\def\eg{e.g\onedot,\xspace} 
\def\ie{i.e\onedot,\xspace} 
\def\cf{c.f\onedot} \def\Cf{\emph{C.f}\onedot}
\def\etc{etc\onedot} 
\def\wrt{w.r.t\onedot} 
\makeatother

\newcommand{\modelspacydp}{\textsc{$\text{SpaCy\textsubscript{DP}}$}\xspace}
\newcommand{\modelspacydpm}{\textsc{$\text{SpaCy\textsubscript{M}}$}\xspace}
\newcommand{\modelpysbd}{\textsc{PySBD}\xspace}
\newcommand{\modelersatz}{\textsc{Ersatz}\xspace}
\newcommand{\modelpunkt}{\textsc{NLTK}\xspace}

\newcommand{\modelwtp}{\textsc{$\text{WtP}$}\xspace}
\newcommand{\modelwtpu}{\textsc{$\text{WtP}$}\xspace}
\newcommand{\modelwtpt}{\textsc{$\text{WtP\textsubscript{T}}$}\xspace}
\newcommand{\modelwtppunct}{\textsc{$\text{WtP\textsubscript{PUNCT}}$}\xspace}

\newcommand{\modelours}{\textsc{SaT}\xspace}

\newcommand{\modeloursft}{\textsc{$\text{SaT\textsubscript{+SM}}$}\xspace}
\newcommand{\modelourslora}{\textsc{$\text{SaT\textsubscript{+LoRA}}$}\xspace}

\newcommand{\modelxlmr}{\textsc{$\text{XLM-R}$}\xspace}

\newcommand{\modelcommandr}{\textsc{Command R}\xspace}
\newcommand{\modelllama}{\textsc{$\text{Llama 3\textsubscript{8B}}$}\xspace}

\newcommand{\modelsbd}{\textsc{mLSBD-T}\xspace}
\newcommand{\modelsbdmonospecific}{
    \textsc{mLSBD-T}
    \raisebox{-.25ex}{
        \begin{tabular}{@{}c@{}}
            \scriptsize Mono\\[-1ex] 
            \scriptsize Specific
        \end{tabular}
    }\xspace
}
\newcommand{\modelsbdmonoboth}{
    \textsc{mLSBD-T}
    \raisebox{-.25ex}{
        \begin{tabular}{@{}c@{}}
            \scriptsize Mono\\[-1ex] 
            \scriptsize Both
        \end{tabular}
    }\xspace
}
\newcommand{\modelsbdmultispecific}{
    \textsc{mLSBD-T}
    \raisebox{-.25ex}{
        \begin{tabular}{@{}c@{}}
            \scriptsize Multi\\[-1ex] 
            \scriptsize Specific
        \end{tabular}
    }\xspace
}
\newcommand{\modelsbdmultiboth}{
    \textsc{mLSBD-T}
    \raisebox{-.25ex}{
        \begin{tabular}{@{}c@{}}
            \scriptsize Multi\\[-1ex] 
            \scriptsize Both
        \end{tabular}
    }\xspace
}

\usepackage{todonotes}

\makeatletter
\newcommand*\iftodonotes{\if@todonotes@disabled\expandafter\@secondoftwo\else\expandafter\@firstoftwo\fi}
\makeatother

\newcommand{\subrparagraph}[1]{\vspace{1.2mm}\noindent\textbf{#1}}
\newcommand{\rparagraph}[1]{\vspace{1.6mm}\noindent\textbf{#1.}}
\newcommand{\sparagraph}[1]{\vspace{0.0mm}\noindent\textbf{#1.}}

\definecolor{bluepigment}{rgb}{0.2, 0.2, 0.6}
\definecolor{ballblue}{rgb}{0.13, 0.67, 0.8}
\definecolor{bleudefrance}{rgb}{0.19, 0.55, 0.91}

\definecolor{spotifygreen}{RGB}{30, 215, 96}
\definecolor{twitterblue}{RGB}{29, 161, 242}

\definecolor{softblue}{RGB}{143, 178, 200}
\definecolor{softblue}{RGB}{132, 197, 250}
\definecolor{t5blue}{RGB}{207, 226, 243}
\definecolor{vibrantorange}{RGB}{240, 177, 110}
\definecolor{t5red}{RGB}{244, 204, 204}
\definecolor{freshgreen}{RGB}{111, 207, 151}
\definecolor{t5green}{RGB}{217, 234, 211}
\definecolor{t5yellow}{RGB}{255, 242, 204}

\NewDocumentCommand\musicnote{}{\scalerel*{\includegraphics{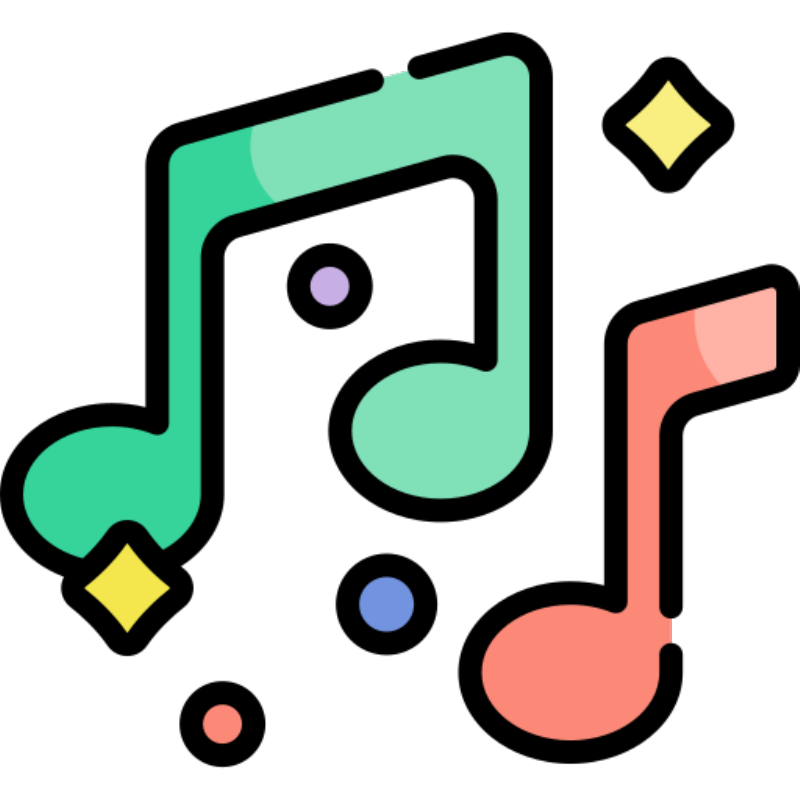}}{X}}

\NewDocumentCommand\globe{}{\scalerel*{\includegraphics{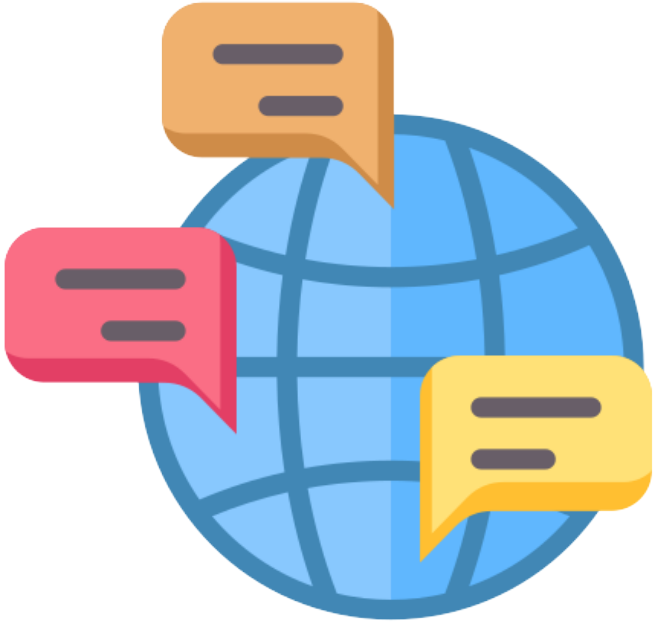}}{X}}
\NewDocumentCommand\mic{}{\scalerel*{\includegraphics{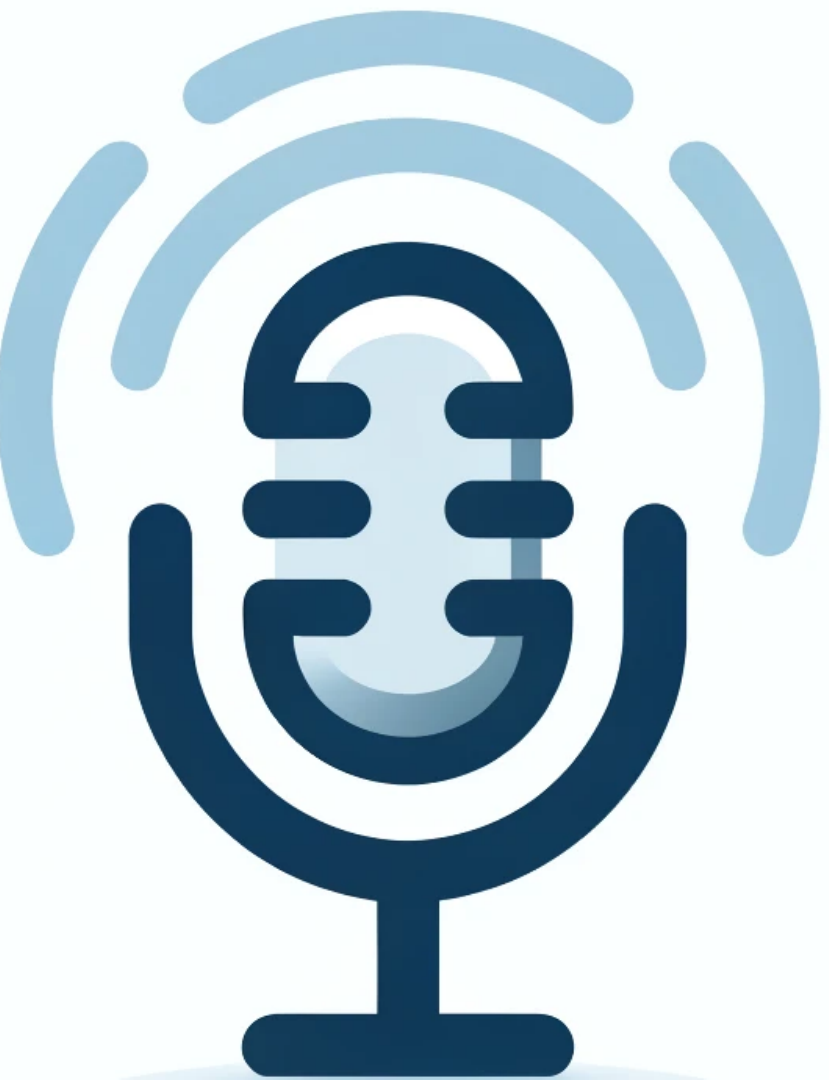}}{X}}

\newcommand{\thickmidrule}{\midrule[\heavyrulewidth]}

\usepackage{amsmath,amsfonts,bm}

\def\eqref#1{equation~\ref{#1}}

\def\1{\bm{1}}

\DeclareMathAlphabet{\mathsfit}{\encodingdefault}{\sfdefault}{m}{sl}
\SetMathAlphabet{\mathsfit}{bold}{\encodingdefault}{\sfdefault}{bx}{n}

\usepackage{times}
\usepackage{latexsym}

\usepackage[T1,T5]{fontenc}

\usepackage[utf8]{inputenc}
\usepackage{microtype}
\usepackage{inconsolata}
\usepackage{amssymb}
\usepackage{pifont}

\renewcommand{\encodingdefault}{T1}
\newcommand{\vietnamese}[1]{{\fontencoding{T5}\selectfont#1}}

\usepackage{xcolor}

\usepackage{graphicx}
\usepackage{tabularx}
\usepackage{amsmath}
\usepackage{booktabs}
\usepackage{multirow}
\usepackage{caption}
\usepackage{subcaption}
\usepackage{algorithm}
\usepackage{algpseudocode}
\usepackage{amsfonts}
\usepackage{verbatim}
\usepackage{tikzducks}
\usepackage{adjustbox}
\usepackage{arydshln}
\usepackage{rotating,booktabs}
\usepackage{makecell}
\usepackage{lipsum}

\usepackage{graphicx}
\usepackage{csquotes}
\usepackage{tablefootnote}
\usepackage[most]{tcolorbox}  

\usepackage{tikz}
\usetikzlibrary{shapes, arrows.meta, positioning}

\usepackage{scalerel,xparse}
\usepackage{fix-cm}

\title{Segment Any Text: A Universal Approach for Robust, Efficient and Adaptable Sentence Segmentation}

\author{
Markus Frohmann$^{1,2}$~~~~~~~~
Igor Sterner$^3$\\~~~~~~~~
\textbf{Ivan Vuli\'c}$^{* 3}$~~~~
\textbf{Benjamin Minixhofer}$^{* 3}$~~~~
\textbf{Markus Schedl}\thanks{Equal senior authorship.}$^{1,2}$ \\
$^1$Johannes Kepler University Linz~~~$^2$Linz Institute of Technology, AI Lab\\
$^3$University of Cambridge
\\
\small{\texttt{markus.\{frohmann, schedl\}@jku.at}}, 
\quad\small{\texttt{\{is473, bm644, iv250\}@cam.ac.uk}}
}

\begin{document}

\maketitle
\begin{abstract}

Segmenting text into sentences plays an early and crucial role in many NLP systems.
This is commonly achieved by using rule-based or statistical methods relying on lexical features such as punctuation.
Although some recent works no longer exclusively rely on punctuation, we find that no prior method achieves all of (i) robustness to missing punctuation, (ii) effective adaptability to new domains, and (iii) high efficiency. We introduce a new model --- Segment any Text (\modelours) --- to solve this problem.
To enhance robustness, we propose a new pretraining scheme that ensures less reliance on punctuation. To address adaptability, we introduce an extra stage of parameter-efficient fine-tuning, establishing state-of-the-art performance
in distinct domains such as verses from lyrics and legal documents.
Along the way, we introduce architectural modifications that result in a threefold gain in speed over the previous state of the art and solve spurious reliance on context far in the future.
Finally, we introduce a variant of our model with fine-tuning on a diverse, multilingual mixture of sentence-segmented data, acting as a drop-in replacement and enhancement for existing segmentation tools.
Overall, our contributions provide a universal approach for segmenting any text. Our method outperforms \emph{all} baselines --- including strong
LLMs --- across 8 corpora spanning diverse domains and languages, especially in practically relevant situations where text is poorly formatted.\footnote{Our models and code, including documentation, are available at \url{https://github.com/segment-any-text/wtpsplit} under the MIT license.}

\end{abstract}

\section{Introduction}
\label{sec:introduction}
\begin{figure}[t]
    \centering
    \includegraphics[width=1\linewidth]{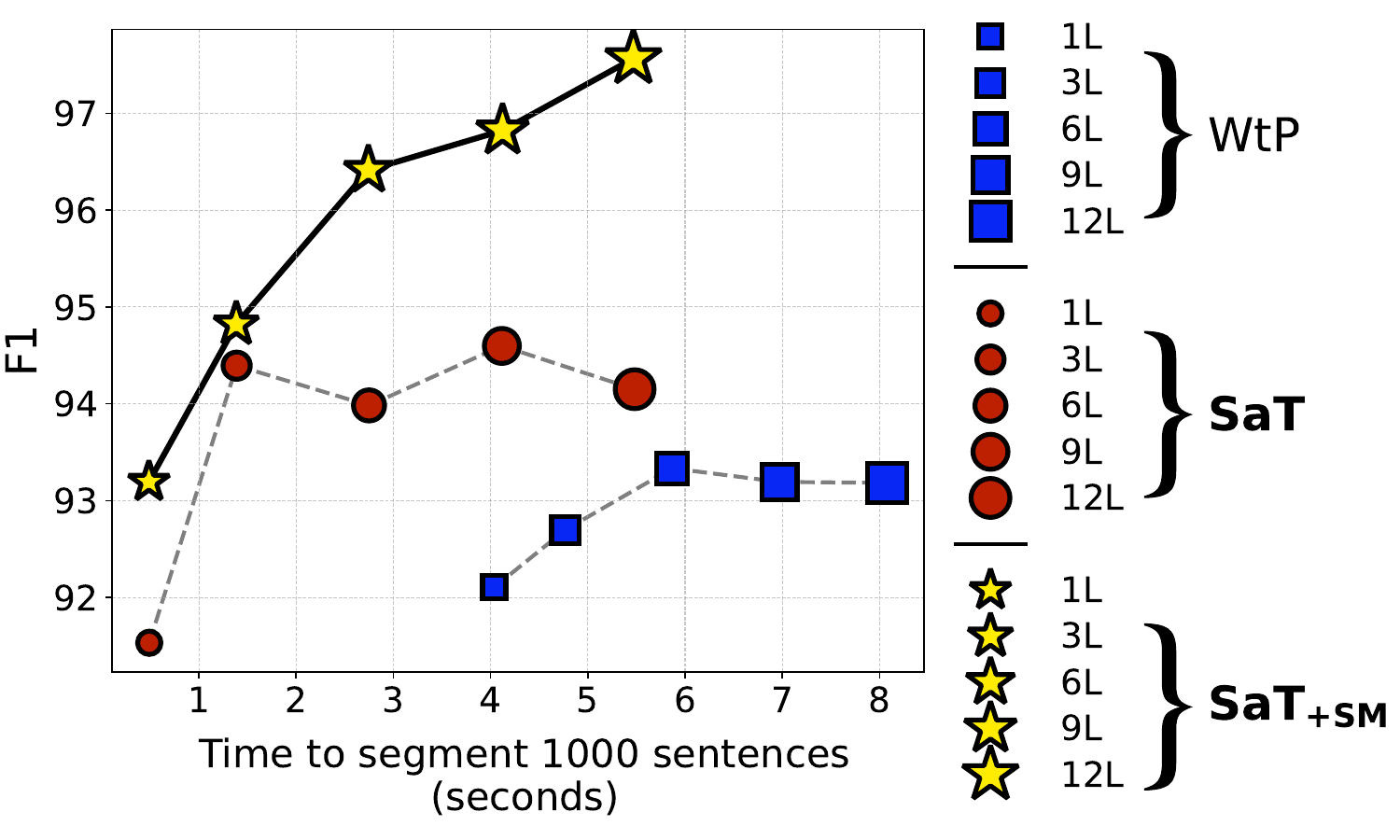}
    \caption{F1 scores and inference time for the prior SoTA (\modelwtpu) and our models (\modelours and \modeloursft), evaluated on the Ersatz sentence segmentation benchmark. We average over all 23 languages and show the average time (10 runs) for variants with different sizes (L = \#layers) to segment 1,000 sentences using consumer hardware (1 Nvidia GTX 2080 Ti GPU).
    }
    \label{fig:pareto-plot}
\end{figure}

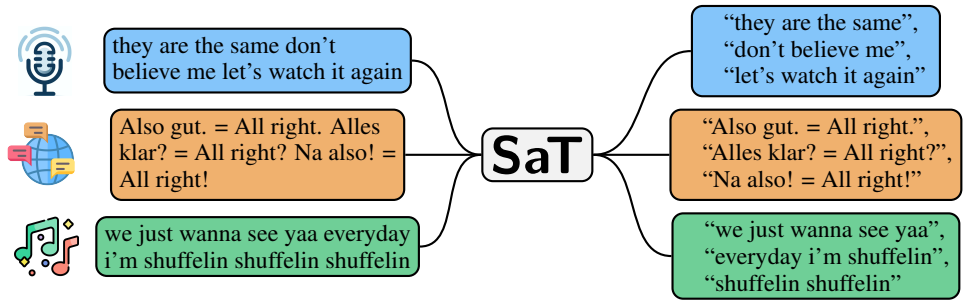
\begin{figure*}[ht]
    \vspace{-3mm}
    \centering
    \small
    \begin{tikzpicture}[
        scale=0.4, 
        node distance=1cm and 1cm, 
        auto,
        every node/.style={
            rounded corners, 
            thick, 
            draw=black, 
            align=left,
        }
    ]

    \node (model) [text centered, fill=gray!10, font=\fontsize{20pt}{18pt}\selectfont] {\fontfamily{lmss}\textbf{Sa\kern-0.1em T}};

    \node (in-1) [fill=vibrantorange, left= of model]{
        Also gut. = All right. Alles\\
        klar? = All right? Na also! =\\
        All right!
    };
    \node [left=3.5mm of in-1, yshift=0.01cm, font=\fontsize{35}{18pt}\selectfont, draw=none, rounded corners=0mm] (cs-icon) {\globe};

    \node (in-2) [fill=softblue, above=0.2cm of in-1]{
       they are the same don't \\
       believe me let's watch it again
    };
    \node [left=3mm of in-2, yshift=0cm, font=\fontsize{38pt}{18pt}\selectfont, draw=none, rounded corners=0mm] (asr-icon) {\mic};

    \node (in-3) [fill=freshgreen, below=0.2cm of in-1]{
        we just wanna see yaa everyday\\
        i'm shuffelin shuffelin shuffelin
    };
    \node [left=1mm of in-3, yshift=0cm, font=\fontsize{35pt}{18pt}\selectfont, draw=none, rounded corners=0mm] (music-icon) {\musicnote};

    \node (out-1) [fill=vibrantorange, right= of model]{
        \quad ``Also gut. = All right.'', \quad \\
        \quad ``Alles klar? = All right?'', \quad \\ 
        \quad ``Na also! = All right!'' \quad 
        \quad
    };

    \node (out-2) [fill=softblue, above=0.15cm of out-1]{
        \quad ``they are the same'', \quad \\
        \quad ``don't believe me'', \quad \\
        \quad ``let's watch it again'' \quad
    };

    \node (out-3) [fill=freshgreen, below=0.15cm of out-1]{
        \quad ``we just wanna see yaa'', \quad\\
        \quad ``everyday i'm shuffelin'', \quad\\
        \quad ``shuffelin shuffelin'' \quad
    };

    \draw [thick, -] (model) to[out=180, in=0, looseness=1] (in-1);
    \draw [thick, -] (model) to[out=180, in=0, looseness=1] (in-2);
    \draw [thick, -] (model) to[out=180, in=0, looseness=1] (in-3);

    \draw [thick, -] (model) to[out=0, in=180, looseness=1] (out-1);
    \draw [thick, -] (model) to[out=0, in=180, looseness=1] (out-2);
    \draw [thick, -] (model) to[out=0, in=180, looseness=1] (out-3);


    \end{tikzpicture}
    \vspace{-1mm}
    \caption{Examples of our model's predictions from \textcolor{softblue}{(i) ASR output}, \textcolor{vibrantorange}{(ii) multilingual text}, and \textcolor{freshgreen}{(iii) verse segmentation}.
    \textcolor{softblue}{(i)} shows part of a transcribed TED talk, demonstrating our method is agnostic to punctuation and casing.
    \textcolor{vibrantorange}{(ii)} is from a Reddit post of German-English translations; existing rule-based systems would segment at nearly every punctuation,
    and existing neural systems are too reliant on punctuation or need a language code.
    \textcolor{freshgreen}{(iii)} shows segmentation of lyrics into verses, showing our model's predictions in a distinct domain.
    }
    \label{fig:system-fig}
\end{figure*}

Sentence segmentation is defined as the task of identifying boundaries between sentences in a given text. 
High-quality sentence boundaries are crucial in many NLP tasks and systems since models often expect individual sentences as input~\citep[\textit{inter alia}]{reimers-gurevych-2019-sentence, reimers-gurevych-2020-making, liu-etal-2021-fast, tiedemann-thottingal-2020-opus}. Further, errors in segmentation can have detrimental effects on downstream task performance, \eg in machine translation~\citep{minixhofer-etal-2023-wheres,wicks-post-2022-sentence, avelka2017SentenceBD}.

Existing sentence segmentation tools predominantly rely on punctuation marks. This limitation renders them 
impractical for text lacking punctuation.
To address this issue, some recent methods aim to overcome this dependency~\citep{honnibal2020spacy, minixhofer-etal-2023-wheres}.
Specifically, during the training of their model, \modelwtp \citep{minixhofer-etal-2023-wheres} randomly removes punctuation characters to increase robustness against missing punctuation.

However, the performance of \modelwtp as the current state-of-the-art (SoTA) model and all other segmenters is still poor on texts from more challenging domains. This includes, among others, user-generated text such as tweets and highly heterogeneous domains such as lyrics.
Segmenting these texts is challenging because of 
missing and/or extra punctuation, inconsistent spacing, and especially irregular casing.
Furthermore, nearly all existing systems, including \modelwtp, require the specification of the texts’ language at inference time. 
This necessitates an additional preprocessing step of language identification, which often proves to be imperfect, particularly with user-generated content \citep{lui-baldwin-2014-accurate, sterner-teufel-2023-tongueswitcher}.
Moreover, this necessity limits their applicability to code-switching text.

To address these challenges, we present a sentence segmentation method that does not rely on language codes or punctuation marks, making it \emph{universally applicable} across a broad range of languages, corpora, and domains.
Specifically, we train subword-based multilingual encoder language models (LMs) in a self-supervised way to predict naturally occurring newlines on web-scale text. We then continue training models on sentence-segmented data in a second, supervised stage to further improve sentence segmentation performance.

We deal with several major issues with previous tools:
To ensure \textit{robustness} against missing punctuation and
noise, we propose a set of corruptions, applied randomly to the input during training.
Crucially, our method does not rely on language codes.
In addition, we mitigate issues observed with \textit{short sequences} via a 
novel
limited lookahead mechanism.
Furthermore, we recognize the \textit{variability of sentence boundaries} across domains and sentence definitions.
To address this, we show how our models can be efficiently adapted to target domains via LoRA~\citep{hu2022lora}, outperforming previous adaptation methods, especially in data-constrained settings.
Further, we improve efficiency by shedding the upper layers of the base model for our default 3-layer models, which segments 1000 sentences in approx. $0.5$ seconds on our hardware.

Figure~\ref{fig:pareto-plot} shows that the standard 3-layer version of \modelours outperforms the current open weights state-of-the-art, \modelwtp, while achieving a $\approx3$x reduction in inference time.
Overall, we present several innovations that overcome each of the shortcomings of previous methods, culminating in a \emph{universal model for sentence segmentation}.
We provide some examples of our model's predictions in Figure~\ref{fig:system-fig}.

\paragraph{Contributions.}
\textbf{1)} We introduce \emph{Segment any Text} (\modelours), an efficient method for sentence segmentation that can reliably segment text across 85 languages regardless of lexical features such as punctuation or casing.
\textbf{2)} We show how our models can be adapted to different domains via data-efficient means, requiring only a minimal set (\eg 16) of sentence-segmented examples.
\textbf{3)} We train and release \modelours models in five sizes, covering 85 languages, and demonstrate state-of-the-art performance across 8 corpora, even outperforming newly introduced strong (open weights) LLM baselines.

\section{Background and Related Work}
\label{sec:related}
We start by providing an overview of existing sentence segmentation systems. Following \citet{read-etal-2012-sentence}, we categorize them into \textbf{1)} rule-based, \textbf{2)} supervised statistical, and \textbf{3)} unsupervised statistical approaches.
Then, we discuss the recently introduced state-of-the-art approach, \modelwtp. 
Moreover, we discuss domain-specific segmentation approaches.
Lastly, since we are the first to evaluate large language models (LLMs) for sentence segmentation broadly, we briefly survey them and discuss their usage in sentence segmentation tasks.

\subsection{General Systems and Baselines}

    \subrparagraph{1. Rule-based methods} segment text into sentences using hand-crafted rules. The segmenters in Moses~\citep{koehn-etal-2007-moses} and SpaCy~\citep{honnibal2020spacy} split on punctuation characters, except for predefined exceptions like abbreviations and acronyms. PySBD~\citep{sadvilkar-neumann-2020-pysbd} relies on exceptions and regular expression rules. Although generally efficient, these methods demand manual per-language effort to incorporate language-specific rules. This also
    necessitates specifying a language code at inference time.
    
    \subrparagraph{2. Supervised statistical methods} learn segmentation from a sentence-segmentation annotated corpus. One early method by \citet{riley-1989-applications} involved a decision tree to determine if each punctuation mark in a text represents a sentence boundary based on linguistic features surrounding punctuation. Satz~\citep{palmer-hearst-1997-adaptive} and Splitta~\citep{gillick-2009-sentence} build on this approach but utilize neural networks and SVMs, respectively. Similarly, in Ersatz, \citet{wicks-post-2021-unified} propose to use a Transformer~\citep{NIPS2017_3f5ee243} with subwords as context around punctuation marks.  However, these methods are limited by their reliance on punctuation to define sentence boundaries. This becomes problematic in poorly punctuated texts as non-punctuation characters cannot serve as sentence boundaries.
    Breaking from this limitation, the dependency parser in the SpaCy library~\citep{honnibal2020spacy} jointly learns dependency parsing and sentence segmentation on a labeled corpus without special treatment of punctuation.
    
    \subrparagraph{3. Unsupervised statistical methods} predict
    sentence boundaries from unsegmented text alone. 
    \citet[\modelpunkt;][]{kiss-strunk-2006-unsupervised} use features such as character length and internal punctuation to identify abbreviations, initials, and ordinal numbers, treating all other punctuation as sentence boundaries. 
    Furthermore, \citet{wicks-post-2021-unified} additionally introduces an unsupervised version of Ersatz, relying on punctuation preceding paragraph breaks.

\subsection{Where's the Point (WtP)}\label{sec:wtp}
    WtP represents the current state-of-the-art in sentence segmentation~\citep{minixhofer-etal-2023-wheres}. 
    Like our method, it can be used in unsupervised and supervised variations.
    Hence, we choose \modelwtp as our main baseline and examine it in the following.
    
    \modelwtp is trained to predict the \textit{newline probability} (\ie the probability for any character to be followed by a \texttt{\textbackslash n} symbol) on web-scale text data in 85 languages. 
    Training is self-supervised since newline symbols occur naturally, typically corresponding to \textit{paragraphs}, each containing multiple sentences.
    \modelwtp thus takes characters as input and generates a probability for each character to be paragraph-ending.
    A character is treated as a boundary if the probability is greater than a selected threshold $\alpha$. 
    To apply models trained in this way to segment text into \textit{sentences}, \citet{minixhofer-etal-2023-wheres}  find it is sufficient to lower the threshold $\alpha$.
    
    \subrparagraph{Robustness to corruptions.}
        To make \modelwtp less reliant on punctuation, \citet{minixhofer-etal-2023-wheres} randomly remove some punctuation during training.
        In addition, they predict the likelihood of commonly occurring punctuation as an auxiliary objective. For details, we refer to Appendix~\ref{sec:aux-obj}.
        While this helps make \modelwtp models less reliant on punctuation, we still find that \modelwtp models have major issues when text is inconsistently formatted, especially irregular casing.

    \subrparagraph{Efficiency.}
        \modelwtp uses the character-level encoder LM Canine-S~\citep{clark-etal-2022-canine} as its backbone.
        Operating on \emph{characters} as the fundamental unit constitutes a major bottleneck in terms of speed, resulting in poor efficiency.
        
    \subrparagraph{Multilinguality.}
        To increase language-specific capacity, \modelwtp utilizes language adapters~\citep{pfeiffer-etal-2022-lifting}.
        This, however, confines its multilingual abilities since a \emph{language code} must be specified at inference time.
        This is especially problematic in code-switching, where multiple languages are present, leading to ambiguity.

    \subrparagraph{Short texts.}
        We also found \modelwtp models deficient in segmenting short sequences, such as tweets or sentence pairs.
        During training, paragraphs are packed to always fully use the model’s context size.
        While being efficient at training, we hypothesize that this renders short sequences out-of-domain.

    \subrparagraph{Domain adaptation.}
        \citet{minixhofer-etal-2023-wheres} also evaluate two supervised adaptation methods.
        First, \modelwtpt tunes the segmentation threshold $\alpha$ based on an already sentence-segmented corpus. Second, based on the auxiliary punctuation prediction objective, \modelwtppunct fits a logistic regression on the probability distribution of the punctuation logits. However, these kinds of adaptations fall short on more challenging domains such as lyrics and code-switched text, as quantified later in Section~\ref{ss:challenging_domains}.

\subsection{Domain-specific Sentence Segmentation}
    Due to deviations from typical sentence structures, differences in sentence lengths, and non-standard punctuation, sentence segmentation is highly dependent on domain-specific characteristics~\citep{Sheik2024LegalSB}, providing a strong basis for domain-specific systems~\citep{read-etal-2012-sentence}.

    Prior studies have focused on creating a dedicated model for a single domain. \citet{Reynar1997AME} 
    utilized features unique to the financial domain.
    \citet{Tuggener2021TheSE} hosted a shared task on transcripts of spoken texts.
    \citet{Brugger2023MultiLegalSBDAM} train models to segment sentences in the legal domain.
    
    Previous approaches to segmenting lyrics into verses require songs to be already pre-segmented into lines. \citet{watanabe-etal-2016-modeling} extract features based on repeated patterns and part-of-speech. \citet{fell-etal-2018-lyrics} improve upon this approach by using convolutions and a more refined set of features.
    
    In contrast to prior domain-specific models, we propose a single model that can be efficiently adapted for segmenting sentences from wildly heterogeneous domains and languages, outperforming previous domain-specific models, even when using only a limited number of examples.

\subsection{Large Language Models}
    Large language models (LLMs) have become a de facto tool for use in many NLP tasks~\citep{Zhao2023ASO, Minaee2024LargeLM}.
    Most modern LLMs are decoder-only Transformers~\citep[\textit{inter alia}]{NIPS2017_3f5ee243, Brown2020LanguageMA, Touvron2023Llama2O, Jiang2024MixtralOE}. Recently, prompting has emerged as the dominating paradigm for solving a task~\citep{Ouyang2022TrainingLM, 10.1145/3560815}.
    
    However, despite widespread use, LLMs have yet to be extensively evaluated for sentence segmentation.
    In this work, we aim to bridge this gap by shedding light on how well popular LLMs can segment sentences when prompted to do so, particularly in more challenging domains such as lyrics, where using LLMs may be especially valuable.

\section{\modelours: Segment any Text}
\label{sec:method}
To create a reliable and effective system across various scenarios, we pre-train a model on paragraph segmentation as in \citet{minixhofer-etal-2023-wheres}. In the following, we outline how we solve each of the major issues of \modelwtp discussed earlier, leading to a \emph{universal model for sentence segmentation}.

\subrparagraph{Efficiency.} We resort to models using subword tokenization, processing tokens consisting of \emph{multiple characters} at a time, making them considerably faster than their character-level counterparts. 

\subrparagraph{Multilinguality.} Unlike \citet{minixhofer-etal-2023-wheres}, we do not rely on language adapters. In addition to improving inference time and storage requirements, this also improves multilinguality since no language has to be specified at inference time. 

\subrparagraph{Robustness to corruptions.} 
We randomly remove common punctuation-only tokens with probability $p$ and use the auxiliary punctuation-prediction objective during training. For details, see §~\ref{sec:wtp} and Appendix~\ref{sec:aux-obj}. Further, we randomly remove \emph{all} casing and punctuation in 10\% of samples within a batch during training.
The resulting model, \textit{Segment any Text} (\modelours), already shows strong segmentation performance at improved efficiency.

Still, to further improve \modelours, we continue training it on a \textbf{S}upervised \textbf{M}ixture of already-segmented sentences.
To be even less dependent on patterns such as punctuation, spaces, and casing, we augment the data by introducing several additional corruption schemes, resulting in our more specialized model, \modeloursft.

Our first corruption scheme removes all casing, if available, and punctuation tokens for all text.
Secondly, we add randomness to the corruption in as many situations as we find useful, aiming to emulate user-generated text in tweets or forums. This includes duplicating punctuation, removing punctuation, lowercasing, and removing/adding spaces between sentences.
Finally, we also use clean, non-corrupted text. 
We then sample uniformly across these three categories. 
For details, see Appendix~\ref{sec:appendix-experiment-details}.

\begin{table*}[t]
   \centering
   \small
   \scalebox{0.76}{ 
        \begin{tabular}{lp{3cm}p{5.5cm}p{5cm}p{3.5cm}}
        \toprule
        \textbf{Domain} & \textbf{Dataset} & \textbf{Description} & \textbf{Characteristics} & \textbf{Source} \\
        \midrule
        \multirow{6}{*}{Clean Text}
        & Universal Dependencies (UD) & Treebanks in many languages. & Includes gold-standard segmentation into sentences. & \citet{de-marneffe-etal-2021-universal, nivre-etal-2020-universal} \\
        \cmidrule{2-5}
        & \multirow{2}{*}{OPUS100} & Sentences from subtitles and news in 100 languages. & A challenging sentence segmentation benchmark~\citep{zhang-etal-2020-improving} & \multirow{2}{*}{\citet{tiedemann-2012-parallel}} \\
        \cmidrule{2-5}
        & \multirow{2}{*}{Ersatz} & Sentences from WMT Shared Tasks, mainly comprising news (commentary). & Includes manual sentence segmentation corrections by \citet{wicks-post-2021-unified}. & \citet{wicks-post-2021-unified, barrault-etal-2020-findings} \\
        \midrule    
        \multirow{4}{*}{Noisy Text}
        & SEPP-NLG Shared Task (surprise test set) & 500 transcribed public TED talks in each of 4 European languages. & Neither casing nor punctuation tokens are present.
        & \citet{Tuggener2021TheSE} \\
        \cmidrule{2-5}
        & \multirow{2}{*}{Tweets} & User-generated content in the form of Slovene (\texttt{sl}) and Serbian (\texttt{sr}) tweets. & 
        Noisy; short in length (70/115 characters on average for \texttt{sl} and \texttt{sr}, respectively). & \citet{fivser2020janes, Miličević_Ljubešić_2016} \\
        \midrule         
        \multirow{4}{*}{\thead[l]{Code-\\switching}}
        & \multirow{4}{*}{\Cf~Table~\ref{tab:noisy-short-stats}.} & Reddit posts for German-English (\texttt{de}-\texttt{en}); data for 3 additional language pairs taken by concatenating code-switching sentences from bilingual transcriptions.
        & We treat all data as transcriptions, removing all punctuation and casing; we only keep sentences with at least one token of each language. 
        & 
        \citet{deuchar2009miami}, \citet{osmelak-wintner-2023-denglisch}, \citet{nguyen-bryant-2020-canvec} and \citet{cetinoglu-2017-code} 
        \\
        \cmidrule{2-5}
        \multirow{2}{*}{Legal}
        & \multirow{2}{*}{MultiLegalSBD} & Laws and judgements from legal documents in 6 languages.
        & Domain-specific jargon and structure; formal and complex sentences. & \multirow{2}{*}{\citet{Brugger2023MultiLegalSBDAM}} \\
        \cmidrule{2-5}
        \multirow{2}{*}{Lyrics}
        & \multirow{2}{*}{Verses} & 35,389 English songs across 16 genres spanning 3 levels of repetitiveness. &  We replicated the setup by \citet{fell-etal-2018-lyrics}. & \citet{meseguerbrocal-wasabi} \\
        \bottomrule
        \end{tabular}
    }
    \caption{Overview of the evaluation corpora we use. For more details, see Appendix~\ref{sec:appendix-experiment-details}.}
    \label{tab:dataset-summary}
\end{table*}

\subrparagraph{Short texts.} To resolve issues with short sequences,
we enforce \modelours to use only the immediate $N$ future tokens for its predictions. We do so via a \emph{limited lookahead} mechanism. Let $k_i$ be the token occurring at position $i$, and $\mathbf{a_{ij}}$ its corresponding attention mask, where $j$ corresponds to the token to be attended to. 
A naive modification of the attention mask would set $\mathbf{a_{ij}}=0$ for $j > i + N$.
However, using Transformer networks with multiple layers results in a lookahead of $N \times L$, where $L$ is the number of Transformer layers~\citep{Jiang2023Mistral7}.
We thus split up the lookahead evenly into $L$ layers, resulting in the following
attention mask:
\begin{equation*}
    \mathbf{a}_{\mathbf{i j}}=0 \text { for } j>i+N_{L},
\end{equation*}
where $N_L$ is the per-layer lookahead, \ie $N_L=\frac{N}{L}$.
Using an intermediate value for $N$ makes \modelours robust to both short and long sequences -- relying on some future context where appropriate, but not so much that it falters on short sequences.

Limited lookahead can be thought of as sliding window attention~\citep{Beltagy2020LongformerTL, Jiang2023Mistral7} with two crucial tweaks: 1) the sliding window extends forward into the future instead of backward, 2) past tokens are not masked out.

\subrparagraph{Domain adaptation.} Finally, some domains
require more sophisticated adaptation than only changing the threshold or relying on punctuation logits. We thus explore low-rank adaptation~\citep[LoRA;][]{hu2022lora} to adapt our models efficiently, denoted by \modelourslora.
We show how this enables state-of-the-art performance on verse segmentation using our models later in §~\ref{sec:results}.
In our setup, it trains $\approx1\%$ of the parameters of \modelours but results in \emph{no inference overhead} since LoRA weights can be merged into the backbone LM weights at inference time~\citep{pfeiffer2023modular}.

\section{Experimental Setup}
\label{sec:setup}
\subsection{Evaluation}

To evaluate our method, we compare ground truth and predicted sentence boundaries on the test sets of corpora spanning a diverse set of languages, sources, and domains,\footnote{We acknowledge the concept of \emph{domains} remains an open issue in NLP~\citep{holtermann-etal-2024-weight, Raffel2019ExploringTL}.} summarized in Table~\ref{tab:dataset-summary}.

In addition, to evaluate how well our method can segment short sequences, we generate non-overlapping sentence pairs from the datasets categorized as clean text.
We additionally simulate a real-time automatic speech recognition (ASR) scenario using transcripts from speeches in 76 languages. We generate sentence pairs in a similar way and remove all punctuation as well as all casing.

We report character-level F1 scores for the positive (\ie sentence-ending) labels. 
For short sequences, we use the proportion of perfectly segmented sequences within a corpus; this is stricter than F1 since any segmentation error results in a score of zero for the entire sequence.
For SEPP-NLG, we use the evaluation script and surprise test set provided by the shared task organizers~\citep{Tuggener2021TheSE},
reporting F1 scores on the \emph{token} level.
In our evaluations on clean text across all 85 languages, we run all competitor and baseline systems ourselves.
For these results, we test all differences for significance with paired two-tailed permutation tests.
We approximate them with $N\!=\!10,000$ and set the significance threshold at $\alpha\!=\!0.05$.
Additional evaluation and dataset details are provided in Appendix~\ref{sec:appendix-experiment-details}.

\begin{table*}[t]
   \centering
   \small
   \def\arraystretch{0.88}
   \scalebox{0.9}{
        \begin{tabular}{lccccccccccccccc}
        \toprule
        \textbf{Model} & \texttt{ar} & \texttt{cs} & \texttt{de} & \texttt{en} & \texttt{es} & \texttt{fi} & \texttt{hi} & \texttt{ja} & \texttt{ka} & \texttt{lv} & \texttt{pl} & \texttt{th} & \texttt{xh} & \texttt{zh} & \thead{\textbf{81}\\\textbf{langs}} \\

        \midrule
        \multicolumn{16}{c}{\textsc{Multilingual}} \\
        \midrule
        \modelspacydpm & - & 91.1 & 84.7 & 91.5 & 94.5 & 93.5 & - & - & - & 91.4 & 94.0 & - & - & - & - \\
        \modelersatz & 77.2 & 90.9 & 87.0 & 91.4 & 95.1 & 93.9 & 84.8 & 69.3 & - & 91.1 & 94.8 & - & - & 77.2 & - \\
        \midrule
        \modelllama & 78.2 & 93.4 & 92.6 & 95.2 & 96.0 & 95.5 & 85.6 & 64.7 & 89.2 & 93.0 & 96.2 & 66.0 & 71.7 & 82.0 & 79.1 \\ 
        \modelcommandr & 58.6 & 68.1 & 79.1 & 84.6 & 81.0 & 74.1 & 72.0 & 52.2 & 25.6 & 74.2 & 78.6 & 10.6 & 56.1 & 73.7 & 55.6 \\
        \midrule
        \modelours & 79.9 & 91.7 & 90.4 & 93.6 & 94.0 & 94.2 & 84.9 & \underline{\textbf{88.6}} & 75.7 & 92.2 & 93.7 & 68.0 & 80.3 & 78.0 & 84.9 \\
        \modeloursft & \underline{\textbf{80.7}} & \underline{\textbf{95.7}} & \underline{\textbf{94.0}} & \underline{\textbf{96.5}} & \underline{\textbf{97.3}} & \underline{\textbf{96.9}} & \underline{\textbf{90.3}} & 88.1 & \underline{\textbf{93.6}} & \underline{\textbf{96.1}} & \underline{\textbf{97.7}} & \underline{\textbf{72.9}} & \underline{\textbf{89.6}} & \underline{\textbf{88.9}} & \underline{\textbf{91.6}} \\
        \midrule
        \multicolumn{16}{c}{\textsc{Monolingual}} \\
        \midrule 
        \modelpunkt  & - & 90.8 & 87.1 & 92.2 & 94.1 & 93.9 & - & - & - & - & 94.5 & - & - & - & - \\
        \modelpysbd & 37.4 & - & 80.6 & 69.6 & 56.9 & - & 70.1 & 76.1 & - & - & 49.3 & - & - & 86.9 & - \\
        \modelspacydp & - & - & 89.0 & 92.9 & 93.5 & 94.1 & - & 77.1 & - & - & 95.3 & - & - & 87.7 & - \\
        \modelwtpu & 77.3 & 91.1 & 89.2 & 93.9 & 93.2 & 93.4 & 85.0 & 72.7 & 91.3 & 90.4 & 93.6 & 66.6 & 77.2 & 90.7 & 84.2 \\
        \midrule
        \modelwtpt & 79.9 & 92.0 & 92.0 & 93.5 & 94.2 & 94.1 & 85.2 & 85.6 & 91.1 & 93.1 & 93.5 & 69.7 & 80.7 & 89.3 & 85.9 \\
        \modelwtppunct & 85.4 & \underline{\textbf{96.4}} & \underline{95.0} & \underline{\textbf{96.7}} & 97.4 & \underline{\textbf{97.7}} & 90.8 & 93.1 & 92.8 & 96.6 & \underline{97.5} & 71.3 & 89.8 & \underline{\textbf{95.5}} & 91.7 \\
        \midrule
        \modelourslora & \underline{\textbf{86.3}} & \underline{96.2} & \underline{\textbf{95.4}} & \underline{\textbf{96.7}} & \underline{\textbf{97.7}} & \underline{97.5} & \underline{\textbf{92.9}} & \underline{\textbf{94.4}} & \underline{\textbf{93.3}} & \underline{\textbf{97.0}} & \underline{\textbf{97.7}} & \underline{\textbf{73.7}} & \underline{\textbf{90.8}} & 94.9 & \underline{\textbf{93.1}} \\
        \bottomrule
        \end{tabular}
    } 
    \caption{
    Mean sentence segmentation F1 scores over OPUS100, UD and Ersatz. For the average, we report macro F1 over languages from all datasets where train and test sets are available. Results are shown using 3-layer variations of all models.
    Numerically best results are in \textbf{bold}, statistically indistinguishable ones from this best are \underline{underlined}.
   }
    \label{tab:main-results}
\end{table*}  
\begin{table}[ht]
\centering
\small
\scalebox{0.9}{
\begin{tabular}{lccccc}
\toprule
\textbf{Model} & \texttt{en} & \texttt{de} & \texttt{fr} & \texttt{it} & \multicolumn{1}{c}{\textbf{Avg.}} \\
\midrule
htw+t2k & 77 & 82 & 76 & 75 & 78 \\
OnPoint & 80 & 82 & 77 & \textbf{77} & 79 \\
Unbabel & \textbf{83} & 78 & \textbf{78} & 76 & 79 \\
\midrule
\modelours & 73.4 & 79.9 & 73.1 & 72.9 & 74.8\\
\modeloursft & 79.7 & \textbf{84.0} & \textbf{78.3} & \textbf{77.1} & \textbf{79.8} \\
\bottomrule
\end{tabular}
}
\caption{F1 scores on the surprise test set of the SEPP-NLG Shared Task. For comparison, we provide results for the 3 best-performing systems from the Shared Task. We use 12-layer versions of our models.
}
\label{tab:ted2020-shared-task}
\end{table}


\rparagraph{Baselines} We compare against \modelpysbd and \modelpunkt as representatives of rule-based and unsupervised statistical methods. For supervised methods, we evaluate the punctuation-agnostic \modelspacydp 
and Spacy's multi-language model, \modelspacydpm.
We also compare against \modelersatz.
Our main comparison is against the current SoTA models: \modelwtpu, \modelwtpt, and \modelwtppunct.

\rparagraph{LLM-based baselines} To evaluate LLMs, we use \textbf{1)} Cohere's \modelcommandr as a recent LLM with claimed strong multilingual performance, and \textbf{2)} Meta's \modelllama due to its popularity and strong performance.
Officially, \modelcommandr supports 23 languages, whereas \modelllama only supports English.\footnote{Due to imperfect filtering of common web-crawled corpora, all LLMs can be considered multilingual to some extent.} 
We split up each dataset into chunks of $10$ sentences to avoid cases where sentences are cut off at critical positions and observed issues with long context lengths. Then, we prepend the prompt to each chunk and let the LLM segment $10$ sentences.\footnote{For a fair comparison, we thus exclude every 10th label when calculating F1 scores. For songs and short sequences, we feed in whole samples and hence do not exclude any labels.}
Finally, to make evaluation metrics robust to unwanted alterations of the input by the LLM, we apply the Needleman-Wunsch algorithm~\citep[NW;][]{Needleman1970AGM} to align sentences within each input and output chunk.
For the prompt and other implementation details, including alignment via NW, we refer to Appendix~\ref{sec:appendix-experiment-details}.

\subsection{Training Setup} 
We train Transformer models operating on subwords, initialized with the weights of XLM-RoBERTa~\citep[XLM-R;][]{conneau-etal-2020-unsupervised}.
We use a lookahead limit of $48$ tokens, which we found to work well in practice on text of any length, leading to \textbf{\modelours}.
We use the mC4~\citep{Raffel2019ExploringTL} corpus and sample text uniformly from the 85 languages also used by \citet{minixhofer-etal-2023-wheres}.

To train \textbf{\modeloursft}, we continue training \modelours on the training set of UD due to its high quality and availability in most of the 85 considered languages. For languages without UD data, we resort to silver-quality
data from OPUS100 or NLLB~\citep{nllb}, whichever is available.

To adapt to different user requirements \wrt \emph{efficiency}, we train and release \modelours and \modeloursft models in different sizes from 1-12 layers, where we remove the upper layers for models $<12$ layers.

For adaptation via LoRA (\textbf{\modelourslora}), we use \modelours as a starting point.\footnote{We include its task head since we found that it improves stability. We also experimented with applying LoRA to \modeloursft, but did not find it to improve upon \modelourslora.} 
We use the respective training set using max. $10,\!000$ sentences. 

The full details of the experiment setup regarding the datasets, infrastructure, training, and hyperparameters are provided in Appendix~\ref{sec:appendix-experiment-details}.

\section{Results}
\label{sec:results}

\subsection{Performance on Clean Text}
Table~\ref{tab:main-results} shows evaluation results on clean text, averaged over OPUS100, UD, and Ersatz on a diverse selection of languages, including an average over 81 languages.\footnote{
For the average, we only consider languages with datasets with both train and test sets for a fair comparison.
While we evaluate on 85 languages, this is the case in 81 languages.
}
We categorize methods into (i) \textit{multilingual}, which take only text as input, and (ii) \textit{monolingual}, which additionally rely on a language code or, in the case of \modelwtpt, \modelwtppunct, and \modelourslora, are adapted to a target domain.


Both \modelours and \modeloursft outperform the current non-domain-adapted SoTA model, \modelwtp.
Meanwhile, unlike \modelwtp, our models do not rely on specifying a language code as input.

Remarkably, \modeloursft and \modelwtppunct are not statistically significantly different, achieving average F1 scores of $91.6$ and $91.7$ respectively.
This is despite \modelwtppunct relying on adaptation to a target sentence-segmented corpus, whilst \modeloursft is a general-purpose multilingual model.
Finally, \modelourslora significantly outperforms the existing domain-adapted SoTA, \modelwtppunct, making it the best overall model. Our domain-adapted model outperforms \modelwtppunct in 63 out of 81 languages.

Among the LLMs, \modelcommandr, despite being trained in 23 languages, does surprisingly poorly, with \modelllama surpassing it by $23.5\%$ absolute avg. F1. 
Nevertheless, \modelllama still falls short compared to all variants of \modelours.
On the English benchmarks, given the abundance of English text, we expected our models to be easily outperformed by LLMs; 
yet, unlike \modelwtpu, \modeloursft outperforms both LLMs on \emph{every} dataset.


We provide full per-dataset results, including all 85 languages, in §~\ref{sec:appendix-add-results}. We also conduct ablation studies on each of our stages' components in §~\ref{sec:appendix-ablations}.

\begin{table}[t]
   \centering
   \def\arraystretch{0.87}
   \small
   \scalebox{0.9}{
        \begin{tabular}{lcccccc}
        \toprule
        \multirow{2}{*}[-0.3em]{\textbf{Model}} & \multicolumn{2}{c}{Tweets} & \multicolumn{2}{c}{Sentence Pairs} & \multirow{2}{*}{\textbf{\thead{Macro\\Avg.}}} \\
        \cmidrule(lr){2-3}
        \cmidrule(lr){4-5}
                       & {\texttt{sl}} & {\texttt{sr}} & {Speeches} & {Ersatz} \\
        \midrule
        \modelllama    & 73.4 & 76.0 & 66.9 & 94.8 & 77.8 \\
        \modelcommandr & 53.8 & 47.4 & 23.0 & 70.0 & 48.6 \\
        \midrule
        \modelwtpu     & 70.8 & 71.4 & 12.6 & 78.0 & 58.2 \\
        \modelwtpt     & 70.4 & 71.4 & 18.9 & 79.0 & 59.9 \\
        \modelwtppunct & 80.1 & 82.3 & 37.9 & 91.5 & 72.9 \\
        \midrule
        \modelours     & 80.5 & 75.5 & 28.8 & 84.0 & 67.2 \\
        \modeloursft   & 78.0 & 72.9 & 41.7 & 92.5 & 71.3 \\
        \modelourslora & \textbf{87.2} & \textbf{89.1} & \textbf{56.8} & \textbf{93.9} & \textbf{81.8} \\
        \bottomrule
        \end{tabular}
    } 
    \caption{Proportion of perfectly segmented short sequences. For Speeches and Ersatz, we are averaging scores over languages. We use 12-layer versions of each model given the task's increased difficulty.\tablefootnote{We exclude other baselines since none of them support \texttt{sl}/\texttt{sr} or all languages from TED/Ersatz.}}
    \label{tab:short-sequences}
\end{table}

\subsection{Performance on Noisy and Short Text}
Table~\ref{tab:ted2020-shared-task} presents the results of our method when evaluated on the SEPP-NLG Shared Task.
\modeloursft establishes a new state-of-the-art, outperforming the SEPP-NLG winners.
This is despite our model supporting 81 additional languages and use cases not considered in the Shared Task.

Furthermore, Table~\ref{tab:short-sequences} shows evaluation results on short sequences, including tweets and sentence pairs taken from manually corrupted speeches and Ersatz.
We observe similar patterns on these corpora: \modelours and \modeloursft outperform \modelwtpu, improving avg.~F1 scores by $9\%$ and $13.1\%$, respectively, 
\modelourslora continues to be the best overall model, also outperforming both LLMs.
We additionally provide an ablation study showing the importance of limited lookahead in \modelours in Table~\ref{tab:ll-ablation}.

\begin{table}[t]
   \centering
   \def\arraystretch{0.9}
   \small
   \scalebox{0.9}{
        \begin{tabular}{lccccc}
        \toprule
        \textbf{Model} & \multicolumn{1}{c}{{\thead{\texttt{es}\\ \texttt{en}}}} & \multicolumn{1}{c}{{\thead{\texttt{de}\\ \texttt{en}}}}  & \multicolumn{1}{c}{{\thead{\texttt{vi}\\ \texttt{en}}}} & \multicolumn{1}{c}{{\thead{\texttt{tr}\\ \texttt{de}}}} & {\textbf{\thead{Macro\\Avg.}}} \\
        \toprule 
        \modelllama    & 47.9 & 56.3 & 35.5 & 33.9 & 43.4 \\
        \modelcommandr & 30.4 & 51.9 & 30.0 & 17.6 & 32.5 \\

        \midrule
        \modelspacydp\hspace{-0.5ex}*  & 17.6 & 8.6 & 11.3 & 12.2 & 12.2 \\
        \modelwtpu\hspace{-0.5ex}*     & 38.6 & 39.0 & 25.5 & 33.5 & 29.1 \\
        \modelwtpt\hspace{-0.5ex}*     & 52.2 & 45.7 & 46.7 & 34.4 & 43.2 \\
        \modelwtppunct\hspace{-0.5ex}* & 62.1 & 60.1 & 59.0 & 41.0 & 54.9 \\
        \midrule
        \modelours     & 54.5 & 49.2 & 49.3 & 39.8 & 48.2 \\
        \modeloursft   & 59.6 & 58.4 & 57.3 & 42.4 & 54.4 \\
        \modelourslora & \textbf{65.0} & \textbf{65.6} & \textbf{67.5} & \textbf{48.8} & \textbf{61.7} \\
        \bottomrule
        \end{tabular}
    } 
    \caption{Sentence segmentation F1 scores for code-switched text.
   We use 12-layer versions of each model. * indicates models using language codes, where we try both language codes and show the better score. We show results using both language codes in Appendix~\ref{sec:appendix-add-results}.
   }
    \label{tab:code-switching}
\end{table}

\subsection{Performance on Challenging Domains}
\label{ss:challenging_domains}

\sparagraph{Code-switching}
The results in Table~\ref{tab:code-switching} reveal that
\modelwtpu achieves an average F1 score of $29.1\%$, while the highest-performing LLM scores $43.4\%$. \modelours and \modeloursft achieve average F1 of $48.2\%$ and $54.4\%$, respectively.
\modelourslora continues to improve performance, achieving $61.7\%$.
To the best of our knowledge, this is the first comprehensive evaluation of sentence segmentation tools on code-switching text.
While our models now represent SoTA, the evaluation results indicate that it is a challenging task.

We now evaluate domain adaptation performance of our method on two highly distinct domains: lyrics and legal data.

\rparagraph{Lyrics}
Table~\ref{tab:main-lyrics} shows results on verse segmentation (\ie segmenting songs into verse, chorus, bridge, \etc).
None of the other baseline systems, including LLMs, can improve over the current domain-specific SotA, $\text{SSM}_{\text{string}}$.
In contrast, \modelourslora outperforms $\text{SSM}_{\text{string}}$ by $10\%$ avg. F1. 
The difference is especially pronounced in hard-to-segment songs that are low in repetitiveness (\eg Rap music), with a $15\%$ difference in F1 scores.
When evaluating \modelourslora on manually corrupted lyrics, it still outperforms \emph{all} baselines,
even when compared to baselines evaluated on non-corrupted songs.
Additionally, \modelourslora\hspace{-0.75ex}@\hspace{-0.5ex}1000, using 1000 songs per genre for adaptation, still outperforms all baselines.
We provide complete results, including those for each genre, in Appendix~\ref{sec:appendix-add-results}.

\rparagraph{Legal and qualitative examples.} We provide comprehensive results on MultiLegalSBD in Appendix~\ref{sec:appendix-add-results}. Finally, We provide qualitative examples from several domains in Appendix~\ref{sec:appendix-examples}.

\begin{table}[t]
   \centering
   \def\arraystretch{0.88}
   \small

   \scalebox{0.9}{
        \begin{tabular}{lccccc}
        \toprule
        \multirow{2}{*}[-0.3em]{\textbf{Model}} & \multicolumn{2}{c}{Corrupted?} & \multicolumn{3}{c}{Repetitiveness} \\
        \cmidrule(lr){2-3} \cmidrule(lr){4-6}
         & \textbf{\checkmark} & \textbf{\ding{55}} & High & Mid & Low \\
        \midrule
        $\text{SSM}_{\text{string}}$$^{\dagger}$ & - & 63.8 & 71.3 & 64.8 & 47.3\\
        \midrule
        \modelllama & 45.5 & 49.7 & 48.9 & 46.7 & 33.8 \\
        \modelcommandr & 36.3 & 38.3 & 38.0 & 37.1 & 28.7 \\
        \midrule
        \modelwtppunct\hspace{-0.75ex}@\hspace{-0.5ex}100  & 46.9 & 53.8 & 55.8 & 55.2 & 35.9\\
        \modelwtppunct\hspace{-0.75ex}@\hspace{-0.5ex}1000 & 49.1 & 56.1 & 58.4 & 57.5 & 44.9 \\
        \modelwtppunct      & 49.2 & 56.2 & 58.4 & 57.6 & 44.9 \\
        \midrule
        \modelourslora\hspace{-0.75ex}@\hspace{-0.5ex}100  & 60.8 & 62.4 & 67.8 & 62.9 & 51.6\\
        \modelourslora\hspace{-0.75ex}@\hspace{-0.5ex}1000 & 67.3 & 72.4 & 76.5 & 73.1 & \textbf{62.7} \\
        \modelourslora      & \textbf{68.5} & \textbf{73.8} & \textbf{77.9} & \textbf{74.8} & 62.3 \\
        \bottomrule
        \end{tabular}
    } 
    \caption{Macro-averaged verse segmentation performance over per-genre F1 scores. $^{\dagger}$Values for $\text{SSM}_{\text{string}}$ taken from \citet{fell-etal-2018-lyrics}, with lyrics already pre-segmented into lines. @N corresponds to using a maximum of N songs per genre for adaptation.}
    \label{tab:main-lyrics}
    \vspace{-3mm}
\end{table}


\section{Discussion}
\label{sec:Discussion}

\sparagraph{LLMs}
Contrary to our expectations, our evaluation results reveal that LLMs generally underperform, particularly in non-English languages.
Notably, when using LLMs for sentence segmentation via prompting, each sentence is processed twice -- once as part of the input, appended to the prompt, and once within the output.
This redundancy leads to inefficient processing, needing to copy the input verbatim to the output, ideally only adding newlines.
However, in practice, LLMs are highly prone to alter their input~\citep{Barbero2024TransformersNG}. We found this issue to be particularly severe for noisy text and lyrics.\footnote{\modelllama and \modelcommandr altered $1.5\%$ and $2\%$ of all characters within lyrics, respectively, even though we prompted them not to alter their input (\cf~Appendix~\ref{sec:appendix-experiment-details}).}
This is highly problematic for a specific task requiring input and output characters to remain the same. Still, we tried to address this by using the Needleman-Wunsch sequence alignment algorithm to make pure segmentation performance comparable to other methods.\footnote{The same objective could be achieved via other means, \eg constrained decoding~\citep{BeurerKellner2024GuidingLT}.}

Aiming to improve the segmentation performance of LLMs, we experimented with few-shot prompting. However, this did not yield the desired improvements; in fact, it degraded performance.
Additionally, we tested varying the number of input-output sentences. The results of both of these ablation studies are presented in Appendix~\ref{sec:appendix-ablations}. 

\begin{figure}[t]
    \center
    \centering
    \includegraphics[width=0.86\linewidth]{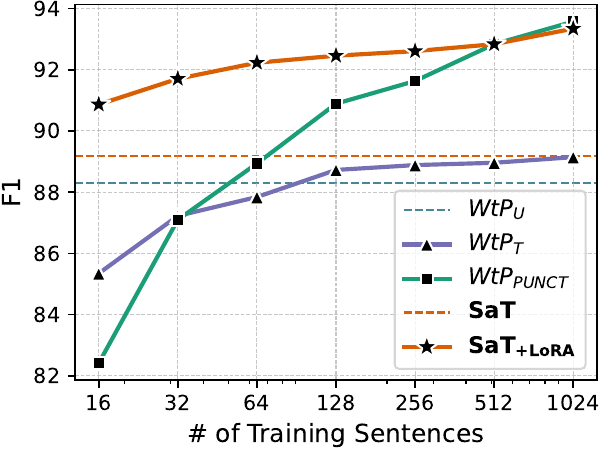}
    \caption{
        Macro avg. F1 vs. number of sentences used for adaptation, averaged over languages in \{OPUS100, UD, Ersatz\}. 
        Per-corpus results shown in Appendix~\ref{sec:appendix-ablations}.
    }
    \label{fig:few-shot}
    \vspace{-3mm}
\end{figure}

\rparagraph{Efficiency}
For our method, we rely on XLM-R as the LM backbone. Operating on subwords makes \modelours considerably faster than \modelwtp. We compare sentence segmentation performance and time on Ersatz across model sizes from 1-12 layers, illustrated in Figure~\ref{fig:pareto-plot}. Additional datasets are shown in Figure~\ref{fig:pareto-plot-all}.
The standard 3-layer variations of \modelours take $\approx0.5$ seconds to segment 1000 sentences on the hardware specified in Appendix~\ref{sec:appendix-experiment-details}, making them $3$ times faster than \modelwtp, while also outperforming \modelwtp models in \emph{all} sizes on Ersatz.
Furthermore,  for \modelours, performance plateaus with sizes $>3$ layers, whereas \modeloursft continues to improve when scaling up its size, making it by far the best model, despite never being exposed to Ersatz.

\rparagraph{Few-shot domain adaptation}
We now analyze how many sentences are needed to adapt our domain adaptation method, \modelourslora, to a target corpus, and compare it to previous methods.
As shown in Figure~\ref{fig:few-shot}, \modelwtpt and \modelwtppunct perform similarly when using $1024$ sentences for domain adaptation. However, \modelwtppunct fails to outperform the fully self-supervised variation of \modelwtp when $\leq32$ sentences are available. In contrast, \modelourslora markedly improves upon the self-supervised \modelours with only $16$ available sentences, and outperforms \modelwtppunct by almost $10\%$ in F1 score, making it substantially \emph{more sample-efficient}.

\section{Conclusion}
We proposed \modelours, an efficient, robust, and highly adaptable multilingual sentence segmentation method that neither relies on language codes nor punctuation. Further, we introduced \modeloursft,  improving \modelours via supervised adaptation using multiple corruption schemes.
Our method consistently achieves state-of-the-art performance among open weights models in experiments across 85 languages and eight diverse corpora, even outperforming newly introduced and optimized strong LLM baselines.
We also demonstrated that \modelours can be efficiently domain-adapted via LoRA, setting new performance standards on segmentation of lyrics and code-switching text.
Overall,
we hope \modelours will unlock significantly improved text data \mbox{(pre-)processing} across a range of NLP applications for multiple languages and domains via its robust and consistently strong \mbox{performance, versatility, and high efficiency}.

\section*{Limitations}
To the best of our knowledge, our evaluations are the most comprehensive to date, spanning 8 diverse corpora across different domains, languages, and noise levels, and sequence lengths. Still, we may not have covered every possible scenario.
Second, since we use \modelxlmr as our backbone, we also use its tokenizer, which has been shown to tokenize text less efficiently in some language~\citep{liang-etal-2023-xlm}, potentially exacerbating existing biases. We try to minimize bias \wrt performance by sampling text from all languages uniformly in both stages.
Furthermore, our use of subword LMs merges characters into subwords. Theoretically, this could limit sentence boundaries to end-of-token positions; however, in practice, we did not find this to be an issue.
Finally, 
language support could be further improved by \eg replacing mC4 with MADLAD-400~\citep{Kudugunta2023MADLAD400AM} for the pre-training stage. We leave this to future work.




\section*{Ethical Considerations}
Our work is multifaceted, as are the ethical dimensions it encompasses.
First, we acknowledge the possibility of NLP datasets and models for encoding unfair stereotypical~\citep{blodgett2020} and exclusive~\citep{dev-etal-2021-harms} biases that may lead to representational and allocational harms~\citep{barocas2017problem}. This issue is a general property of pre-trained LMs, and the models and datasets utilized in our study are similarly at risk.
We advise practitioners to use these models with the appropriate care and we refer to existing works \citep{bias_mitigation_liang, lauscher-etal-2021-sustainable-modular} for discussions on bias mitigation.
Second, one key aspect of our work deals with efficiency. On the one hand, considering the well-documented relationship between model training efforts and potential CO$_2$ emissions~\citep{strubell-etal-2019-energy}, our research contributes to Green AI by improving the environmental sustainability of state-of-the-art sentence segmentation systems.
On the other hand, since the training of language models often comes with high infrastructure prerequisites only available to certain user groups~\citep{StochasticParrots}, we hope that our work also contributes to the continued democratization of language technology by reducing resource- and language-related usage barriers.

\section*{Acknowledgments}
This research was funded in whole or in part by the Austrian Science Fund (FWF): \url{https://doi.org/10.55776/P33526}, \url{https://doi.org/10.55776/DFH23}, \url{https://doi.org/10.55776/COE12}, \url{https://doi.org/10.55776/P36413}.
In addition, Ivan Vuli\'{c} and Benjamin Minixhofer have been supported through the Royal Society University Research Fellowship \textit{‘Inclusive and Sustainable Language Technology for a Truly Multilingual World’} (no 221137) awarded to Ivan Vuli\'{c}. 
This research has also been supported with Cloud TPUs from Google's TPU Research Cloud (TRC). 
This work was also supported by compute credits from a Cohere For AI Research Grant, these grants are designed to support academic partners conducting research with the goal of releasing scientific artifacts and data for good projects.
We also thank Simone Teufel for fruitful discussions. 

\bibliography{anthology,cites}

\clearpage
\appendix

\section{Appendix}
\label{sec:appendix}
\begin{figure*}[t]
    \vspace{-6mm}
    \centering
    \begin{subfigure}[t]{0.325\textwidth}
        \centering
        \includegraphics[width=\linewidth]{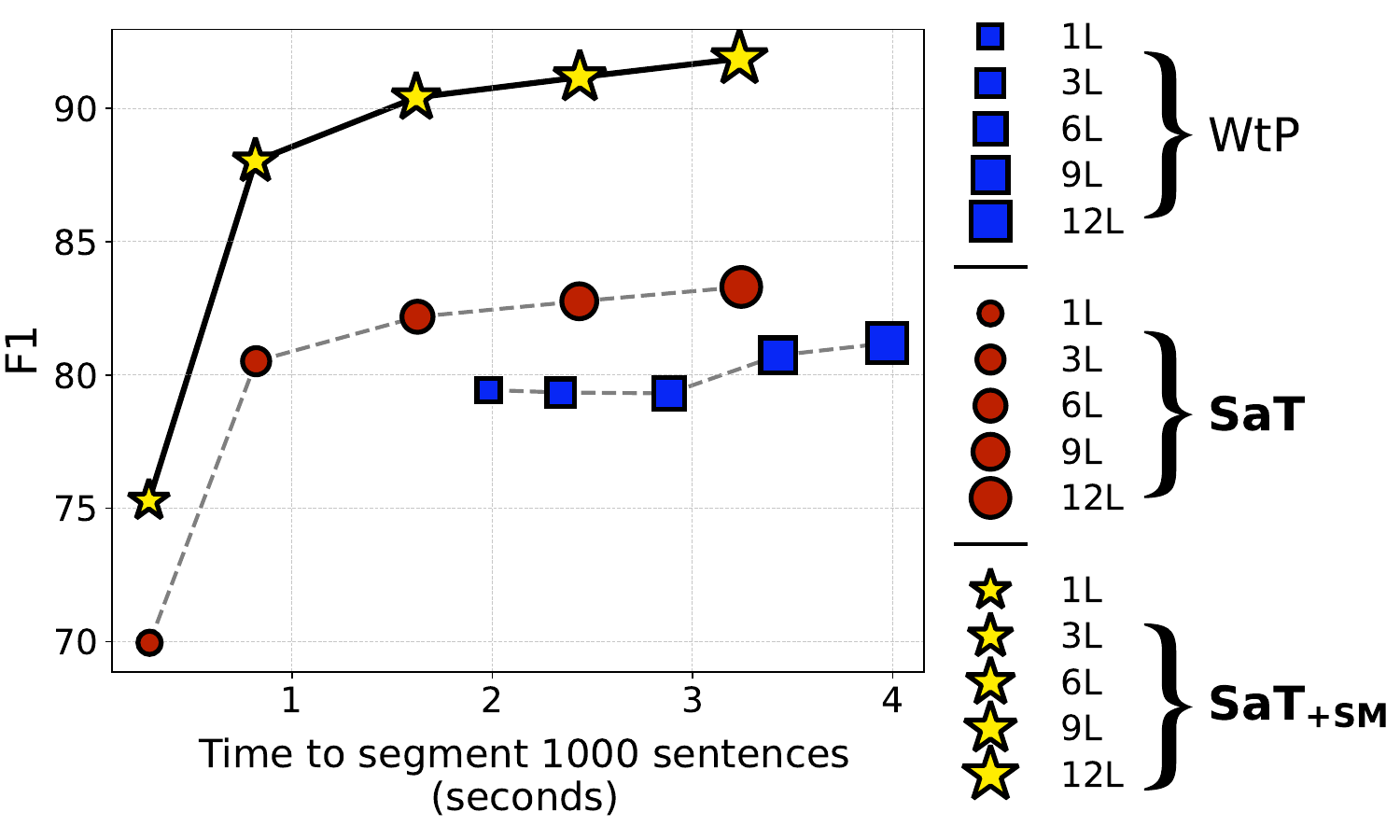}
        \caption{OPUS100}
        \label{fig:opus100-pareto}
    \end{subfigure}
    \hfill
    \begin{subfigure}[t]{0.325\textwidth}
        \centering
        \includegraphics[width=\linewidth]{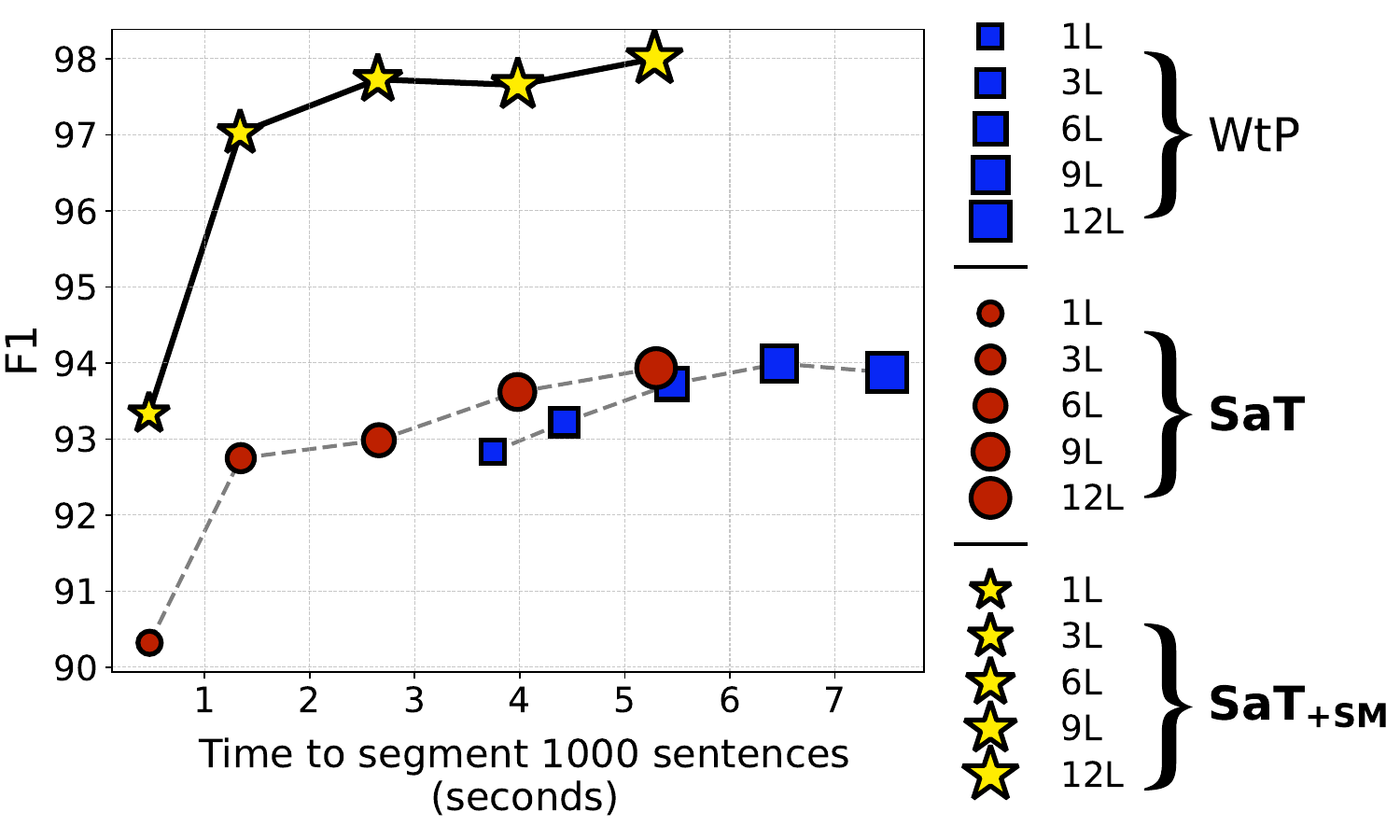}
        \caption{UD}
        \label{fig:ud-pareto}
    \end{subfigure}
    \hfill
    \begin{subfigure}[t]{0.325\textwidth}
        \centering
        \includegraphics[width=\linewidth]{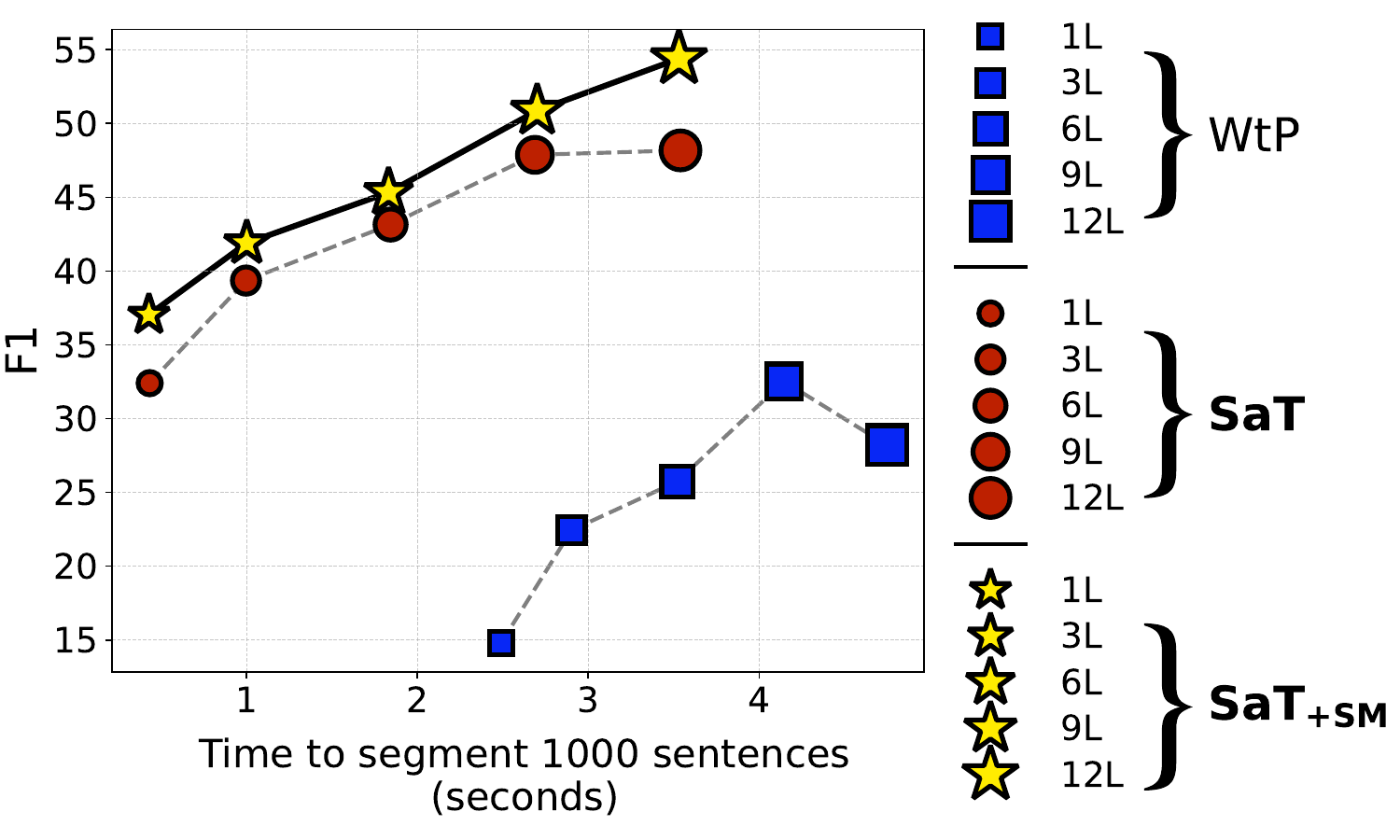}
        \caption{Code-Switching}
        \label{fig:code-switching-pareto}
    \end{subfigure}
    \caption{
    F1 scores and inference time for the prior SoTA (\modelwtpu) and our models (\modelours and \modeloursft), evaluated on additional sentence segmentation benchmarks. We average over all 23 languages and show the average time (10 runs) for variants with different sizes (L = \#layers) to segment 1,000 sentences using consumer hardware (1 Nvidia GTX 2080 Ti GPU).
    Performance on Ersatz is shown in Figure~\ref{fig:pareto-plot}.
    }
    \label{fig:pareto-plot-all}
\end{figure*}

\subsection{Ablation Studies}  
\label{sec:appendix-ablations}
    \begin{figure*}[t]
    \centering
    \begin{subfigure}[t]{0.395\textwidth}
        \centering
        \includegraphics[width=\linewidth]{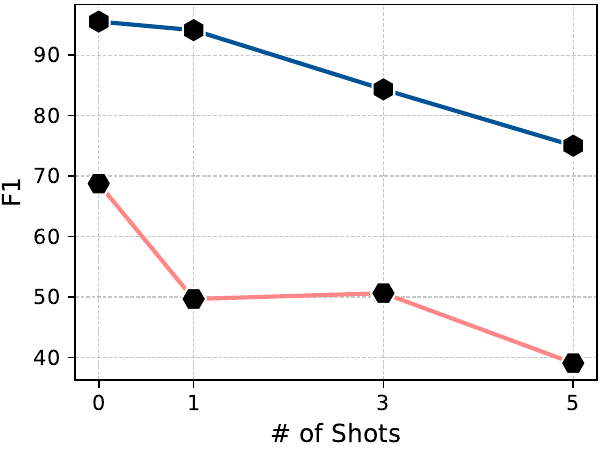}
        
        \caption{Effect of few-shot prompting.}
        \label{fig:llm-few-shot}
    \end{subfigure}
    \hfill
    \begin{subfigure}[t]{0.4\textwidth}
        \centering
        \includegraphics[width=\linewidth]{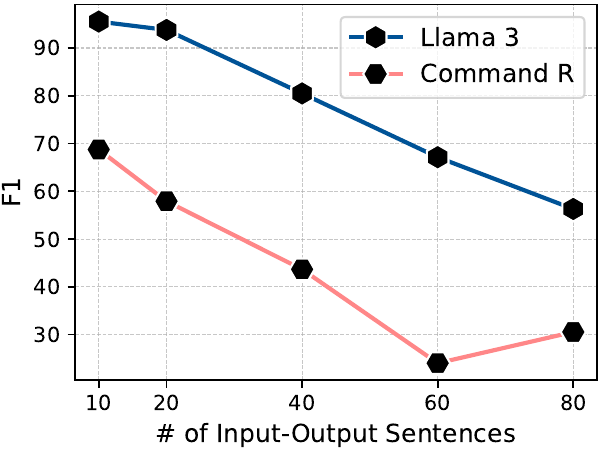}
        \caption{Effect of varying the number of input-output sentences.}
        \label{fig:llm-k}
    \end{subfigure}
    \hfill
    \caption{
        Ablation study on sentence segmentation performance of LLMs.
    }
    \label{fig:llm-ablation}
\end{figure*}

    \begin{table}[h]
   \centering
   \small
   \scalebox{0.85}{
        \begin{tabular}{llcccc}
        \toprule
        \textbf{Model} & \textbf{Variation} & \thead{Clean\\Text} & Tweets & \thead{Code\\Switching} & \\
        \midrule
        \multirow{2}{*}{\modelours} & - & \textbf{84.9} & 78.0 & \textbf{48.2} \\
        & Only clean text & 84.7 & 33.5 & 16.5 \\
        \midrule
        \multirow{3}{*}{\modeloursft} & - & \textbf{91.6} & 75.5 & \textbf{54.4} \\
         & Only clean text & 91.5 & \textbf{77.2} & 10.2\\
         & No pre-training & 89.9 & 42.1 & 44.0\\
        \midrule
        \multirow{2}{*}{\modelourslora} & - & \textbf{93.1} & \textbf{88.2} & \textbf{61.7} \\
        & No pre-training & 88.4 & 74.5 & 12.4 \\
        \bottomrule
    \end{tabular}
    } 
    \caption{
    Effect of various components of our method's variants. 
    We report macro average F1 scores for each domain and use models with the same number of layers for each category of text as in the main text.
    \textit{Only clean text} does not apply any corruptions.
    \textit{No pre-training} skips the paragraph segmentation stage on web-scale mC4, and thus starts from \modelxlmr weights.
    Best per-category results are \textbf{bold}.
   }
    \label{tab:components-ablation}
\end{table}
    \begin{table}[h]
   \centering
   \small
   \scalebox{0.85}{
        \begin{tabular}{llccccc}
        \toprule
        \multirow{3}{*}{\textbf{Model}} & \multirow{3}{*}{\textbf{Variation}} & \multicolumn{2}{c}{Lyrics} & \multicolumn{2}{c}{Legal} & \\
        & & \multicolumn{2}{c}{Corrupted?} & \multicolumn{2}{c}{Corrupted?} \\
        \cmidrule(lr){3-4} \cmidrule(lr){5-6}
        & & \textbf{\checkmark} & \textbf{\ding{55}} & \textbf{\checkmark} & \textbf{\ding{55}} \\
        \midrule
        \multirow{2}{*}{\modelourslora\hspace{-0.75ex}@\hspace{-0.5ex}100} & - & \textbf{60.8} & \textbf{62.4} & \textbf{81.1} & \textbf{93.6} \\
         & No pre-training & 20.3 & 34.3 & 61.2 & 79.8 \\
        \midrule
        \multirow{2}{*}{\modelourslora} & - & \textbf{68.5} & \textbf{73.8} & \textbf{83.3} & \textbf{95.1} \\
        & No pre-training & 59.9 & 62.1 & 82.5 & 94.9 \\
        \bottomrule
    \end{tabular}
    } 
    \caption{
    Effect of the web-scale pre-training stage on adaptation to hard domains, averaged over genres/legal categories. @100 corresponds to using a maximum of 100 songs/documents per genre/category for adaptation.
   }
    \label{tab:components-ablation-lora}
\end{table}

    \paragraph{\modelours components.} We show the effect of removing different components of our corruption schemes used in \modelours and \modeloursft in Table~\ref{tab:components-ablation}. 
    For \modelours, only using clean text even slightly hurts performance on clean text and strongly degrades performance in our more noisy tweets and code-switching evaluations.
    A similar pattern occurs for \modeloursft: Only using clean text hurts performance. Moreover, skipping the web-scale pre-training stage (\textit{No pre-training}) also decreases performance to a large extent, with the difference being particularly large for tweets and code-switching.
    Finally, for \modelourslora, \emph{no pre-training} similarly degrades performance, with the difference being particularly large in code-switching, where \modelourslora is better by $49.3\%$ absolute F1.
    
    Moreover, Table~\ref{tab:components-ablation-lora} compares domain adaptation performances via LoRA to models without our web-scale pre-training to \modelours models with it.
    As observed before, \emph{no pre-training} markedly degrades performance in both lyrics and legal data. The difference is especially large in cases where only 100 songs or documents are available, clearly showing that our pre-training stage \emph{improves sample-efficiency}.

    \paragraph{Limited lookahead.} We further provide an ablation study on the effect of disabling the limited lookahead mechanism using sentence pairs in Table~\ref{tab:ll-ablation}.
    Without limited lookahead, \modelours is outperformed by \modelwtpu. On the contrary, with limited lookahead, \modelours outperforms \modelwtpu by a considerable margin, where the difference is even more pronounced for 12-layer variations.
    Moreover, \modeloursft hardly benefits from limited lookahead, justifying our decision to disable it for \modeloursft.

    \paragraph{Effect of model size.} Figure~\ref{fig:pareto-plot-all} shows the effect of scaling up model sizes on OPUS100, UD, and code-switching, respectively.
    Remarkably, all 3-layer variations of \modeloursft clearly outperform \modelwtpu, despite not relying on language codes and being $\approx5$x faster.
    The difference is particularly pronounced in code-switching, where even the 1-layer variations of both \modelours and \modeloursft outperform the best variation of \modelwtpu.
    In general, performance continues to increase when further scaling up model sizes up to 12 layers. 

    \begin{table}[h]
   \centering
   \small
   \scalebox{0.85}{
        \begin{tabular}{ccccccc}
        \toprule
        \textbf{Model} & \textbf{Layers} & \textbf{\thead{Look-\\ahead}} & OPUS100 & UD  & Ersatz  & TED \\
        \midrule
        \multirow{2}{*}{\modelwtpu} & 3 & $\infty$ & 52.4 & 80.6 & 78.0 & 9.8 \\
        \cmidrule{2-7}
        & 12 & $\infty$ & 52.8 & 77.9 & 78.0 & 12.6\\
        \midrule
        \multirow{4}{*}{\modelours} 
        & \multirow{2}{*}{3} & $\infty$ & 4.4 & 1.8 & 3.5 & 1.9 \\
        & & 48 & 56.9 & 82.4 & 82.2 & 16.9 \\
        \cmidrule{2-7}
        & \multirow{2}{*}{12} & $\infty$ & 31.0 & 55.0 & 55.5 & 20.9 \\
        & & 48 & \textbf{63.3} & \textbf{85.2} & \textbf{84.0} & \textbf{28.8} \\
        \midrule
        \multirow{4}{*}{\modeloursft} 
        & \multirow{2}{*}{3} & $\infty$ & 72.2 & 91.4 & 85.9 & 29.4  \\
        & & 48 & 73.7 & 93.1 & 85.8 & 28.4 \\
        \cmidrule{2-7}
        & \multirow{2}{*}{12} & $\infty$ & 78.0 & 93.5 & \textbf{92.5} & \textbf{41.7} \\
        &  & 48 & \textbf{78.6} & \textbf{93.6} & 91.3 & 38.3 \\
        \bottomrule
        \end{tabular}
    } 
    \caption{
    Proportion of perfectly segmented sequences within additional corpora.
   }
    \label{tab:ll-ablation}
\end{table}

    \paragraph{LLMs.} Figure~\ref{fig:llm-ablation} shows the effect of few-shot prompting and varying the number of input-output sentences for \modelllama and \modelcommandr.
    Contrary to our expectations, in-context learning via few-shot prompting does not improve sentence segmentation performance for both LLMs in consideration. Providing only a single example already degrades performance, and providing more examples further degrades it.
    Furthermore, increasing the number of input-output sentences from our favorably low default of $10$ results in considerable performance decreases. Notably, when using $80$ input-output sentences per chunk, both LLMs achieve F1 scores of only $<60\%$.


\subsection{Complete Experiment Details}
\label{sec:appendix-experiment-details}
    \begin{table}[h]
   \centering
   \small
   \scalebox{0.83}{
        \begin{tabular}{cccc}
        \toprule
        \multirow{2}{*}[-0.3em]{Language} & \multicolumn{2}{c}{Number of Sentences} & \multirow{2}{*}[-0.3em]{Source} \\
        \cmidrule(lr){2-3}
                       & Train & Test & \\
        \midrule
        sl & 2728 & 2728 & \citet{fivser2020janes}\\
        sr & 1727 & 192 & \citet{Miličević_Ljubešić_2016}\\
        \midrule
        es-en & 1335 & 1334 & \citet{deuchar2009miami}\\
        en-de & 678 & 599 & \citet{osmelak-wintner-2023-denglisch}\\
        tr-de & 578 & 805 & \citet{cetinoglu-2017-code}\\
        vi-en & 1360 & 1361 & \citet{nguyen-bryant-2020-canvec}\\
        \bottomrule
        \end{tabular}
    } 
    \caption{Number of train and test sentences from tweets and code-switched text, including their source.}
    \label{tab:noisy-short-stats}
\end{table}

    \begin{table}[h]
   \centering
   \small
   \scalebox{0.85}{
        \begin{tabular}{llcc}
        \toprule
        \multirow{2}{*}[-0.3em]{Repetitiveness} & \multirow{2}{*}[-0.3em]{Genre} & \multicolumn{2}{c}{Number of Songs}\\
        \cmidrule(lr){3-4}
                       & & Train & Test \\
        \midrule
        \multirow{3}{*}{High} & Punk Rock        & 778 & 190 \\
                              & Pop Punk         & 512 & 141 \\
                              & Country          & 2916 & 711 \\
        \midrule
        \multirow{11}{*}{Mid} & Rock             & 4611 & 1182 \\
                              & Pop              & 3490 & 891 \\
                              & RnB              & 3542 & 915 \\
                              & Alternative Rock & 3370 & 856 \\
                              & Alternative Metal & 651 & 155 \\
                              & Soul             & 494 & 110 \\
                              & Hard Rock        & 1821 & 430 \\
                              & Indie Rock       & 1193 & 305 \\
                              & Pop Rock         & 1633 & 412 \\
                              & Heavy Metal      & 988 & 216 \\
                              & Indie Rock       & 1193 & 305 \\
        \midrule
        \multirow{2}{*}{Low}  & Southern Hip Hop & 836 & 208 \\
                              & Gangsta Rap      & 270 & 64 \\
        \bottomrule
        \end{tabular}
    } 
    \caption{Number of train and test songs per genre.}
    \label{tab:music-stats}
\end{table}

    \begin{table}[h]
   \centering
   \small
   \scalebox{0.85}{
        \begin{tabular}{cccccc}
        \toprule
        \multirow{3}{*}[-0.3em]{Language} &  \multicolumn{4}{c}{Number of Documents}\\
        \cmidrule(lr){2-6}
                       & \multicolumn{2}{c}{Laws} &  \multicolumn{2}{c}{Judgements} \\
                       \cmidrule(lr){2-3}
                       \cmidrule(lr){4-5}
                            & Train & Test & Train & Test \\
                       
        \midrule
        de & 10 & 3 & 104 & 27 \\
        en & - & - & 64 & 16 \\
        es & 494 & 183 & 151 & 39 \\  
        fr & 1672 & 459 & 252 & 63 \\
        it & 2206 & 704 & 194 & 49 \\
        \bottomrule
        \end{tabular}
    } 
    \caption{Number of legal train and test documents per category. We discard Portuguese since there is no training data available.}
    \label{tab:legal-stats}
\end{table}
    \
    \paragraph{Dataset details.}

        We give an overview of all used languages and their evaluation dataset sizes for clean text in Table~\ref{tab:list_of_languages}.
        Furthermore, we provide statistics of splits for noisy text and additional domains in Table~\ref{tab:noisy-short-stats} for tweets and code-switching, Table~\ref{tab:music-stats} for lyrics, and Table~\ref{tab:legal-stats} for legal data.

        If a given corpus does not provide train and test splits, we set aside 10,000 sentences for testing and keep the rest for training if more than 10,000 sentences are available. If a corpus is smaller, we set aside 50\% for testing and use the rest for adaptation.
        To train \modeloursft, if neither UD nor OPUS100 train data is available, we resort to NLLB. This is the case in Cebuano (\texttt{ceb}), Javanese (\texttt{jv}), Mongolian (\texttt{mn}), and Yoruba (\texttt{yo}), where we take 10,000 sentences each.

        To simulate our real-time automatic speech recognition (ASR) scenario, we take publicly available TED talk transcripts in 76 languages, available at \url{opus.nlpl.eu/TED2020/corpus/version/TED2020}. We generate non-overlapping sentence pairs as done in other experiments to evaluate performance on short sequences. Additionally, we remove fully lowercase all pairs and all punctuation tokens. We generally derive punctuation tokens with the commonly used Moses tokenizer~\citep{koehn-etal-2007-moses} for languages where it is available. For all other languages, we simply remove all punctuation characters.

        Moreover, we observe that used tweets in \texttt{sl} and \texttt{sr} are inconsistent \wrt segmenting single emojis. We thus filter out all emojis as a simple pre-processing step. Similarly, we normalize tweets by filtering out words starting with \textit{http}, \textit{\#}, and \textit{@}.

    \paragraph{Computing infrastructure.}
        We train \modelours on a TPUv4 VM with 8 cores, \modelourslora on a TPUv3 VM with 1 core, and \modeloursft using a single A100 GPU.
        To measure inference time, we use a consumer-grade GPU, the Nvidia GTX 2080 Ti with an AMD EPYC 7402P CPU.
    \paragraph{Implementation details.}
        We use the \texttt{PyTorch}~\citep{pytorch} and \texttt{transformers}~\citep{wolf-etal-2020-transformers} libraries for all experiments.
        For adaptation via LoRA, we make use of the \texttt{adapters} library~\citep{poth-etal-2023-adapters, pfeiffer2020AdapterHub} library, a wrapper around the \texttt{transformers} library. Our code and models are released under the MIT License, ensuring open access to the community for further development.
    \paragraph{Training of \modelours.}
        We train \modelours using a context window of $256$ since we observed that it improves performance. During inference, we use the full context size of \modelxlmr, $512$.
        Moreover, we follow \citet{minixhofer-etal-2023-wheres} and sample paragraphs to ensure that a maximum of 10\% of paragraphs do not end in punctuation (except for Thai, which does not use sentence-ending punctuation). We also sample paragraphs of languages uniformly.
        We continue training \modelxlmr on naturally occurring newline symbols for 200k training steps using a batch size of 512. We use a linearly increasing learning rate warmup from 0 to 1e-4, and decay the learning rate to 0 for the remaining 195k steps. We use the AdamW optimizer~\citep{AdamW}.
        For the auxiliary objective as introduced by \citet{minixhofer-etal-2023-wheres}, we set the removal probability $p=0.25$ using the union of the 30 most common punctuation characters in every language. We then take the corresponding tokens as used by \modelxlmr as labels for the auxiliary objective.

        For models without limited lookahead (cf. Table~\ref{tab:ll-ablation}), we follow \citet{minixhofer-etal-2023-wheres} and use a threshold of $0.01$ for sentence boundary detection. When using limited lookahead, we observe that the optimal threshold increases. We thus use a constant threshold of $0.025$ with limited lookahead.

    \paragraph{Training of \modeloursft.}
        In our supervised mixture stage, we continue training \modelours using the same context window of $256$.
        The training data now consists of sentence-segmented text, and we train \modeloursft predicting sentence-ending tokens. 
    
        For each language, we corrupt the data in two ways.
        In the first, we lowercase and remove all punctuation tokens.
        This aims to roughly emulate the output of an automatic speech recognition (ASR) system.
        In the second, we lowercase all text with probability $0.5$, remove all punctuation with probability $0.5$, duplicate punctuation (\eg changing ! to !!!) with geometric distribution scaling with the number of duplications (\ie doubling with probability $0.5$, tripling with probability $0.25$, \etc), and join sentences without a whitespace with probability $0.1$ (or with a space for the four languages which do not generally use a whitespace to split sentences, see Table~\ref{tab:dataset-summary}.)
        This aims to emulate user-generated text.
        
        We generally pack sentences into chunks.
        For the uncorrupted sentences and sentences corrupted with the first scheme, we pack until each chunk fully fills up the model's context window.
        For our second corruption scheme, we include $s$ sentences in each block, where $s$ is drawn from the same geometric distribution as used before.
    
        We train with a batch size of 128, linear learning rate warmup from 0 to 3e-5 for 500 steps, and linearly decay for another 19,500 steps.
        We uniformly sample batches of sentences from a single language and evenly sample batches corrupted with one of the three corruption schemes.
        During inference, we use a constant threshold of $0.25$.

    \paragraph{Training of \modelourslora.} 
        For adaptation via LoRA, we use a learning rate of 3e-4. We train LoRA modules with AdamW for a target domain for 30 epochs, where we linearly warm up the learning rate for the first 10\% of training, followed by a decay to 0 for the remaining training steps. We do not apply early stopping. We apply LoRA to the query and value matrices of the attention block, as well as the intermediate layer of the Transformer, using a rank $r=16$ and scaling factor $a=32$. We noticed this to positively impact performance at a comparably low computational cost. Moreover, similarly to \modelwtpt and \modelwtppunct, we additionally tune the classification threshold on the same training data if more than $512$ \textit{sentences} are available. We noticed that this helps performance in such cases.
        For verse segmentation, we use \modelours models without limited lookahead, since, with verses, it is both helpful and desirable to rely on future verses.

    \paragraph{LLM details.}
    We use default hyperparameters for both \modelllama and \modelcommandr. For \modelcommandr, We used the Cohere API. This led to some API refusals, particularly for lyrics. We thus only consider chunks that were not refused when calculating metrics.
    To align input and output chunks using the Needleman-Wunsch sequence alignment algorithm, we use a gap penalty of $-0.5$, a gap extension penalty of $-0.5$, a match reward of $1$, and a mismatch penalty of $-0.5$. If no alignment is found, the LLM produced output that strongly deviated from the input chunk. We thus assign no sentence boundaries to the predictions of the LLMs for this input chunk.

    We experiment with several prompts, optimizing performance on the training set, resulting in the following final prompt:
    \begin{tcolorbox}[colback=white, colframe=gray!60!white, sharp corners, boxrule=0.5mm, title=General LLM Prompt]
        Separate the following text into sentences by adding a newline between each sentence. Do not modify the text in any way and keep the exact ordering of words! If you modify it, remove or add anything, you get fined \$1000 per word. Provide a concise answer without any introduction. Indicate sentence boundaries only via a single newline, no more than this!
    \end{tcolorbox}

    We then append \textit{\textbackslash n\textbackslash n\# Input: \textbackslash n\textbackslash n}, followed by the input chunk, followed by \textit{\textbackslash n\textbackslash n\# Output: \textbackslash n\textbackslash n}, resulting in the complete input to the LLM.

    For few-shot prompting, we append the prompt with \textit{When provided with multiple examples, you are to respond only to the last one}. In addition, we indicate chunk $n$ with \textit{Input N: } and \textit{Output N: }, respectively.

    Since it is a highly distinct task, we use the following prompt for verse segmentation:
    \begin{tcolorbox}[colback=white, colframe=gray!70!white, sharp corners, boxrule=0.5mm, title=LLM Lyrics Prompt]
        Separate the following song's lyrics into semantic units (e.g., verse, chorus, bridge, intro/outro, etc - similar to how they are presented in a lyrics booklet) via double newlines, but do not annotate them. Only include the song in the output, no annotations. Do not modify the song in any way and keep the exact ordering of words! If you modify it, remove or add anything, you get fined \$1000 per word. Indicate semantic units by double newlines.
    \end{tcolorbox}

\subsection{Qualitative Examples}
\label{sec:appendix-examples}
    \paragraph{ASR output.} We show predictions of \modeloursft on parts of transcribed TED talks in different languages in Table~\ref{tab:example.ted}.

    \paragraph{Code-switching.} We also show predictions of \modeloursft on code-switching text in four language pairs in Table~\ref{tab:example.code-switching}.

    \paragraph{Verse segmentation.} In addition, we show predictions of \modelourslora on verse segmentation in Tables~\ref{tab:example.lyrics-high},  ~\ref{tab:example.lyrics-mid}, and ~\ref{tab:example.lyrics-low} for songs of high, mid, and low levels of repetitiveness, respectively.

\subsection{Additional Results} 
\label{sec:appendix-add-results}

\begin{figure*}[h]
    \centering
    \begin{subfigure}[b]{0.32\textwidth}
        \centering
        \includegraphics[width=\linewidth]{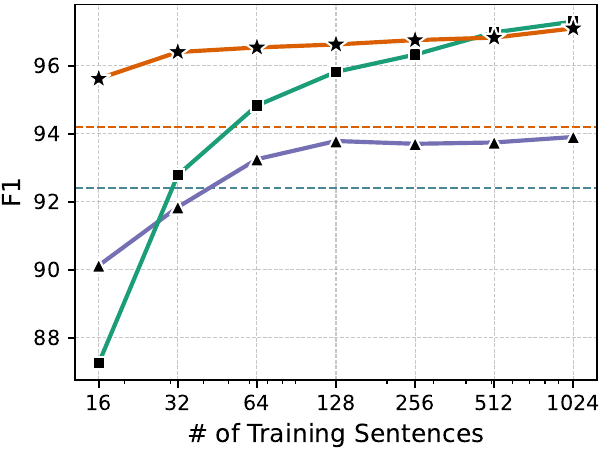}
        \caption{Ersatz}
        \label{fig:ersatz}
    \end{subfigure}
    \begin{subfigure}[b]{0.32\textwidth}
        \centering
        \includegraphics[width=\linewidth]{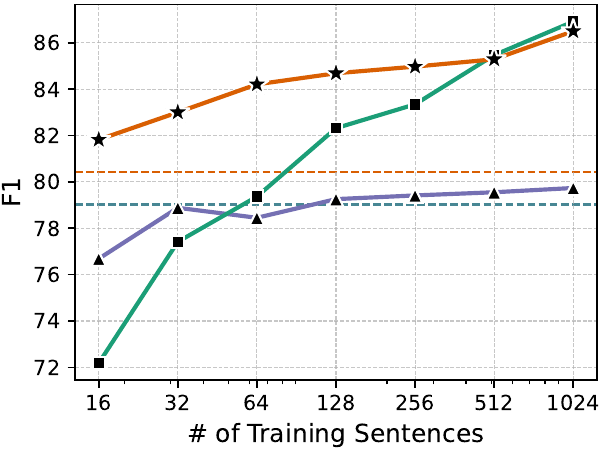}
        \caption{OPUS100}
        \label{fig:opus100}
    \end{subfigure}
    \begin{subfigure}[b]{0.32\textwidth}
        \centering
        \includegraphics[width=\linewidth]{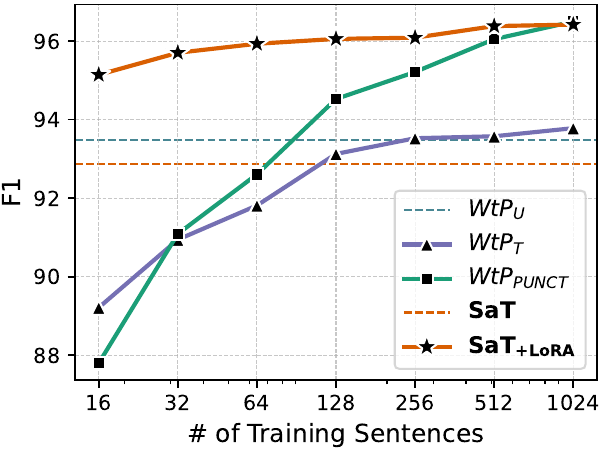}
        \caption{UD}
        \label{fig:ud}
    \end{subfigure}
    \caption{Avg. F1 vs. number of sentences used for adaptation, averaged over languages within a given dataset.}
    \label{fig:few-shot-all}
\end{figure*}


    \paragraph{Results in English.} We provide an overview of the performance of different models on different corpora in Table~\ref{tab:en-summary-results}.
    
\begin{table}[t]
   \centering
   \small
   \scalebox{0.76}{
        \begin{tabular}{lcccc}
        \toprule
        \textbf{Model} & \multicolumn{1}{c}{OPUS100} & \multicolumn{1}{c}{UD}  & \multicolumn{1}{c}{Ersatz} & \multicolumn{1}{c}{\textbf{\thead{Macro\\Avg.}}} \\
        \midrule
        \modelspacydpm & 88.8 & 91.7 & 94.0 & 91.5 \\
        \modelersatz & 87.6 & 89.1 & 97.5 & 91.4 \\
        \midrule
        \modelllama & 92.8 & 94.8 & \underline{98.2} & 95.3 \\
        \modelcommandr & 89.5 & 77.1 & 87.2 & 84.6 \\
        \midrule
        \modelours & 90.4 & 93.9 & 96.7 & 93.7 \\
        \modeloursft & \underline{\textbf{94.6}} & \underline{\textbf{96.7}} & \underline{\textbf{98.3}} & \underline{\textbf{96.5}} \\
        \thickmidrule
        \modelpunkt & 88.2 & 90.8 & 97.7 & 92.2 \\
        \modelpysbd & 59.6 & 75.3 & 73.9 & 69.6 \\
        \modelspacydp & 89.0 & 91.3 & 98.5 & 92.9 \\
        \modelwtpu & 90.6 & 94.5 & 96.5 & 93.9 \\
        \modelwtpt & 89.4 & 94.5 & 96.7 & 93.5 \\
        \modelwtppunct & \underline{94.7} & \underline{\textbf{96.9}} & \underline{98.6} & \underline{96.7} \\
        \midrule
        \modelourslora & \underline{\textbf{94.8}} & \underline{96.8} & \underline{\textbf{98.7}} & \underline{\textbf{96.8}} \\
        \bottomrule
        \end{tabular}
    } 
    \caption{
    English (\texttt{en}) sentence segmentation F1 scores. We use 3-layer versions of each model.
    Numerically best results are in \textbf{bold}, statistically indistinguishable ones from this best are \underline{underlined}.
   }
    \label{tab:en-summary-results}
\end{table}  

    \paragraph{Legal data.} Table~\ref{tab:all-legal-clean} shows the sentence segmentation performance of different models on MultiLegalSBD. Furthermore, Table~\ref{tab:all-legal-corrupted} shows performance on MultiLegalSBD when applying the same corruptions as on Speeches, removing all casing and punctuation tokens.

    \paragraph{More few-shot results.} Figure~\ref{fig:few-shot-all} shows the per-dataset few-shot domain adaptation results, comparing \modelwtpt, \modelwtppunct, and \modelourslora.

\begin{figure}[t]
    \centering
    \includegraphics[width=1\linewidth]{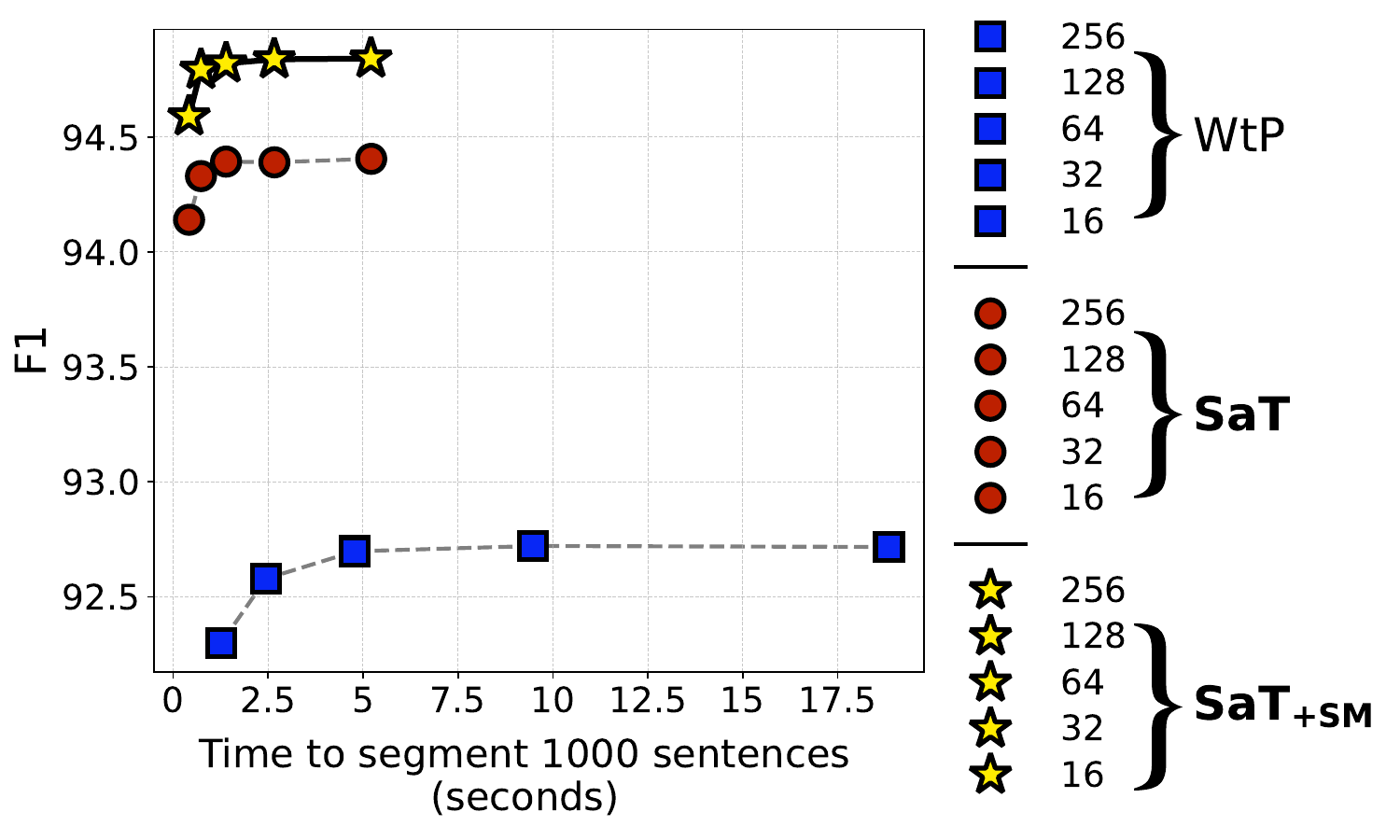}
    \caption{Sentence segmentation F1 scores vs. execution time across different strides (default 64), evaluated on Ersatz. We use the standard 3-layer variants of each model. Higher \texttt{stride} values result in faster inference. }
    \label{fig:strides}
    \vspace{-2mm}
\end{figure}

    \paragraph{Effect of stride.} For both \modelours and \modelwtp, we use a default stride of $64$ during evaluation. Each subword or character is thus processed multiple times, where we average predictions for overlapping positions. Since \modelours operates on subwords but \modelwtp on characters, this results in different scaling behaviors, illustrated in Figure~\ref{fig:strides}.

    \paragraph{Complete verse segmentation results.} We provide complete per-genre results for verse segmentation in Tables ~\ref{tab:high-low-lyrics-clean} and ~\ref{tab:mid-lyrics-clean}. Furthermore, Tables~\ref{tab:high-low-lyrics-corrupted} and \ref{tab:mid-lyrics-corrupted}. show verse segmentation performance when applying the same corruptions as on Speeches, removing all casing and punctuation tokens.

    \paragraph{Complete results on clean data.} Results of \modelours, its variations, and other methods on all languages are shown in Tables~\ref{tab:all_comparisons_1}-\ref{tab:all_comparisons_6}.

    \paragraph{Complete code-switching result.} We show results using both language codes on code-switched text for models using language codes in Table~\ref{tab:code-switching-all-codes}.

\begin{table}[t]
    \small
    \scalebox{0.76}{
         \begin{tabular}{lccccc}
         \toprule
         \textbf{Model} & \multicolumn{1}{c}{{\thead{\texttt{es}\\ \texttt{en}}}} & \multicolumn{1}{c}{{\thead{\texttt{de}\\ \texttt{en}}}}  & \multicolumn{1}{c}{{\thead{\texttt{tr}\\ \texttt{de}}}} & \multicolumn{1}{c}{{\thead{\texttt{vi}\\ \texttt{en}}}} & {\textbf{\thead{Macro\\Avg.}}} \\
         \toprule
         \modelpunkt    & 0.0/0.0 & 0.0/1.1 & 0.0/0.0 & -/0.0 & 0.3/- \\
         \modelpysbd    & 0.0/0.0 & 2.1/2.1 & -/0.0 & -/0.0 & 0.5/0.5 \\
         \modelspacydp  & 0.0/17.6 & 8.6/8.0 & -/12.2 & -/11.3 & 12.2/- \\
         \modelwtpu     & 20.8/38.6 & 39.0/31.4 & 33.5/21.0 & 22.7/25.5 & 29.1/29.0 \\
         \modelwtpt     & 46.9/52.2 & 45.7/39.1 & 33.3/34.4 & 46.7/36.7 & 40.6/43.2 \\
         \modelwtppunct & 60.7/62.1 & 60.1/58.1 & 39.9/41.0 & 59.0/50.7 & 53.0/54.9 \\
    
         \bottomrule
         \end{tabular}
    } 
    \caption{Complete sentence segmentation F1 scores for code-switched text for systems relying on language codes, where the first number corresponds to the first language shown.}
    \label{tab:code-switching-all-codes}
\end{table}

\subsection{Auxiliary Punctuation Prediction Objective}\label{sec:aux-obj}
    As mentioned in Section~\ref{sec:method}, we adopt the auxiliary punctuation prediction objective from WtP~\citep{minixhofer-etal-2023-wheres} as our base corruption scheme. 
    
    For clarity, we first specify the target without the auxiliary objective. Here, the original sequence of tokens $\bm{c}$ within some corpus is first stripped of newline characters:
    \begin{equation}
        \bm{x} = \{c_i \:|\: c_i \in \bm{c}, c_i \neq \mathtt{\backslash n}\}.
    \end{equation}
    We then create labels, which we set positive if the following token in the original sequence is a newline character:
    \begin{equation}
        \bm{y} = \left\{\!
        \begin{array}{l}
        \!1 \text{\; if $c_{i+1} = \mathtt{\backslash n}$}\\
        \!0 \text{\; otherwise}
        \end{array}
        \:|\: c_i \in \bm{x}
        \right\},
    \end{equation}
    where $c_i$ indexes into the original sequence $\bm{c}$. Using these labels, we optimize the standard cross-entropy of these labels and the model's predictions. Note that the newline character is not contained in our base model's vocabulary and will thus only appear as a single character. We also tokenize the whole batch at the start before applying any corruptions and do not re-tokenize later. We found this to be more effective than re-tokenizing the sequence after applying corruptions.
    
    \rparagraph{Auxiliary Punctuation Prediction} For the auxiliary objective, we adapt the methodology from \citet{minixhofer-etal-2023-wheres} to tokens and identify the union of the 30 most common punctuation-only \textit{tokens} within the training set. For simplicity, we ignore tokens containing multiple (potentially non-punctuation) characters. We also include the <UNK> token in this resulting set $P$, resulting in 109 punctuation tokens. We then define a random binary mask that determines which punctuation characters to remove among $P$, resulting in the new sequence $\bm{x}'$:
    \begin{equation}
        \bm{x}' = \left\{
        c_i \:|\:
        \begin{array}{l}
        c_i \in \bm{c}, c_i \neq \mathtt{\backslash n},\\
        c_i \notin P \text{ or } p_i = 0\\
        \end{array}
        \right\}
    \end{equation}
    Here, we do not remove two consecutive character tokens to be able to reconstruct the original sequence. In addition, unlike WtP, we only remove character tokens if the following token is not a newline token.
    For the remaining characters, the auxiliary labels $\bm{z}$ indicate which (if any) character among $P$ followed them in the original sequence:
    \begin{equation}
        \bm{z} = \left\{\!
        \begin{array}{l}
        \!c_{i+1} \text{\; if $c_{i+1} \in P$}\\
        \!0 \text{\;\;\;\;\;\;otherwise}
        \end{array}
        \:|\: c_i \in \bm{x}'
        \right\}
    \end{equation}
    To avoid needing two separate forward passes through the model, we substitute the input $\bm{x}$ with $\bm{x}'$ also for the main (newline prediction) objective.
    The final loss $\mathcal{L}$ is obtained by summing up the main newline prediction objective and the auxiliary objective of predicting punctuation:
    \begin{equation}
    \mathcal{L} = \mathcal{L}^{\mathrm{main}} + \mathcal{L}^{\mathrm{aux}}
    \label{equation:total_loss}
    \end{equation}

    \begin{table*}[t]
\centering
\small
\resizebox{\textwidth}{!}{%
\begin{tabular}{llc|rrrr|}
\toprule
Language & iso & Space & UD & OPUS100 & Ersatz & Speeches\\
\midrule
Afrikaans & af &  & AfriBooms (425) & 1.9k & & 1.3k\\
Amharic & am &  &  & 2.0k & & 514\\
Arabic & ar &  & PADT (680) & 2.0k & 1.5k & 10.0k\\
Azerbaijani & az &  &  & 2.0k & & 6.8k\\
Belarusian & be &  & HSE (1.1k) & 2.0k & & 6.2k\\
Bulgarian & bg &  & BTB (1.1k) & 2.0k & & 10.0k\\
Bengali & bn &  & BRU (56) & 2.0k & & 5.2k\\
Catalan & ca &  & AnCora (1.8k) & 2.0k & & 10.0k\\
Cebuano & ceb &  & GJA (188) &  & & 55\\
Czech & cs &  & PDT (10.1k) & 2.0k & 1.7k & 10.0k\\
Welsh & cy &  & CCG (953) & 1.8k & & -\\
Danish & da &  & DDT (565) & 2.0k & & 10.0k\\
German & de &  & GSD (977) & 1.9k & 2.0k & 10.0k\\
Greek & el &  & GDT (456) & 2.0k & & 10.0k\\
English & en &  & GUM (1.1k) & 2.0k & 7.7k & 10.0k\\
Esperanto & eo &  &  & 2.0k & & 10.0k\\
Spanish & es &  & AnCora (1.7k) & 2.0k & 3.1k & 10.0k\\
Estonian & et &  & EDT (3.2k) & 2.0k & 2.0k & 5.9k\\
Basque & eu &  & BDT (1.8k) & 2.0k & & 10.0k\\
Persian & fa &  & PerDT (1.5k) & 2.0k & & 10.0k\\
Finnish & fi &  & TDT (1.6k) & 2.0k & 2.0k & 10.0k\\
French & fr &  & GSD (416) & 2.0k & 1.7k & -\\
Western Frisian & fy &  &  & 1.9k & & 46\\
Irish & ga &  & IDT (454) & 2.0k & & -\\
Scottish Gaelic & gd &  & ARCOSG (545) & 1.1k & & 10.0k\\
Galician & gl &  & TreeGal (400) & 2.0k & & 7.9k\\
Gujarati & gu &  &  & 1.9k & 1.0k & 13\\
Hausa & ha &  &  & 2.0k & & 10.0k\\
Hebrew & he &  & IAHLTwiki (393) & 2.0k & & 10.0k\\
Hindi & hi &  & HDTB (1.7k) & 2.0k & 2.5k & 10.0k\\
Hungarian & hu &  & Szeged (449) & 2.0k & & 10.0k\\
Armenian & hy &  & BSUT (595) & 7.0k & & 104\\
Indonesian & id &  & PUD (1.0k) & 2.0k & & 1.6k\\
Igbo & ig &  &  & 1.7k & & 10.0k\\
Icelandic & is &  & IcePaHC (5.2k) & 2.0k & & 10.0k\\
Italian & it &  & ISDT (482) & 2.0k & & 5.8k\\
Japanese & ja & \ding{55} & GSD (543) & 2.0k & 1.1k & 533\\
Javanese & jv &  & CSUI (125) &  & & 1.1k\\
Georgian & ka &  &  & 2.0k & & 10.0k\\
Kazakh & kk &  & KTB (1.0k) & 1.9k & 1.0k & 4.1k\\
Khmer & km & \ding{55} &  & 1.9k & 2.4k & 12\\
Kannada & kn &  &  & 906 & & 10.0k\\
Korean & ko &  & Kaist (2.3k) & 2.0k & & 215\\
\bottomrule
\end{tabular}

\begin{tabular}{|llc|rrrr}
\toprule
Language & iso & Space & UD & OPUS100 & Ersatz & Speeches\\
\midrule
Kurdish & ku &  &  & 1.9k & & 10.0k\\
Kirghiz & ky &  &  & 1.7k & & 3.0k\\
Latin & la &  & ITTB (2.1k) &  & & 10.0k\\
Lithuanian & lt &  & ALKSNIS (684) & 2.0k & 1.0k & 2.1k\\
Latvian & lv &  & LVTB (2.3k) & 2.0k & 2.0k & 10.0k\\
Malagasy & mg &  &  & 2.0k & & 103\\
Macedonian & mk &  &  & 2.0k & & 10.0k\\
Malayalam & ml &  &  & 2.0k & & 2.1k\\
Mongolian & mn &  &  & 4.2k & & 5.5k\\
Marathi & mr &  & UFAL (47) & 2.0k & & 10.0k\\
Malay & ms &  &  & 1.9k & & 2.1k\\
Maltese & mt &  & MUDT (518) & 2.0k & & 10.0k\\
Burmese & my & \ding{55} &  & 2.0k & & 5.5k\\
Nepalese & ne &  &  & 1.9k & & 6.7k\\
Dutch & nl &  & Alpino (596) & 2.0k & & 10.0k\\
Norwegian & no &  & Bokmaal (1.9k) & 2.0k & & 2.1k\\
Panjabi & pa &  &  & 2.0k & & 3.8k\\
Polish & pl &  & PDB (2.2k) & 2.0k & 1.0k & 548\\
Pushto & ps &  &  & 1.8k & 2.7k & 10.0k\\
Portuguese & pt &  & Bosque (1.2k) & 2.0k & & 5.5k\\
Romanian & ro &  & Nonstandard (1.1k) & 2.0k & 2.0k & 10.0k\\
Russian & ru &  & Taiga (881) & 2.0k & 991 & 10.0k\\
Sinhala & si &  &  & 2.0k & & 516\\
Slovak & sk &  & SNK (1.1k) & 2.0k & & 10.0k\\
Slovenian & sl &  & SSJ (1.3k) & 2.0k & & 10.0k\\
Albanian & sq &  & TSA (60) & 2.0k & & 10.0k\\
Serbian & sr &  & SET (520) & 2.0k & & 5.6k\\
Swedish & sv &  & LinES (1.0k) & 2.0k & & 2.6k\\
Tamil & ta &  & TTB (120) & 2.0k & 1.0k & 3.0k\\
Telugu & te &  &  & 2.0k & & 303\\
Tajik & tg &  &  & 2.0k & & 10.0k\\
Thai & th &  & PUD (1.0k) & 2.0k & & 7.8k\\
Turkish & tr &  & IMST (983) & 2.0k & 3.0k & 3.8k\\
Ukrainian & uk &  & IU (892) & 2.0k & & 10.0k\\
Urdu & ur &  & UDTB (535) & 1.9k & & 7.8k\\
Uzbek & uz &  &  & 2.0k & & 3.8k\\
Vietnamese & vi &  & VTB (800) & 1.9k & & 10.0k\\
Xhosa & xh &  &  & 1.9k & & 5.6k\\
Yiddish & yi &  &  & 1.3k & & 2.6k\\
Yoruba & yo &  & YTB (318) & 9.4k & & 10.0k\\
Chinese & zh & \ding{55} & GSDSimp (500) & 2.0k & 2.0k & 1.7k\\
Zulu & zu &  &  & 1.9k & & 8.1k\\
& & & & &\\
\bottomrule
\end{tabular}
}
\caption{List of the 85 languages considered, whether they generally use whitespace to split sentences, and the corresponding evaluation dataset size, measured in sentences. For UD, we use UDv2.13, where the treebank name used is also shown. We use \textit{Speeches} only in pairwise evaluations.}
\label{tab:list_of_languages}
\end{table*}

    \begin{table*}[t]
\centering
\footnotesize
\begin{tabular}{lp{0.875\linewidth}}
\toprule
\multirow{12}{*}{\thead{English\\(\texttt{en})}} & we use science to create something wonderful\\
& we use story and artistic touch to get us to a place of wonder\\
& this guy wall-e is a great example of that\\
& he finds beauty in the simplest things\\
& but when he came in to lighting we knew we had a big problem\\
& we got so geeked-out on making wall-e this convincing robot that we made his binoculars practically optically \newline \hspace*{0.5cm}perfect\\
& laughter\\
& \textcolor{red!70!black}{\textbf{\textcolor{red!70!black}{\textbf{(*)}}}} his binoculars are one of the most critical acting devices he has\\
& he doesn't have a face or even traditional dialogue for that matter \\
& so the animators were heavily dependent on the binoculars to sell his acting and emotions \\
& we started lighting and we realized the triple lenses inside his binoculars were a mess of reflections \\
& he was starting to look glassy-eyed \\

\midrule

\multirow{15}{*}{\thead{German\\(\texttt{de})}} 
& aber ich habe auch das marfan-syndrom\\
& das ist eine erbkrankheit\\
& 1992 nahm ich an einer genetikstudie teil\\
& zu meinem entsetzen erfuhr ich dass wie sie hier sehen meine aorta ascendens nicht im normalbereich war die \newline \hspace*{0.5cm} grüne linie hier unten\\
& alle hier im raum werden bei 3,2 und 3,6 cm liegen\\
& ich war bereits bei 4,4\\
& wie sie sehen können erweiterte sich meine aorta zunehmend und allmählich geriet ich an den punkt dass eine \newline \hspace*{0.5cm} operation nötig sein würde\\
& die angebotene operation war ziemlich gruselig\\
& anästhesie öffnen des brustkorbs  man hängt sie an eine künstliche herzlungenmaschine lässt ihre \newline \hspace*{0.5cm} körpertemperatur auf etwa 18 grad fallen hält ihr herz an schneidet die aorta raus ersetzt sie mit einer klappe und aorta aus plastik\\
& und am wichtigsten verdonnert sie lebenslang zu antikoagulationstherapie\\
& \textcolor{red!70!black}{\textbf{\textcolor{red!70!black}{\textbf{(*)}}}} normalerweise mit warfarin\\
& der gedanke an diese operation war nicht gerade ansprechend\\
\midrule

\multirow{13}{*}{\thead{French\\(\texttt{fr})}} & 
j'ai montré mon intro et j'ai mis la scène de la méduse\\
& le réalisateur est resté silencieux pendant un très long moment \textcolor{red}{\textbf{(|)}} assez long pour que je puisse me dire oh non \newline \hspace*{0.5cm} c'est foutu\\
& et il a commencé à applaudir\\
& puis le concepteur artistique\\
& \textcolor{red!70!black}{\textbf{\textcolor{red!70!black}{\textbf{(*)}}}} et finalement toute la salle\\
& c'est pour ces moments que je fais ce travail\\
& le moment où tout fait sens et où l'on crée un monde auquel on peut croire\\
& on utilise la science et la programmation pour créer ces mondes incroyables\\
& on utilise les histoires et l'art pour leur donner vie\\
& c'est la coexistence de l'art et la science qui transforme le monde en un lieu magique un lieu avec une âme un \newline \hspace*{0.5cm} lieu auquel on peut croire un lieu où les choses qu'on imagine deviennent réelles -- et un monde où tout d'un \newline \hspace*{0.5cm} coup une fille réalise qu'elle n'est pas seulement une scientifique mais aussi une artiste\\
& merci \textcolor{red!70!black}{\textbf{\textcolor{red!70!black}{\textbf{(*)}}}} applaudissements\\
\midrule
\multirow{13}{*}{\thead{Italian\\(\texttt{it})}}
& cosa state leggendo \\
& beau lotto \\
& \textcolor{red!70!black}{\textbf{\textcolor{red!70!black}{\textbf{(*)}}}} cosa state leggendo \\
& mancano metà delle lettere \\
& giusto \\
& non c'è nessuna ragione a priori perché una h debba comparire tra la w e la a \\
& ma ne collocate una lì \\
& perché \\
& perché nella statistica della vostra esperienza passata sarebbe stato utile fare così \\
& quindi lo fate di nuovo \\
& e tuttavia non collocate una lettera dopo quella prima t \\
& perché \textcolor{red}{\textbf{(|)}} perché non si sarebbe dimostrato utile nel passato \\
& quindi non lo fate di nuovo\\
\bottomrule
\end{tabular}
\caption{Examples of predictions of \modeloursft taken from random positions from transcribed TED talks in four langauges.
\textcolor{red}{\textbf{(|)}} marks a missing sentence boundary (false negative), and \textcolor{red!70!black}{\textbf{\textcolor{red!70!black}{\textbf{(*)}}}} marks a wrongly inserted sentence boundary (false positive). All others are correctly segmented, according to the ground truth segmentation.
}
\label{tab:example.ted}
\end{table*}

\begin{table*}[t]
\centering
\footnotesize
\begin{tabular}{lp{0.84\linewidth}}
\toprule
\multirow{14}{*}{\thead{German-\\English\\(\texttt{de}-\texttt{en})}} 
& its about the rundfunkstaatsvertrag and the licence you need to stream to more than 500 viewers \\
& just go to the amt on your day off arrive at 9 and bring all the papers and a book \\
& hat ihr kein editor angekreidet \\
& guess op thought everyone pays freiwillige gesetzliche krankenversicherung und pflegeversicherung which \newline \hspace*{0.5cm} is around 780eurs month per person family depends \textcolor{red}{\textbf{(|)}} is the date that matters \\
& the bescheinigungszeitraum on my ausdruck der elektronischen lohnsteuerbescheinigung für 2013 or my \newline \hspace*{0.5cm} first anmeldung \\
& na anti-establishment anti-kapitalismus oder generell anti-zwang \\
& \textcolor{red!70!black}{\textbf{\textcolor{red!70!black}{\textbf{(*)}}}} zum beispiel \textcolor{red}{\textbf{(|)}} ich bin software engineer mit data-management data-security  background und er is \newline \hspace*{0.5cm} front-end mobile dev \\
& noch besser in flughäfen in england kommt keine höflich message wie achten sie bitte auf ihr eigenes \newline \hspace*{0.5cm} gepäck... sondern direkt eine drohung for security reasons baggage left unattended will be removed and \newline \hspace*{0.5cm} destroyed \\
& da bleibt irgendwie nicht mehr viel übrig \\

\midrule
\multirow{14}{*}{\thead{Spanish-\\English\\(\texttt{es}-\texttt{en})}}
& in the morning over there cada vez que yo decía algo él me decía algo\\
& the best thing about her ella no complain you know \textcolor{red}{\textbf{(|)}} tiene she has a great personality\\
& hasta que tú pushed the wrong button\\
& linda lópez la testing\\
& ay but she's cool\\
& el teniente y ella han tenido tú sabes conflicts\\
& entonces tina es the computer person\\
& so ella ella es la jefa de linda\\
& sí i i have a room\\
& no\\
& \textcolor{red!70!black}{\textbf{\textcolor{red!70!black}{\textbf{(*)}}}} pero tal vez consigue un roommate\\
& un roommate \textcolor{red}{\textbf{(|)}} mandó un e-mail diciendo que le había que había otra persona en la dirección en lugar de \newline \hspace*{0.5cm} ella\\
& yo me dio a entender según como leí yo que era ella e edith\\

\midrule
\multirow{12}{*}{\thead{Turkish-\\German\\(\texttt{tr}-\texttt{de})}}
& ja bence ich probiere es einfach\\
& ja vor allem sınav olduğu zaman muss ja muss ja schon so sein\\
& geçen sene ich weiß noch eh sınavların olduğu günlerde \textcolor{red}{\textbf{(|)}} ja tam denk geldi weißt du \textcolor{red}{\textbf{(|)}} die woche noch \newline \hspace*{0.5cm} ich denke mir so tutayım mı tutmayayım mı kann gar nicht mehr\\
& ben tutmuştum bir tanesinde\\
& \textcolor{red!70!black}{\textbf{(*)}} und ich dachte so tamam bittin sen die\\
& \textcolor{red!70!black}{\textbf{(*)}} hani das ist nicht gut gegangen\\
& \textcolor{red!70!black}{\textbf{(*)}} die prüfung das war auch so\\
& \textcolor{red!70!black}{\textbf{(*)}} ich konnte mich gar konzentrieren überhaupt nicht\\
& hani yemek de değil\\
& \textcolor{red!70!black}{\textbf{(*)}} weißt du einfach nur wasser\\
& ja bir de o geçen sene da war es so heiß\\
\midrule

\multirow{13}{*}{\thead{Vietnamese-\\English\\(\texttt{vi}-\texttt{en})}} 
& \vietnamese{bởi vậy lux đường có overconfident \textcolor{red}{\textbf{(|)}} ai cũng cần phải improve \textcolor{red}{\textbf{(|)}} nên lux phải phải phải đưa cho chị ti với \newline \hspace*{0.5cm} anh alex check nha \textcolor{red}{\textbf{(|)}} thì design đến đâu rồi lux \textcolor{red}{\textbf{(|)}} cái graphic design của lux}\\
& \vietnamese{\textcolor{red!70!black}{\textbf{(*)}} \vietnamese{lux design được đến đâu rồi come up with idea để ghi ra}}\\
& \vietnamese{\textcolor{red!70!black}{\textbf{(*)}} chị ti sẽ cùng help lux to write down}\\
& \vietnamese{\textcolor{red!70!black}{\textbf{(*)}} hoặc là ở woden qua đây ăn dinner với chị ti}\\
& \vietnamese{get it out of the way là done với một cái đó rồi là xong}\\
& \vietnamese{thái writing cũng đâu có good đâu}\\
& \vietnamese{deadline như vậy được chưa}\\
& \vietnamese{vậy là tối mai lux biết spend time write it tomorrow \textcolor{red}{\textbf{(|)}} được rồi ta design như vậy nè}\\
& \vietnamese{mọi thứ là lux phải plan ahead như vậy chứ}\\
\bottomrule
\end{tabular}
\caption{Examples of predictions of \modeloursft taken from random positions from code-switching text in four language pairs.
\textcolor{red}{\textbf{(|)}} marks a missing sentence boundary (false negative), and \textcolor{red!70!black}{\textbf{\textcolor{red!70!black}{\textbf{(*)}}}} marks a wrongly inserted sentence boundary (false positive). All others are correctly segmented according to the ground truth segmentation.
There are many ambiguous sentence boundaries in these corpora.
}
\label{tab:example.code-switching}
\end{table*}
\begin{table*}[t]
\centering
\footnotesize
\begin{tabular}{p{0.52\linewidth}p{0.40\linewidth}}
\toprule
\textbf{Original (non-corrupted) song} & \textbf{Corrupted song} \\
\midrule
\begin{tabular}{@{}p{\linewidth}@{}}
\small
Have yourself a merry little Christmas\\
Let your hearts be light\\
From now on\\
Our troubles will be out of sight\\
\\
Have yourself a merry little Christmas\\
Make the Yule-tide gay\\
From now on\\
Our troubles will be miles away.\\
\\
Here we are as in olden days\\
Happy golden days of yore\\
Faithful friends who are dear to us\\
Will be near to us once more\\
\\
Through the years We all will be together\\
If the Fates allow\\
Until then we'll have to muttle through somehow\\
So have yourself a merry little Christmas now\\
\\
Here we are as in olden days\\
Happy golden days of yore\\
Faithful friends who are dear to us\\
Will be near to us once more\\
\\
Through the years We all will be together\\
If the Fates allow\\
Until then we'll have to muttle through somehow\\
So have yourself a merry little Christmas now\\
\\
Have yourself a merry little Christmas now\\
Merry Christmas
\end{tabular} &
\begin{tabular}{@{}p{\linewidth}@{}}
\small
i'd never leave the perfect girl\\
or rip apart the perfect world\\
just up and leave in the middle of a song\\
\\
i'd never pack my things in a silverado
\\drive on out to colorado\\
just to find some freedom i thought was gone\\
\textcolor{red}{\textbf{(|)}} 
ooooh\\
there are things i'd never do\\
\\
but here i am in this hotel room\\
thinking bout you and what i 've done\\
oh what have i done\\
head in my hands\\
thinking about a lot of things\\
i wish that i could change\\
it's sad but it's true\\
i 've done a lot of things i'd never do\\
\\
i'd never ever work so much\\
that i'd lose sight i'd lose touch\\
of everything a man could ever want\\
\\
i'd never lose my cool and say\\
those words that cut just like a blade\\
and leave you dying crying all alone\\
\\
i'd never leave the perfect girl\\
or rip apart the perfect world\\
just up and leave in the middle of a song

\end{tabular} \\
\bottomrule
\end{tabular}
\caption{Examples of predictions of \modelourslora taken from songs categorized as \emph{Country} (\emph{High Repetitiveness}).\newline
\textcolor{red}{\textbf{(|)}} marks a missing verse boundary (false negative), and \textcolor{red!70!black}{\textbf{\textcolor{red!70!black}{\textbf{(*)}}}} marks a wrongly inserted verse boundary (false positive).
All others are correctly segmented according to the ground truth segmentation.
While the task was only to segment songs into \emph{verses} and no line segmentation was provided, we format songs using both lines and verses for clarity.
}
\label{tab:example.lyrics-high}
\end{table*}

\begin{table*}[t]
\centering
\footnotesize
\begin{tabular}{p{0.52\linewidth}p{0.40\linewidth}}
\toprule
\textbf{Original (non-corrupted) song} & \textbf{Corrupted song} \\
\midrule
\begin{tabular}{@{}p{\linewidth}@{}}
\small
Hope, a new beginning\\
Time, time to start living\\
Just like just before we died\\
\\
There's no going back to the place we started from\\
\\
Hurt, falling through fingers\\
Trust, trust in the feeling\\
There's something left inside\\
\\
There's no going back to the place we started from\\
All secrets known\\
\\
Calm, old wounds are healing\\
Strong, truth is worth saving\\
I want to feel alive\\
\\
There's no going back to the place we started from\\
All secrets known\\
\end{tabular} &
\begin{tabular}{@{}p{\linewidth}@{}}
\small
 lock me up inside my room\\
 leave me without toys and food\\
 keep that monster in my bed\\
 just remember i'm not dead\\
 \\
 you forget my memory lives way beyond these walls\\
 you forget my indecision's taking all control\\
 \\
 once in a far land i grabbed you\\
 and you woke me up to my origin\\
 \\
 people have to understand my innocence has gone\\
\textcolor{red}{\textbf{(|)}} 
 go beyond my urge\\
 or make an effort to\\
 living on my own plagued by images\\
 \\
 once in a far land i grabbed you\\
 and you woke me up to my origin \\
\end{tabular} \\
\bottomrule
\end{tabular}
\caption{
Examples of predictions of \modelourslora from songs categorized as \emph{Alternative Metal} (\emph{Mid Repetitiveness}). 
}
\label{tab:example.lyrics-mid}
\end{table*}

\begin{table*}[t]
\centering
\footnotesize
\begin{tabular}{p{0.52\linewidth}p{0.40\linewidth}}
\toprule
\textbf{Original (non-corrupted) song} & \textbf{Corrupted song} \\
\midrule
\begin{tabular}{@{}p{\linewidth}@{}}
\small
Let me chirp these fools\\
\\
Juice got weed Juice got pills\\
Juice got the work on the corner cutting deals\\
Juice know you haters out there snitching ain't for real\\
So Juice got some gang niggas down for the kill\\
Juice know the feds got surveillance on the field\\
We never had a job but we sitting on a mill\\
We ball out in the club wit our niggas staying trill\\
We never wrote a check just them big face bills\\
A player drinking Makers Marka, cranberry vodka\\
Wearing a mink coat thats furry as Chewbacca\\
I saw ya main girl and a player had to stop her\\
Her name wasn't Silkk but her face was The Shocker\\
The feds taking pictures of us balling but I got 'em\\
A 7 footer hole for his body we gonna drop 'em\\
We always on the grind we be watching when they watching\\
And when they turn they back its the clucka-clucka-rock 'em yeah!\\
\\
If you boys got beef we can (roll wit it)\\
In the club or the street we can (go wit it)\\
It don't make me none (blow for blow wit it)\\
Crack his head wit a gun (I'ma sho split it)\\
\\
We got them tones in the club and them bulletproof vests\\
Them three fifty seven titanium Smith-N-Wess\\
And plus we deep as hell and prepared to bust\\
You gonna have hell if you fuck wit us and thats whats up\\
\\
\textcolor{red!70!black}{\textbf{\textcolor{red!70!black}{\textbf{(*)}}}}
The whole club we maintain\\
These hydrashock bullets mushroom in ya brain\\
We in bed with the med we give 'em something to do\\
Cause clown ass niggas love to act a fool\\
\\
My hood is real nigga my hood ain't fake\\
My hood is home nigga everything straight\\
My hood will rob you with mask on they face\\
My hood will do it to put food on they plate\\
My hood ain't tame dog they wanna jump fool\\
My hood they hang together they all jump you\\
And if you don't believe me then come to my hood\\
And you will see that it ain't all good\\
\end{tabular} &
\begin{tabular}{@{}p{\linewidth}@{}}
\small
zaytoven on the track\\
\textcolor{red}{\textbf{(|)}} 
zay-tiggy\\
gucci\\
gucci\\
\\
so watch entertainment\\
lets go\\
\\
\\
they call me chef-boy-r.g.\\
but hold that thought\\
its a kodak moment\\
but hold that thought\\
hurricane wrist game\\
turn that junk off\\
hot as piggly wiggly\\
cant kermit the frog dog\\
\\
\\
early in the mornin\\
i aint even yawnin\\
cookin up a cake\\
like i'm doin a performance\\
when it come to flossin\\
i aint even talkin\\
diamonds on my joint\\
got my chevy moonwalkin\\
10 bricks on my bart simpson just look\\
my watch 35 pounds of kush\\
my ring 36 oz's my nig\\
my bracelet 500 lbs of mid\\
a gucci wrapped tour bus\\
yall hoes follow us\\
party pack pills man\\
hoes gonna swallow us\\
naturally a loner\\
but love my kid\\
mix the soda with the cola\\
i can buy me a friend\\
new swag somethin like\\
trap house times 10\\
ery nigga round me \\
bust heads, ya-dig\\
iced out grill \\
i can't buy that bullshit\\
i'm wit some street shit,\\
like a reverend in the pulpit\\
\\
they call me chef-boy-r.g.\\
but hold that thought\\
its a kodak moment\\
but hold that thought\\
hurricane wrist game\\
turn that junk off\\
hot as piggly wiggly\\
cant kermit the frog dog\\
...\\
\end{tabular} \\
\bottomrule
\end{tabular}
\caption{
Examples of predictions of \modelourslora from songs categorized as \emph{Southern Hip Hop} (\emph{Low Repetitiveness}). 
}
\label{tab:example.lyrics-low}
\end{table*}

    \begin{table*}[h]
\centering
\resizebox{0.85\textwidth}{!}{
    \begin{tabular}{lccccccccccc} 
        \toprule
        \multirow{2}{*}{\textbf{Model}} & \multicolumn{2}{c}{\texttt{fr}} & \multicolumn{2}{c}{\texttt{es}} & \multicolumn{2}{c}{\texttt{it}} & \multicolumn{1}{c}{\texttt{en}} & \multicolumn{2}{c}{\texttt{de}} & \multirow{2}{*}{\textbf{\thead{Macro\\Avg.}}} \\
        \cmidrule{2-10}
        & Judg. & Laws & Judg. & Laws & Judg. & Laws & Judg. & Judg. & Laws  \\
        \midrule
        \modelspacydpm & 82.7 & 61.2 & 67.7 & 85.2 & 73.6 & 53.8 & 81.4 & 65.8 & 67.9 & 71.0 \\
        \modelersatz & 81.5 & 51.7 & 63.6 & 81.9 & - & - & 79.5 & 59.8 & 68.5 & 69.5 \\
        \midrule
        \modelllama & 85.5 & 52.7 & 70.4 & 66.5 & 81.1 & 77.2 & 89.2 & 78.3 & 83.4 & 76.0 \\
        \modelcommandr & 62.5 & 51.2 & 55.5 & 52.8 & 56.2 & 70.3 & 69.0 & 55.7 & 64.3 & 59.7 \\
        \midrule
        \modelours & 82.7 & 86.8 & 68.8 & 73.2 & 85.1 & 71.2 & 82.6 & 67.8 & 86.4 & 78.3 \\
        \modeloursft & \textbf{85.1} & \textbf{95.8} & \textbf{80.1} & \textbf{89.3} & \textbf{88.3} & \textbf{80.6} & \textbf{93.1} & \textbf{81.3} & \textbf{92.7} & \textbf{87.4} \\
        \bottomrule
        \modelpunkt & 75.6 & 51.5 & 65.2 & 87.7 & 72.9 & 45.3 & 76.3 & 65.1 & 73.3 & 68.1 \\
        \modelpysbd & 74.2 & 50.5 & 60.8 & 79.7 & 74.1 & 55.0 & 75.0 & 67.6 & 70.4 & 67.5 \\
        \modelspacydp & 71.6 & 74.3 & 64.9 & 87.3 & 74.0 & 54.4 & 84.5 & 68.2 & 65.2 & 71.6 \\
        \modelwtpu & 87.8 & 63.0 & 75.6 & 84.6 & 84.3 & 80.0 & 88.8 & 79.7 & 85.8 & 81.1 \\
        \modelwtpt & 87.0 & 80.7 & 76.3 & 87.1 & 85.0 & 79.5 & 88.7 & 80.4 & 85.9 & 83.4 \\
        \modelwtppunct & \textbf{96.8} & 98.4 & 88.7 & 94.6 & 94.0 & 96.9 & 96.3 & 87.3 & 93.7 & 94.1 \\
        \midrule
        \modelsbdmonospecific & 96.4 & 97.7 & 88.7 & 93.8 & 93.7 & 98.0 & 95.3 & 86.9 & 93.5 & 93.8 \\
        \modelsbdmonoboth & 96.5 & 98.5 & 88.2 & 93.7 & 87.1 & 84.4 & 95.3 & 87.6 & 92.7 & 91.6 \\
        \modelsbdmultispecific & 96.4 & \textbf{98.9} & 89.0 & \textbf{94.8} & 94.5 & 98.1 & 95.7 & 87.5 & 93.6 & 94.3 \\
        \modelsbdmultiboth & 96.6 & \textbf{98.9} & 88.7 & \textbf{94.8} & 94.6 & 98.2 & 95.6 & 78.6 & 93.2 & 93.2 \\
        \midrule
        \modelourslora\hspace{-0.75ex}@\hspace{-0.5ex}10 & 95.1 & 96.6 & 86.5 & 93.5 & 94.0 & 85.6 & 96.3 & 87.7 & \textbf{97.3} & 92.5 \\
        \modelourslora\hspace{-0.75ex}@\hspace{-0.5ex}100 & 96.7 & 97.0 & \textbf{89.3} & 94.0 & 95.4 & 87.1 & 97.0 & 88.8 & \textbf{97.3} & 93.6 \\
        \modelourslora & \textbf{96.8} & 98.5 & 89.1 & 94.4 & \textbf{95.5} & \textbf{98.3} & \textbf{97.3} & \textbf{88.9} & 97.1 & \textbf{95.1} \\
        \bottomrule
    \end{tabular}}
    \caption{Sentence segmentation F1 score for legal data (MultiLegalSBD). We take the macro F1 scores over documents within a given category. @N correspond to using a maximum of N documents per category for adaptation, respectively. Transformer-based \modelsbd baselines are taken from \citet{Brugger2023MultiLegalSBDAM}. For these domain-specific baselines, \textit{mono} and \textit{multi} correspond to models trained on documents from only one or all languages, respectively. \textit{Both} corresponds to models trained on both laws and judgments, whereas \textit{specific} corresponds to models trained on a given category (laws/judgments).
    Best per-category results are in \textbf{bold}.
    }
    \label{tab:all-legal-clean}
\end{table*}


\begin{table*}[h]
\centering
\resizebox{0.85\textwidth}{!}{
    \begin{tabular}{lcccccccccc} 
        \toprule
        \multirow{2}{*}{\textbf{Model}} & \multicolumn{2}{c}{\texttt{fr}} & \multicolumn{2}{c}{\texttt{es}} & \multicolumn{2}{c}{\texttt{it}} & \multicolumn{1}{c}{\texttt{en}} & \multicolumn{2}{c}{\texttt{de}} & \multirow{2}{*}{\textbf{\thead{Macro\\Avg.}}} \\
        \cmidrule{2-10}
        & Judg. & Laws & Judg. & Laws & Judg. & Laws & Judg. & Judg. & Laws &  \\
        \midrule
        \modelspacydpm & 0.0 & 0.0 & 1.5 & 0.0 & 0.4 & 0.0 & 0.2 & 2.2 & 0.0 & 0.5 \\
        \modelersatz & 0.7 & 0.1 & 4.7 & 0.0 & - & - & 0.6 & 4.1 & - & 1.7 \\
        \midrule
        \modelllama & 60.0 & 40.7 & 50.5 & 55.9 & 54.2 & 35.6 & 69.2 & 59.2 & 74.2 & 55.5 \\
        \modelcommandr & 54.3 & 57.7 & 42.8 & 44.5 & 46.4 & 42.9 & 56.0 & 45.4 & 64.1 & 50.5 \\
        \midrule
        \modelours & 63.7 & 80.6 & 54.9 & 63.3 & 56.0 & 55.9 & 49.9 & 57.6 & 75.0 & 61.9 \\
        \modeloursft & \textbf{68.8} & \textbf{89.5} & \textbf{65.3} & \textbf{83.3} & \textbf{62.3} & \textbf{71.9} & \textbf{74.5} & \textbf{72.0} & \textbf{80.9} & \textbf{74.3} \\
        \bottomrule
        \modelpunkt & 1.4 & 0.0 & 3.8 & 0.0 & 1.3 & 1.6 & 0.2 & 4.9 & 0.0 & 1.5 \\
        \modelpysbd & 21.7 & 0.2 & 3.8 & 0.0 & 20.0 & 0.2 & 1.3 & 0.4 & 0.0 & 5.3 \\
        \modelspacydp & 7.9 & 4.4 & 4.6 & 0.0 & 5.3 & 2.3 & 2.4 & 13.1 & 8.8 & 5.4 \\
        \modelwtpu & 32.2 & 19.5 & 38.0 & 41.1 & 36.9 & 28.8 & 28.1 & 25.0 & 19.0 & 29.8 \\
        \modelwtpt & 48.9 & 64.7 & 51.8 & 71.5 & 46.5 & 43.8 & 54.5 & 52.4 & 62.0 & 55.1 \\
        \modelwtppunct & 65.0 & 83.4 & 67.2 & 83.1 & 58.3 & 66.9 & 73.5 & 73.0 & 84.1 & 72.7 \\
        \midrule
        \modelsbdmonospecific & 9.6 & 45.1 & 7.9 & 13.2 & 6.9 & 47.1 & 3.6 & 12.1 & 0.3 & 16.2 \\
        \modelsbdmonoboth & 9.0 & 43.0 & 8.1 & 15.6 & 6.7 & 39.6 & 3.6 & 7.7 & 0.6 & 14.9 \\
        \modelsbdmultispecific & 8.8 & 44.4 & 7.3 & 34.4 & 6.9 & 47.2 & 2.5 & 9.2 & 0.7 & 17.9 \\
        \modelsbdmultiboth & 8.9 & 43.5 & 5.6 & 24.6 & 6.9 & 47.4 & 1.3 & 2.4 & 0.4 & 15.7 \\
        \midrule
        \modelourslora\hspace{-0.75ex}@\hspace{-0.5ex}10 & 70.8 & 82.9 & 67.8 & 79.9 & 63.5 & 62.6 & 80.6 & 78.6 & \textbf{93.5} & 75.6 \\
        \modelourslora\hspace{-0.75ex}@\hspace{-0.5ex}100 & 76.9 & 86.6 & 75.9 & 84.1 & 69.3 & 75.3 & \textbf{84.2} & \textbf{84.1} & \textbf{93.5} & 81.1 \\
        \modelourslora & \textbf{77.5} & \textbf{90.2} & \textbf{76.1} & \textbf{85.5} & \textbf{71.2} & \textbf{87.5} & \textbf{84.2} & 83.8 & \textbf{93.5} & \textbf{83.3} \\
        \bottomrule
    \end{tabular}}
    \caption{Sentence segmentation F1 score for corrupted legal data (MultiLegalSBD), where we \emph{remove all casing and punctuation tokens}. We take the macro F1 scores over documents within a given category.
    }
    \label{tab:all-legal-corrupted}
\end{table*}

    \begin{table*}[h]
   \centering
   \small
   \scalebox{1}{
        \begin{tabular}{lccc|cc}
        \toprule
        & \multicolumn{5}{c}{Repetitiveness} \\
        \cline{2-6}
        & \multicolumn{3}{c}{High} & \multicolumn{2}{c}{Low} \\
        \cline{2-4} \cline{5-6}
        \textbf{Model} & Country & \thead{Punk\\Rock} & \thead{Pop\\Punk} & \thead{Southern\\Hip Hop} & \thead{Gangsta\\Rap} \\
        \toprule
        $\text{SSM}_{\text{string}}$$^{\dagger}$ &- &- & -&- &- \\
        \modelllama & 47.0 & 50.2 & 49.5 & 34.7 & 32.8 \\
        \modelcommandr & 35.3 & 39.7 & 39.2 & 27.5 & 29.9 \\
        \modelwtppunct\hspace{-0.75ex}@\hspace{-0.5ex}100 & 56.5 & 56.1 & 54.9 & 43.3 & 42.4 \\
        \modelwtppunct\hspace{-0.75ex}@\hspace{-0.5ex}1000 & 58.9 & 58.8 & 57.5 & 45.9 & 43.9 \\
        \modelwtppunct & 58.9 & 58.8 & 57.5 & 45.9 & 43.9 \\
        \midrule
        \modelourslora\hspace{-0.75ex}@\hspace{-0.5ex}100 & 66.7 & 67.5 & 69.2 & 53.0 & 50.3 \\
        \modelourslora\hspace{-0.75ex}@\hspace{-0.5ex}1000 & 76.7 & 76.0 & 76.8 & \textbf{64.5} & \textbf{60.9} \\
        \modelourslora & \textbf{79.3} & \textbf{76.8} & \textbf{77.6} & 64.2 & 60.3 \\
        \bottomrule
        \end{tabular}
    } 
    \caption{Complete verse segmentation F1 scores for songs categorized as \emph{low} and \emph{high} repetitiveness. Categorization of songs into genres and repetitiveness taken from \citet{fell-etal-2018-lyrics}. We report the macro average over songs.  $^{\dagger}$Values for $\text{SSM}_{\text{string}}$ taken from \citet{fell-etal-2018-lyrics}, with lyrics already pre-segmented into lines. \hspace{-0.75ex}@\hspace{-0.5ex}N corresponds to using a maximum of N and 1000 songs per genre for adaptation, respectively.}
    \label{tab:high-low-lyrics-clean}
\end{table*}

\begin{table*}[h]
   \centering
   \small
   \scalebox{1}{
        \begin{tabular}{lccccccccc}
        \toprule
        & \multicolumn{9}{c}{Mid Repetitiveness} \\
        \cline{2-10}
        \textbf{Model} & \thead{\small Rock} & \thead{\small Pop} & \thead{\small RnB} & \thead{\small Soul} & \thead{\small Alt.\\\small Rock} & \thead{\small Alt.\\\small Metal} & \thead{\small Indie\\\small Rock} & \thead{\small Pop\\\small Rock} & \thead{\small Hard\\\small Rock} \\
        \toprule
        $\text{SSM}_{\text{string}}$$^{\dagger}$ & 64.8 & 66.6 & 65.6 & 63.0 & 67.9 & 68.5 & 65.6 & 65.8 & 67.7 \\
        \modelllama & 48.2 & 47.0 & 45.7 & 48.7 & 49.3 & 47.6 & 48.4 & 47.1 & 48.7 \\
        \modelcommandr & 37.8 & 36.3 & 33.3 & 36.5 & 40.1 & 39.1 & 40.2 & 37.6 & 38.4 \\
        \modelwtppunct\hspace{-0.75ex}@\hspace{-0.5ex}100 & 57.5 & 55.9 & 51.4 & 52.5 & 59.9 & 58.6 & 56.3 & 57.5 & 57.0 \\
        \modelwtppunct\hspace{-0.75ex}@\hspace{-0.5ex}1000 & 60.5 & 57.7 & 53.1 & 55.0 & 63.0 & 60.5 & 59.8 & 58.1 & 59.4 \\
        \modelwtppunct & 60.7 & 58.2 & 53.5 & 55.0 & 63.3 & 60.5 & 60.1 & 58.3 & 59.5 \\
        \midrule
        \modelourslora\hspace{-0.75ex}@\hspace{-0.5ex}100 & 65.4 & 63.8 & 61.5 & 59.8 & 64.9 & 68.6 & 63.9 & 63.8 & 64.1 \\
        \modelourslora\hspace{-0.75ex}@\hspace{-0.5ex}1000 & 74.7 & 72.7 & 71.2 & 71.1 & 75.3 & \textbf{77.7} & 72.6 & 74.5 & 75.7 \\
        \modelourslora & \textbf{78.1} & \textbf{75.6} & \textbf{73.4} & \textbf{71.7} & \textbf{77.4} & \textbf{77.7} & \textbf{73.5} & \textbf{75.6} & \textbf{76.6} \\
        \bottomrule
        \end{tabular}
    } 
    \caption{Complete verse segmentation F1 scores for songs categorized as \emph{mid} repetitiveness.
    }
    \label{tab:mid-lyrics-clean}
\end{table*}

\begin{table*}[h]
   \centering
   \small
   \scalebox{1}{
        \begin{tabular}{lccc|cc}
        \toprule
        & \multicolumn{5}{c}{Repetitiveness} \\
        \cline{2-6}
        & \multicolumn{3}{c}{High} & \multicolumn{2}{c}{Low} \\
        \cline{2-4} \cline{5-6}
        \textbf{Model} & Country & \thead{Punk\\Rock} & \thead{Pop\\Punk} & \thead{Southern\\Hip Hop} & \thead{Gangsta\\Rap} \\
        \toprule
        $\text{SSM}_{\text{string}}$$^{\dagger}$ & 70.2 & 70.9 & 72.7 & 47.0 & 47.7 \\
        \modelllama & 54.6 & 53.5 & 57.1 & 36.6 & 36.2 \\
        \modelcommandr & 38.8 & 42.6 & 41.4 & 35.0 & 26.3 \\
        \modelwtppunct\hspace{-0.75ex}@\hspace{-0.5ex}100 & 48.9 & 52.5 & 50.0 & 35.5 & 36.4 \\
        \modelwtppunct\hspace{-0.75ex}@\hspace{-0.5ex}1000 & 51.7 & 55.6 & 53.1 & 36.5 & 39.5 \\
        \modelwtppunct & 51.9 & 55.6 & 53.1 & 36.5 & 39.5 \\
        \midrule
        \modelourslora\hspace{-0.75ex}@\hspace{-0.5ex}100 & 66.7 & 67.2 & 67.7 & 44.9 & 48.4 \\
        \modelourslora\hspace{-0.75ex}@\hspace{-0.5ex}1000 & 72.7 & \textbf{71.8} & 74.5 & 54.7 & \textbf{55.8} \\
        \modelourslora & \textbf{75.2} & 71.4 & \textbf{74.7} & \textbf{54.8} & \textbf{55.8} \\
        \bottomrule
        \end{tabular}
    } 
    \caption{Complete verse segmentation F1 scores for songs categorized as \emph{low} and \emph{high} repetitiveness, where we \emph{remove all casing and punctuation tokens}. We report the macro average over songs.
    }
    \label{tab:high-low-lyrics-corrupted}
\end{table*}

\begin{table*}[h]
   \centering
   \small
   \scalebox{1}{
        \begin{tabular}{lccccccccc}
        \toprule
        & \multicolumn{9}{c}{Mid Repetitiveness} \\
        \cline{2-10}
        \textbf{Model} & \thead{\small Rock} & \thead{\small Pop} & \thead{\small RnB} & \thead{\small Soul} & \thead{\small Alt.\\\small Rock} & \thead{\small Alt.\\\small Metal} & \thead{\small Indie\\\small Rock} & \thead{\small Pop\\\small Rock} & \thead{\small Hard\\\small Rock} \\
        \toprule
        $\text{SSM}_{\text{string}}$$^{\dagger}$ &- &- &- &- &- &- &- &- &- \\
        \modelllama & 53.8 & 51.7 & 47.4 & 51.9 & 54.1 & 52.3 & 54.0 & 52.8 & 52.2 \\
        \modelcommandr & 41.1 & 37.5 & 34.0 & 40.0 & 41.4 & 40.7 & 42.6 & 39.5 & 40.6 \\
        \modelwtppunct\hspace{-0.75ex}@\hspace{-0.5ex}100 & 51.5 & 47.0 & 41.5 & 47.3 & 52.6 & 52.7 & 52.4 & 48.8 & 50.4 \\
        \modelwtppunct\hspace{-0.75ex}@\hspace{-0.5ex}1000 & 52.8 & 48.9 & 43.1 & 48.6 & 54.9 & 54.2 & 55.1 & 51.2 & 52.0 \\
        \modelwtppunct & 53.1 & 49.4 & 43.2 & 48.6 & 55.0 & 54.2 & 54.8 & 51.0 & 52.6 \\
        \midrule
        \modelourslora\hspace{-0.75ex}@\hspace{-0.5ex}100 & 64.5 & 62.2 & 58.8 & 60.8 & 67.4 & 66.9 & 62.7 & 63.5 & 63.6 \\
        \modelourslora\hspace{-0.75ex}@\hspace{-0.5ex}1000 & 70.4 & 68.5 & 65.2 & \textbf{65.6} & 71.4 & 73.1 & \textbf{69.9} & 68.3 & 71.4 \\
        \modelourslora & \textbf{73.2} & \textbf{71.5} & \textbf{67.1} & 65.1 & \textbf{73.3} & \textbf{73.5} & 69.5 & \textbf{70.4} & \textbf{72.7} \\
        \bottomrule
        \end{tabular}
    } 
    \caption{Complete verse segmentation F1 scores for songs categorized as \emph{mid} repetitiveness, where we \emph{remove all casing and punctuation tokens}. We report the macro average over songs.
    }
    \label{tab:mid-lyrics-corrupted}
\end{table*}
    \begin{table*}[t]
   \centering
   \small
   \scalebox{0.825}{
        \begin{tabular}{llcccccccccccccccc}
        \toprule
        & \textbf{Model} & \texttt{af} & \texttt{am} & \texttt{ar} & \texttt{az} & \texttt{be} & \texttt{bg} & \texttt{bn} & \texttt{ca} & \texttt{ceb} & \texttt{cs} & \texttt{cy} & \texttt{da} & \texttt{de} & \texttt{el} & \texttt{en} \\
        \midrule
        \multirow{13}{*}{UD} &
        \modelspacydpm & 98.3 & - & 79.1 & - & 80.0 & 93.8 & 47.9 & 98.8 & 98.5 & 89.6 & 98.9 & 94.9 & 87.7 & 93.8 & 91.7 \\
        & \modelersatz & - & - & 79.7 & - & - & - & - & - & - & 89.3 & - & - & 92.4 & - & 89.1 \\
        & \modelllama & \underline{\textbf{100.0}} & - & 80.1 & - & 88.1 & 97.3 & \underline{96.9} & 99.4 & \underline{\textbf{99.7}} & 93.0 & 98.7 & 95.2 & 96.7 & 94.7 & 94.8 \\
        & \modelcommandr & 86.1 & - & 75.6 & - & 68.1 & 76.9 & 59.2 & 76.6 & 89.9 & 68.9 & 80.8 & 73.8 & 85.3 & 66.7 & 77.1 \\
        & \modelours & 99.1 & - & 82.9 & - & 89.0 & 97.9 & \underline{96.2} & 98.9 & 99.7 & 91.5 & \underline{99.1} & 95.5 & 96.4 & \underline{97.5} & 93.9 \\
        & \modeloursft & \underline{\textbf{100.0}} & - & \underline{\textbf{84.5}} & - & \underline{\textbf{93.5}} & \underline{\textbf{99.3}} & \underline{\textbf{100.0}} & \underline{\textbf{99.7}} & \underline{\textbf{99.7}} & \underline{\textbf{94.3}} & \underline{\textbf{99.4}} & \underline{\textbf{98.5}} & \underline{\textbf{97.8}} & \underline{\textbf{97.9}} & \underline{\textbf{96.7}} \\
        \cmidrule{2-17}
        & \modelpunkt & - & - & - & - & - & - & - & - & - & 89.1 & - & 94.4 & 92.6 & 92.7 & 90.8 \\
        & \modelpysbd & - & - & 28.1 & - & - & 74.9 & - & - & - & - & - & 72.6 & 80.0 & 91.0 & 75.3 \\
        & \modelspacydp & - & - & - & - & - & - & - & \underline{99.8} & - & - & - & 94.0 & \underline{96.7} & 94.0 & 91.3 \\
        & \modelwtpu & 98.3 & - & 80.5 & - & 88.9 & 98.2 & \underline{\textbf{93.5}} & 98.3 & \underline{\textbf{99.7}} & \underline{92.3} & 99.2 & 95.1 & \underline{95.6} & 97.3 & 94.5 \\
        & \modelwtpt & 99.0 & - & \underline{86.4} & - & 88.8 & 98.1 & - & 98.4 & - & 92.0 & 99.2 & 94.3 & \underline{95.8} & 97.7 & 94.5 \\
        & \modelwtppunct & \underline{99.9} & - & \underline{\textbf{87.4}} & - & \underline{\textbf{91.9}} & \underline{\textbf{99.6}} & - & 99.6 & - & \underline{\textbf{95.4}} & \underline{99.5} & \underline{\textbf{98.9}} & \underline{96.5} & 97.8 & \underline{\textbf{96.9}} \\
        & \modelourslora & \underline{\textbf{100.0}} & - & \underline{86.6} & - & \underline{91.2} & \underline{99.4} & - & \underline{\textbf{99.9}} & - & \underline{95.2} & \underline{\textbf{99.6}} & \underline{98.7} & \underline{\textbf{96.8}} & \underline{\textbf{98.9}} & \underline{96.8} \\
        \midrule
        \multirow{13}{*}{OPUS100} & 
        \modelspacydpm & 41.9 & 6.3 & 57.9 & 72.3 & 33.9 & 93.4 & 37.2 & 88.0 & - & 87.3 & 25.8 & 90.2 & 72.9 & 89.2 & 88.8 \\
        & \modelersatz & - & - & 59.2 & - & - & - & - & - & - & 86.5 & - & - & 73.1 & - & 87.6 \\
        & \modelllama & 65.4 & 36.2 & 62.2 & 76.8 & 54.5 & 94.6 & 80.0 & 90.9 & - & 91.3 & 46.3 & 93.2 & 83.8 & 94.3 & 92.8 \\
        & \modelcommandr & 55.8 & 6.6 & 44.5 & 60.5 & 36.1 & 57.9 & 4.8 & 66.0 & - & 69.0 & 39.8 & 75.4 & 76.2 & 65.1 & 89.5 \\
        & \modelours & 78.4 & 58.0 & \underline{\textbf{67.0}} & 75.0 & 70.4 & 93.5 & 80.5 & 87.7 & - & 89.2 & 72.0 & 90.9 & 78.2 & 92.0 & 90.4 \\
        & \modeloursft & \underline{\textbf{86.2}} & \underline{\textbf{70.8}} & 65.2 & \underline{\textbf{85.3}} & \underline{\textbf{87.8}} & \underline{\textbf{96.2}} & \underline{\textbf{86.1}} & \underline{\textbf{93.0}} & - & \underline{\textbf{94.3}} & \underline{\textbf{79.6}} & \underline{\textbf{94.0}} & \underline{\textbf{86.9}} & \underline{\textbf{95.9}} & \underline{\textbf{94.6}} \\
        \cmidrule{2-17}
        & \modelpunkt & - & - & - & - & - & - & - & - & - & 86.7 & - & 89.9 & 73.3 & 84.8 & 88.2 \\
        & \modelpysbd & - & 5.9 & 38.0 & - & - & 72.9 & - & - & - & - & - & 70.2 & 66.5 & 62.7 & 59.6 \\
        & \modelspacydp & - & - & - & - & - & - & - & 87.3 & - & - & - & 90.2 & 74.0 & 91.1 & 89.0 \\
        & \modelwtpu & 74.6 & 58.2 & 64.5 & 74.9 & 71.7 & 93.2 & 77.9 & 87.7 & - & 87.5 & 68.7 & 88.2 & 76.7 & 90.9 & 90.6 \\
        & \modelwtpt & 76.4 & 63.7 & 64.6 & 74.6 & 72.5 & 92.8 & 82.1 & 88.6 & - & 90.0 & 74.2 & 90.1 & 84.6 & 91.9 & 89.4 \\
        & \modelwtppunct & \underline{87.8} & 70.5 & 76.1 & 83.0 & \underline{89.1} & \underline{96.2} & 86.5 & 94.0 & - & \underline{94.9} & 81.0 & 94.5 & \underline{89.4} & \underline{95.9} & \underline{94.7} \\
        & \modelourslora & \underline{\textbf{88.5}} & \underline{\textbf{75.9}} & \underline{\textbf{79.1}} & \underline{\textbf{85.0}} & \underline{\textbf{89.4}} & \underline{\textbf{96.4}} & \underline{\textbf{88.6}} & \underline{\textbf{94.6}} & - & \underline{\textbf{95.0}} & \underline{\textbf{83.0}} & \underline{\textbf{95.2}} & \underline{\textbf{90.1}} & \underline{\textbf{96.1}} & \underline{\textbf{94.8}} \\
        \midrule
        \multirow{13}{*}{Ersatz} 
        & \modelspacydpm & - & - & 91.1 & - & - & - & - & - & - & 96.4 & - & - & 93.5 & - & 94.0 \\
        & \modelersatz & - & - & 92.8 & - & - & - & - & - & - & 96.7 & - & - & 95.4 & - & 97.5 \\
        & \modelllama & - & - & \underline{\textbf{92.4}} & - & - & - & - & - & - & 95.9 & - & - & \underline{\textbf{97.3}} & - & \underline{98.2} \\
        & \modelcommandr & - & - & 55.8 & - & - & - & - & - & - & 66.4 & - & - & 75.9 & - & 87.2 \\
        & \modelours & - & - & 89.7 & - & - & - & - & - & - & 94.3 & - & - & \underline{96.6} & - & 96.7 \\
        & \modeloursft & - & - & \underline{92.3} & - & - & - & - & - & - & \underline{\textbf{98.5}} & - & - & \underline{97.2} & - & \underline{\textbf{98.3}} \\
        \cmidrule{2-17}
        & \modelpunkt & - & - & - & - & - & - & - & - & - & 96.7 & - & - & 95.3 & - & 97.7 \\
        & \modelpysbd & - & - & 46.2 & - & - & - & - & - & - & - & - & - & 95.3 & - & 73.9 \\
        & \modelspacydp & - & - & - & - & - & - & - & - & - & - & - & - & 96.3 & - & \underline{98.5} \\
        & \modelwtpu & - & - & 87.0 & - & - & - & - & - & - & 93.6 & - & - & 95.3 & - & 96.5 \\
        & \modelwtpt & - & - & 88.7 & - & - & - & - & - & - & 93.9 & - & - & 95.6 & - & 96.7 \\
        & \modelwtppunct & - & - & \underline{92.7} & - & - & - & - & - & - & \underline{\textbf{99.0}} & - & - & \underline{\textbf{99.3}} & - & \underline{98.6} \\
        & \modelourslora & - & - & \underline{\textbf{93.1}} & - & - & - & - & - & - & 98.5 & - & - & \underline{\textbf{99.3}} & - & \underline{\textbf{98.7}} \\
        \bottomrule
        \end{tabular}
    } 
    \caption{
    Sentence segmentation test F1 scores on languages \texttt{af}-\texttt{en}. Results are shown using 3-layer variations of all models.
    Numerically best results are in \textbf{bold}, statistically indistinguishable ones from this best are \underline{underlined}.
   }
    \label{tab:all_comparisons_1}
\end{table*}  

\begin{table*}[t]
   \centering
   \small
   \scalebox{0.825}{
        \begin{tabular}{llcccccccccccccccc}
        \toprule
        & \textbf{Model} & \texttt{eo} & \texttt{es} & \texttt{et} & \texttt{eu} & \texttt{fa} & \texttt{fi} & \texttt{fr} & \texttt{fy} & \texttt{ga} & \texttt{gd} & \texttt{gl} & \texttt{gu} & \texttt{ha} & \texttt{he} & \texttt{hi} \\

        \midrule
        \multirow{13}{*}{UD} & 
        \modelspacydpm & - & 98.6 & 93.6 & 95.7 & 99.8 & 92.7 & 96.1 & - & 96.3 & 62.9 & 92.1 & - & - & 93.9 & - \\
        & \modelersatz & - & 97.5 & 92.9 & - & - & 92.8 & 97.3 & - & - & - & - & - & - & - & 99.5 \\
        & \modelllama & - & 98.5 & 94.1 & 97.3 & \underline{99.7} & 95.9 & \underline{\textbf{99.1}} & - & \underline{94.5} & 67.6 & 94.6 & - & - & 93.4 & \underline{99.8} \\
        & \modelcommandr & - & 81.8 & 74.9 & 79.4 & 95.5 & 75.3 & 92.3 & - & 73.5 & 54.2 & 73.7 & - & - & 77.2 & 91.8 \\
        & \modelours & - & 97.0 & 92.8 & 97.0 & 98.5 & 93.8 & 97.5 & - & 87.3 & 68.1 & 97.4 & - & - & 94.2 & 96.2 \\
        & \modeloursft & - & \underline{\textbf{99.4}} & \underline{\textbf{98.2}} & \underline{\textbf{100.0}} & \underline{\textbf{99.9}} & \underline{\textbf{97.2}} & \underline{98.3} & - & \underline{\textbf{95.5}} & \underline{\textbf{84.3}} & \underline{\textbf{98.9}} & - & - & \underline{95.5} & \underline{\textbf{99.9}} \\
        \cmidrule{2-17}
        & \modelpunkt & - & 98.5 & 93.6 & - & - & 92.6 & 97.0 & - & - & - & - & - & - & - & - \\
        & \modelpysbd & - & 46.2 & - & - & 98.8 & - & 62.1 & - & - & - & - & - & - & - & 99.8 \\
        & \modelspacydp & - & 99.1 & - & - & - & 94.9 & 91.6 & - & - & - & - & - & - & - & - \\
        & \modelwtpu & - & 96.5 & 92.6 & 97.1 & 96.6 & 92.1 & 96.4 & - & 84.3 & 71.2 & 97.5 & - & - & 95.1 & 96.1 \\
        & \modelwtpt & - & 96.9 & 92.5 & 97.3 & 97.8 & 92.7 & 96.6 & - & 90.4 & 70.7 & \underline{\textbf{98.7}} & - & - & 96.1 & 96.8 \\
        & \modelwtppunct & - & \underline{\textbf{99.7}} & \underline{\textbf{98.0}} & \underline{99.9} & \underline{\textbf{100.0}} & \underline{\textbf{98.1}} & \underline{98.3} & - & \underline{98.2} & 79.6 & \underline{98.6} & - & - & \underline{\textbf{97.1}} & \underline{\textbf{99.9}} \\
        & \modelourslora & - & \underline{99.6} & \underline{97.7} & \underline{\textbf{100.0}} & \underline{\textbf{100.0}} & \underline{97.7} & \underline{\textbf{98.9}} & - & \underline{\textbf{98.8}} & \underline{\textbf{82.2}} & \underline{98.7} & - & - & \underline{96.4} & \underline{99.9} \\
        \midrule
        \multirow{13}{*}{OPUS100} 
        & \modelspacydpm & 88.4 & 90.4 & 87.0 & 80.7 & 54.5 & 92.9 & 85.1 & 21.7 & 61.0 & 36.4 & 88.2 & 5.2 & 88.5 & 91.8 & 52.8 \\
        & \modelersatz & - & 90.0 & 87.4 & - & - & 92.7 & 86.1 & - & - & - & - & 21.2 & - & - & 58.0 \\
        & \modelllama & 91.6 & 93.6 & 89.0 & 84.5 & 60.4 & 93.9 & 91.4 & 42.4 & 71.3 & 62.8 & 90.8 & 33.0 & 88.1 & \underline{91.6} & 59.6 \\
        & \modelcommandr & 84.6 & 80.1 & 67.2 & 57.8 & 42.9 & 68.4 & 80.8 & 30.6 & 55.9 & 50.5 & 71.6 & 16.7 & 65.8 & 62.7 & 39.3 \\
        & \modelours & 90.3 & 91.1 & 84.6 & 82.2 & 56.8 & 91.3 & 87.3 & 64.9 & \underline{\textbf{82.3}} & \underline{81.6} & 87.9 & 71.5 & 83.8 & 89.4 & 62.9 \\
        & \modeloursft & \underline{\textbf{95.2}} & \underline{\textbf{95.3}} & \underline{\textbf{93.0}} & \underline{\textbf{90.0}} & \underline{\textbf{60.1}} & \underline{\textbf{95.3}} & \underline{\textbf{92.6}} & \underline{\textbf{91.0}} & \underline{80.8} & \underline{\textbf{82.5}} & \underline{\textbf{93.1}} & \underline{\textbf{83.4}} & \underline{\textbf{91.4}} & \underline{\textbf{92.2}} & \underline{\textbf{72.8}} \\
        \cmidrule{2-17}
        & \modelpunkt & - & 89.8 & 87.6 & - & - & 93.1 & 85.8 & - & - & - & - & - & - & - & - \\
        & \modelpysbd & - & 67.6 & - & - & 41.3 & - & 80.9 & - & - & - & - & - & - & - & 23.0 \\
        & \modelspacydp & - & 88.0 & - & - & - & 92.1 & 84.1 & - & - & - & - & - & - & - & - \\
        & \modelwtpu & 90.9 & 89.9 & 82.5 & 84.4 & 59.6 & 90.8 & \underline{\textbf{86.8}} & 44.4 & 77.8 & 83.5 & 88.5 & 69.2 & 82.8 & 90.2 & 65.1 \\
        & \modelwtpt & 90.5 & 91.4 & 87.7 & 86.0 & 59.3 & 92.5 & - & 61.0 & 77.3 & 83.7 & 88.9 & 69.6 & 88.9 & 89.4 & 64.3 \\
        & \modelwtppunct & \underline{95.4} & 95.0 & \underline{94.6} & 91.6 & 72.7 & 95.5 & - & \underline{88.1} & 87.5 & 92.7 & 93.9 & 76.5 & \underline{\textbf{91.5}} & \underline{93.8} & 76.3 \\
        & \modelourslora & \underline{\textbf{95.8}} & \underline{\textbf{95.8}} & \underline{\textbf{94.7}} & \underline{\textbf{92.8}} & \underline{\textbf{74.1}} & \underline{\textbf{96.3}} & - & \underline{\textbf{88.8}} & \underline{\textbf{90.6}} & \underline{\textbf{94.5}} & \underline{\textbf{94.6}} & \underline{\textbf{80.5}} & \underline{91.0} & \underline{\textbf{94.1}} & \underline{\textbf{81.5}} \\
        \midrule
        \multirow{13}{*}{Ersatz} 
        & \modelspacydpm & - & 97.2 & 97.0 & - & - & 95.0 & 96.4 & - & - & - & - & 3.8 & - & - & 17.9 \\
        & \modelersatz & - & 96.6 & 98.0 & - & - & 96.0 & 96.3 & - & - & - & - & 94.3 & - & - & 96.8 \\
        & \modelllama & - & 98.3 & 97.0 & - & - & 96.7 & 97.9 & - & - & - & - & \underline{\textbf{93.1}} & - & - & 97.3 \\
        & \modelcommandr & - & 76.8 & 77.2 & - & - & 78.8 & 84.6 & - & - & - & - & 74.8 & - & - & 84.9 \\
        & \modelours & - & 98.4 & 95.9 & - & - & 97.6 & 97.3 & - & - & - & - & \underline{92.0} & - & - & 95.5 \\
        & \modeloursft & - & \underline{\textbf{99.5}} & \underline{\textbf{98.9}} & - & - & \underline{\textbf{98.4}} & \underline{\textbf{98.5}} & - & - & - & - & 70.7 & - & - & \underline{\textbf{98.1}} \\
        \cmidrule{2-17}
        & \modelpunkt & - & 96.4 & 97.5 & - & - & 95.9 & 96.0 & - & - & - & - & - & - & - & - \\
        & \modelpysbd & - & 84.2 & - & - & - & - & 96.0 & - & - & - & - & - & - & - & 87.5 \\
        & \modelspacydp & - & - & - & - & - & - & 91.0 & - & - & - & - & - & - & - & - \\
        & \modelwtpu & - & \underline{\textbf{98.7}} & 96.0 & - & - & 97.4 & 97.2 & - & - & - & - & 89.7 & - & - & 93.9 \\
        & \modelwtpt & - & - & 95.7 & - & - & 97.2 & 96.6 & - & - & - & - & 88.9 & - & - & 94.6 \\
        & \modelwtppunct & - & - & 98.0 & - & - & \underline{\textbf{99.4}} & 98.4 & - & - & - & - & \underline{\textbf{96.7}} & - & - & 96.3 \\
        & \modelourslora & - & - & \underline{\textbf{99.0}} & - & - & 98.6 & \underline{\textbf{99.1}} & - & - & - & - & \underline{95.4} & - & - & \underline{\textbf{97.3}} \\
        \bottomrule
        \end{tabular}
    } 
    \caption{
    Sentence segmentation test F1 scores on languages \texttt{eo}-\texttt{hi}.
   }
    \label{tab:all_comparisons_2}
\end{table*}

\begin{table*}[t]
   \centering
   \small
   \scalebox{0.825}{
        \begin{tabular}{llcccccccccccccccc}
        \toprule
        & \textbf{Model} & \texttt{hu} & \texttt{hy} & \texttt{id} & \texttt{ig} & \texttt{is} & \texttt{it} & \texttt{ja} & \texttt{jv} & \texttt{ka} & \texttt{kk} & \texttt{km} & \texttt{kn} & \texttt{ko} & \texttt{ku} & \texttt{ky} \\
        \midrule
        \multirow{13}{*}{UD} & 
        \modelspacydpm & 98.3 & 11.3 & 97.8 & - & 95.1 & 92.3 & 0.0 & 97.9 & - & 96.1 & - & - & 99.8 & - & - \\
        & \modelersatz & - & - & - & - & - & - & 93.4 & - & - & 95.6 & - & - & - & - & - \\
        & \modelllama & 98.1 & 92.0 & \underline{99.3} & - & 94.1 & \underline{98.1} & \underline{\textbf{97.9}} & \underline{99.1} & - & 98.5 & - & - & \underline{99.9} & - & - \\
        & \modelcommandr & 73.2 & 50.3 & 91.7 & - & 74.6 & 88.8 & 90.7 & 83.3 & - & 86.1 & - & - & 79.9 & - & - \\
        & \modelours & 97.0 & \underline{97.6} & 98.5 & - & 73.4 & 96.3 & 96.1 & \underline{98.8} & - & 95.8 & - & - & 99.6 & - & - \\
        & \modeloursft & \underline{\textbf{99.5}} & \underline{\textbf{98.4}} & \underline{\textbf{99.6}} & - & \underline{\textbf{95.7}} & \underline{\textbf{98.7}} & \underline{97.1} & \underline{\textbf{99.3}} & - & \underline{\textbf{99.3}} & - & - & \underline{\textbf{100.0}} & - & - \\
        \cmidrule{2-17}
        & \modelpunkt & - & - & - & - & - & 95.2 & - & - & - & - & - & - & - & - & - \\
        & \modelpysbd & - & 92.8 & - & - & - & 74.9 & \underline{97.9} & - & - & 95.5 & - & - & - & - & - \\
        & \modelspacydp & - & - & - & - & - & \underline{99.5} & \underline{97.8} & - & - & - & - & - & 99.9 & - & - \\
        & \modelwtpu & 96.0 & 96.0 & \underline{\textbf{98.0}} & - & 85.8 & 93.7 & 93.4 & \underline{\textbf{97.8}} & - & \underline{\textbf{97.4}} & - & - & 99.2 & - & - \\
        & \modelwtpt & 96.3 & 96.2 & - & - & 88.9 & 93.7 & 95.7 & - & - & 82.8 & - & - & 99.4 & - & - \\
        & \modelwtppunct & \underline{\textbf{99.5}} & \underline{\textbf{98.5}} & - & - & \underline{\textbf{96.6}} & \underline{99.3} & \underline{\textbf{98.1}} & - & - & 84.7 & - & - & \underline{\textbf{99.9}} & - & - \\
        & \modelourslora & \underline{\textbf{99.5}} & \underline{98.3} & - & - & \underline{\textbf{96.8}} & \underline{\textbf{99.7}} & \underline{98.0} & - & - & \underline{96.9} & - & - & \underline{\textbf{99.9}} & - & - \\
        \midrule
        \multirow{13}{*}{OPUS100} 
        & \modelspacydpm & 93.0 & 24.2 & 89.6 & 29.8 & 95.0 & 85.8 & 0.1 & - & 38.3 & 42.1 & 0.0 & 9.8 & 50.9 & 26.8 & 21.7 \\
        & \modelersatz & - & - & - & - & - & - & 28.8 & - & - & 37.9 & 0.0 & - & - & - & - \\
        & \modelllama & 94.2 & 73.1 & 92.3 & 39.3 & 95.1 & \underline{90.7} & 9.0 & - & 89.2 & 56.4 & 9.9 & 26.7 & 70.3 & 41.9 & 58.3 \\
        & \modelcommandr & 62.6 & 46.7 & 71.3 & 28.0 & 77.7 & 78.6 & 6.0 & - & 25.6 & 34.5 & 4.3 & 5.5 & 38.9 & 29.8 & 26.5 \\
        & \modelours & 92.3 & 85.3 & 86.6 & 81.7 & 93.7 & 87.0 & \underline{\textbf{83.1}} & - & 75.7 & 80.3 & 70.4 & 70.2 & 72.4 & 77.2 & 84.1 \\
        & \modeloursft & \underline{\textbf{95.8}} & \underline{\textbf{90.6}} & \underline{\textbf{93.7}} & \underline{\textbf{92.1}} & \underline{\textbf{96.5}} & \underline{\textbf{90.9}} & 78.0 & - & \underline{\textbf{93.6}} & \underline{\textbf{92.2}} & \underline{\textbf{86.6}} & \underline{\textbf{87.9}} & \underline{\textbf{78.9}} & \underline{\textbf{91.1}} & \underline{\textbf{91.8}} \\
        \cmidrule{2-17}
        & \modelpunkt & - & - & - & - & - & 87.5 & - & - & - & - & - & - & - & - & - \\
        & \modelpysbd & - & 58.4 & - & - & - & 70.2 & 43.4 & - & - & 35.7 & - & - & - & - & - \\
        & \modelspacydp & - & - & - & - & - & 85.3 & 42.6 & - & - & - & - & - & 46.6 & - & - \\
        & \modelwtpu & 91.6 & \underline{\textbf{85.1}} & 88.9 & 78.6 & 93.9 & 84.6 & 44.6 & - & 91.3 & 73.4 & 71.8 & 64.5 & 56.7 & 78.1 & 84.5 \\
        & \modelwtpt & 92.1 & - & 89.5 & 82.1 & 94.4 & 88.4 & 79.5 & - & 91.1 & 74.9 & 71.1 & 60.4 & 70.4 & 66.4 & 84.4 \\
        & \modelwtppunct & \underline{96.2} & - & \underline{93.9} & 90.2 & \underline{96.6} & 93.4 & 86.7 & - & 92.8 & \underline{92.1} & 79.0 & 78.0 & \underline{81.6} & 85.1 & 90.3 \\
        & \modelourslora & \underline{\textbf{96.3}} & - & \underline{\textbf{94.1}} & \underline{\textbf{92.3}} & \underline{\textbf{96.6}} & \underline{\textbf{94.3}} & \underline{\textbf{90.5}} & - & \underline{\textbf{93.3}} & \underline{\textbf{92.1}} & \underline{\textbf{87.8}} & \underline{\textbf{83.9}} & \underline{\textbf{81.8}} & \underline{\textbf{90.9}} & \underline{\textbf{92.0}} \\
        \midrule
        \multirow{13}{*}{Ersatz} 
        & \modelspacydpm & - & - & - & - & - & - & 0.0 & - & - & 96.9 & 0.0 & - & - & - & - \\
        & \modelersatz & - & - & - & - & - & - & 85.7 & - & - & \underline{99.6} & 31.3 & - & - & - & - \\
        & \modelllama & - & - & - & - & - & - & 87.1 & - & - & 99.0 & 85.6 & - & - & - & - \\
        & \modelcommandr & - & - & - & - & - & - & 59.7 & - & - & 86.7 & 6.7 & - & - & - & - \\
        & \modelours & - & - & - & - & - & - & 86.5 & - & - & 97.1 & \underline{\textbf{89.4}} & - & - & - & - \\
        & \modeloursft & - & - & - & - & - & - & \underline{\textbf{89.1}} & - & - & \underline{\textbf{99.7}} & 83.9 & - & - & - & - \\
        \cmidrule{2-17}
        & \modelpunkt & - & - & - & - & - & - & - & - & - & - & - & - & - & - & - \\
        & \modelpysbd & - & - & - & - & - & - & 87.0 & - & - & 64.7 & - & - & - & - & - \\
        & \modelspacydp & - & - & - & - & - & - & 91.0 & - & - & - & - & - & - & - & - \\
        & \modelwtpu & - & - & - & - & - & - & 80.2 & - & - & 96.3 & 70.2 & - & - & - & - \\
        & \modelwtpt & - & - & - & - & - & - & 81.5 & - & - & 95.7 & 91.4 & - & - & - & - \\
        & \modelwtppunct & - & - & - & - & - & - & \underline{96.7} & - & - & \underline{99.7} & \underline{92.0} & - & - & - & - \\
        & \modelourslora & - & - & - & - & - & - & \underline{\textbf{94.6}} & - & - & \underline{\textbf{99.8}} & \underline{\textbf{92.0}} & - & - & - & - \\
        \bottomrule
        \end{tabular}
    } 
    \caption{
    Sentence segmentation test F1 scores on languages \texttt{hu}-\texttt{ky}.
   }
    \label{tab:all_comparisons_3}
\end{table*}

\begin{table*}[t]
   \centering
   \small
   \scalebox{0.825}{
        \begin{tabular}{llccccccccccccccc}
        \toprule
        & \textbf{Model} & \texttt{la} & \texttt{lt} & \texttt{lv} & \texttt{mg} & \texttt{mk} & \texttt{ml} & \texttt{mn} & \texttt{mr} & \texttt{ms} & \texttt{mt} & \texttt{my} & \texttt{ne} & \texttt{nl} & \texttt{no} & \texttt{pa} \\

        \midrule
        \multirow{13}{*}{UD} 
        & \modelspacydpm & 0.0 & 93.8 & 98.5 & - & - & - & - & 92.5 & - & 85.7 & - & - & 92.8 & 96.9 & - \\
        & \modelersatz & - & 92.2 & 96.9 & - & - & - & - & - & - & - & - & - & - & - & - \\
        & \modelllama & 90.0 & 94.3 & 97.0 & - & - & - & - & 90.0 & - & \underline{\textbf{94.4}} & - & - & \underline{\textbf{96.0}} & 96.5 & - \\
        & \modelcommandr & 65.1 & 65.7 & 69.5 & - & - & - & - & 71.6 & - & 70.1 & - & - & 66.3 & 69.5 & - \\
        & \modelours & 67.8 & \underline{97.1} & 96.4 & - & - & - & - & 83.5 & - & 86.7 & - & - & 92.6 & 98.6 & - \\
        & \modeloursft & \underline{\textbf{96.6}} & \underline{\textbf{97.9}} & \underline{\textbf{99.2}} & - & - & - & - & \underline{\textbf{100.0}} & - & \underline{93.0} & - & - & \underline{95.7} & \underline{\textbf{99.0}} & - \\
        \cmidrule{2-17}
        & \modelpunkt & - & - & - & - & - & - & - & - & - & - & - & - & 95.7 & 95.6 & - \\
        & \modelpysbd & - & - & - & - & - & - & - & 60.3 & - & - & - & - & 93.6 & - & - \\
        & \modelspacydp & - & 92.0 & - & - & - & - & - & - & - & - & - & - & 93.1 & - & - \\
        & \modelwtpu & 89.2 & 98.2 & 96.5 & - & - & - & - & 89.4 & - & 89.8 & - & - & 94.1 & 98.2 & - \\
        & \modelwtpt & 89.3 & 97.9 & 96.4 & - & - & - & - & \underline{92.0} & - & 87.3 & - & - & 92.9 & 98.4 & - \\
        & \modelwtppunct & \underline{97.3} & \underline{\textbf{99.6}} & \underline{\textbf{99.0}} & - & - & - & - & \underline{\textbf{98.8}} & - & \underline{\textbf{93.6}} & - & - & \underline{\textbf{97.0}} & \underline{\textbf{99.4}} & - \\
        & \modelourslora & \underline{\textbf{97.5}} & 98.2 & \underline{98.9} & - & - & - & - & \underline{96.5} & - & 90.9 & - & - & \underline{96.0} & \underline{99.1} & - \\
        \midrule
        \multirow{13}{*}{OPUS100} & 
        \modelspacydpm & - & 76.5 & 76.9 & 83.1 & 93.2 & 38.7 & 32.8 & 86.5 & 87.6 & 55.6 & 0.0 & 6.4 & 93.0 & 95.0 & 4.9 \\
        & \modelersatz & - & 77.0 & 77.6 & - & - & - & - & - & - & - & - & - & - & - & - \\
        & \modelllama & - & 82.8 & 83.1 & 85.7 & 94.1 & 80.9 & 49.3 & 88.1 & 91.0 & 78.8 & 16.3 & 39.7 & 93.7 & \underline{95.3} & 26.7 \\
        & \modelcommandr & - & 70.3 & 70.4 & 64.0 & 48.2 & 0.7 & 26.6 & 65.2 & 67.9 & 65.1 & 17.1 & 23.3 & 74.8 & 71.5 & 14.4 \\
        & \modelours & - & 82.6 & 82.6 & 88.5 & 93.0 & 77.6 & 73.5 & 89.9 & 85.8 & 62.2 & 74.1 & 69.8 & 92.1 & 94.8 & 69.5 \\
        & \modeloursft & - & \underline{\textbf{90.0}} & \underline{\textbf{89.7}} & \underline{\textbf{91.6}} & \underline{\textbf{95.4}} & \underline{\textbf{85.9}} & \underline{\textbf{90.1}} & \underline{\textbf{93.3}} & \underline{\textbf{94.1}} & \underline{\textbf{85.1}} & \underline{\textbf{89.4}} & \underline{\textbf{82.0}} & \underline{\textbf{94.8}} & \underline{\textbf{95.8}} & \underline{\textbf{81.0}} \\
        \cmidrule{2-17}
        & \modelpunkt & - & - & - & - & - & 80.2 & - & - & - & - & - & - & \underline{\textbf{93.4}} & 94.5 & - \\
        & \modelpysbd & - & - & - & - & - & - & - & 86.2 & - & - & 27.4 & - & 18.2 & - & - \\
        & \modelspacydp & - & 78.9 & - & - & 82.2 & - & - & - & - & - & - & - & 92.4 & - & - \\
        & \modelwtpu & - & 76.5 & 78.0 & 89.1 & 92.3 & 80.0 & \underline{\textbf{80.7}} & 88.5 & 87.0 & 60.6 & 68.9 & 68.9 & 91.5 & 94.2 & 55.6 \\
        & \modelwtpt & - & 84.1 & 85.6 & 91.5 & 92.4 & 81.7 & - & 88.6 & 87.9 & 80.4 & 74.3 & 68.7 & - & 94.3 & 62.2 \\
        & \modelwtppunct & - & 90.1 & 91.5 & \underline{95.3} & \underline{\textbf{95.6}} & \underline{\textbf{86.5}} & - & 93.5 & \underline{\textbf{94.0}} & \underline{88.4} & 82.2 & 74.3 & - & \underline{96.1} & 77.3 \\
        & \modelourslora & - & \underline{\textbf{92.6}} & \underline{\textbf{92.8}} & \underline{\textbf{95.4}} & \underline{95.5} & \underline{86.5} & - & \underline{\textbf{94.9}} & \underline{93.8} & \underline{\textbf{89.2}} & \underline{\textbf{86.7}} & \underline{\textbf{77.3}} & - & \underline{\textbf{96.3}} & \underline{\textbf{79.6}} \\
        \midrule
        \multirow{13}{*}{Ersatz} 
        & \modelspacydpm & - & 93.3 & 98.6 & - & - & - & - & - & - & - & - & - & - & - & - \\
        & \modelersatz & - & 95.0 & 98.7 & - & - & - & - & - & - & - & - & - & - & - & - \\
        & \modelllama & - & 94.9 & 98.8 & - & - & - & - & - & - & - & - & - & - & - & - \\
        & \modelcommandr & - & 78.3 & 82.7 & - & - & - & - & - & - & - & - & - & - & - & - \\
        & \modelours & - & 96.3 & 97.5 & - & - & - & - & - & - & - & - & - & - & - & - \\
        & \modeloursft & - & \underline{\textbf{98.3}} & \underline{\textbf{99.3}} & - & - & - & - & - & - & - & - & - & - & - & - \\
        \cmidrule{2-17}
        & \modelpunkt & - & - & - & - & - & - & - & - & - & - & - & - & - & - & - \\
        & \modelpysbd & - & - & - & - & - & - & - & - & - & - & - & - & - & - & - \\
        & \modelspacydp & - & 74.9 & - & - & - & - & - & - & - & - & - & - & - & - & - \\
        & \modelwtpu & - & 96.5 & 96.9 & - & - & - & - & - & - & - & - & - & - & - & - \\
        & \modelwtpt & - & 96.4 & 97.2 & - & - & - & - & - & - & - & - & - & - & - & - \\
        & \modelwtppunct & - & \underline{\textbf{99.2}} & \underline{99.3} & - & - & - & - & - & - & - & - & - & - & - & - \\
        & \modelourslora & - & 98.4 & \underline{\textbf{99.4}} & - & - & - & - & - & - & - & - & - & - & - & - \\
        \bottomrule
        \end{tabular}
    } 
    \caption{
    Sentence segmentation test F1 scores on languages \texttt{la}-\texttt{pa}.
   }
    \label{tab:all_comparisons_4}
\end{table*}

\begin{table*}[t]
   \centering
   \small
   \scalebox{0.825}{
        \begin{tabular}{llccccccccccccccc}
        \toprule
        & \textbf{Model} & \texttt{pl} & \texttt{ps} & \texttt{pt} & \texttt{ro} & \texttt{ru} & \texttt{si} & \texttt{sk} & \texttt{sl} & \texttt{sq} & \texttt{sr} & \texttt{sv} & \texttt{ta} & \texttt{te} & \texttt{tg} & \texttt{th} \\

        \midrule
        \multirow{13}{*}{UD} & \modelspacydpm \\
        & \modelersatz & 97.4 & - & - & 98.3 & 77.7 & - & - & - & - & - & - & 90.5 & - & - & - \\
        & \modelllama & \underline{\textbf{99.3}} & - & 93.5 & 93.5 & 88.5 & - & 90.6 & 98.6 & \underline{\textbf{100.0}} & 96.5 & 95.5 & 96.7 & - & - & \underline{\textbf{81.4}} \\
        & \modelcommandr & 91.4 & - & 72.2 & 59.6 & 61.2 & - & 73.5 & 75.4 & 92.0 & 82.0 & 72.8 & 0.0 & - & - & 12.4 \\
        & \modelours & 95.3 & - & \underline{96.6} & 73.2 & 82.7 & - & 94.8 & 96.8 & \underline{\textbf{100.0}} & 98.0 & 95.0 & \underline{\textbf{99.5}} & - & - & 72.7 \\
        & \modeloursft & \underline{99.2} & - & \underline{\textbf{97.2}} & \underline{\textbf{99.0}} & \underline{\textbf{92.4}} & - & \underline{\textbf{96.5}} & \underline{\textbf{99.2}} & \underline{99.1} & \underline{\textbf{99.4}} & \underline{\textbf{96.2}} & \underline{\textbf{99.5}} & - & - & 71.2 \\
        \cmidrule{2-17}
        & \modelpunkt & 97.2 & - & 91.9 & - & 78.3 & - & - & - & - & - & 93.9 & - & - & - & - \\
        & \modelpysbd & 84.8 & - & - & - & 67.9 & - & 86.2 & - & - & - & - & - & - & - & - \\
        & \modelspacydp & 98.5 & - & 98.1 & 95.3 & 80.3 & - & - & - & - & - & 87.0 & - & - & - & - \\
        & \modelwtpu & 94.4 & - & 95.9 & 80.9 & 84.9 & - & 96.0 & 95.7 & \underline{\textbf{100.0}} & 97.9 & 94.7 & 96.9 & - & - & \underline{\textbf{67.3}} \\
        & \modelwtpt & 95.5 & - & 95.6 & 93.5 & 86.8 & - & 95.8 & 96.2 & - & 98.2 & 95.0 & \underline{97.7} & - & - & - \\
        & \modelwtppunct & \underline{\textbf{99.3}} & - & \underline{\textbf{98.4}} & \underline{99.4} & \underline{\textbf{93.1}} & - & \underline{\textbf{98.1}} & \underline{99.1} & - & \underline{\textbf{99.8}} & \underline{96.5} & \underline{\textbf{100.0}} & - & - & - \\
        & \modelourslora & 98.9 & - & 97.8 & \underline{\textbf{99.5}} & \underline{92.7} & - & 96.9 & \underline{\textbf{99.2}} & - & \underline{99.7} & \underline{\textbf{96.5}} & \underline{\textbf{99.5}} & - & - & - \\
        \midrule
        \multirow{13}{*}{OPUS100}
        & \modelspacydpm & 92.0 & 2.4 & 91.6 & 91.3 & 75.2 & 75.4 & 91.6 & 92.6 & 92.6 & 94.6 & 93.1 & 36.6 & 64.1 & 69.5 & 21.7 \\
        & \modelersatz & 92.1 & 1.8 & - & 92.8 & 68.6 & - & - & - & - & - & - & 45.3 & - & - & - \\
        & \modelllama & 93.8 & 26.1 & \underline{\textbf{94.1}} & 94.9 & \underline{\textbf{89.3}} & 76.9 & 94.3 & 93.8 & 92.3 & 94.6 & 94.3 & 45.3 & 65.9 & 76.1 & 66.0 \\
        & \modelcommandr & 74.4 & 7.2 & 80.8 & 70.4 & \underline{88.1} & 8.6 & 74.8 & 63.8 & 59.3 & 67.2 & 76.9 & 1.9 & 36.1 & 54.9 & 10.6 \\
        & \modelours & 92.4 & 68.6 & 91.7 & 91.1 & 82.8 & 79.2 & 91.4 & 91.6 & 89.7 & 94.0 & 91.3 & 60.5 & 76.5 & 79.6 & 68.0 \\
        & \modeloursft & \underline{\textbf{95.6}} & \underline{\textbf{85.6}} & \underline{93.8} & \underline{\textbf{95.9}} & 84.1 & \underline{\textbf{85.4}} & \underline{\textbf{95.9}} & \underline{\textbf{95.3}} & \underline{\textbf{95.7}} & \underline{\textbf{95.9}} & \underline{\textbf{95.0}} & \underline{\textbf{75.0}} & \underline{\textbf{86.8}} & \underline{\textbf{92.3}} & \underline{\textbf{72.9}} \\
        \cmidrule{2-17}
        & \modelpunkt & 92.5 & - & 92.2 & - & 75.8 & - & - & - & - & - & 92.5 & - & - & - & - \\
        & \modelpysbd & 17.5 & - & - & - & 64.9 & - & 29.5 & - & - & - & - & - & - & - & - \\
        & \modelspacydp & 92.9 & - & 90.8 & 91.9 & 75.4 & - & - & - & - & - & 90.2 & - & - & - & - \\
        & \modelwtpu & 91.8 & 63.3 & 89.9 & 88.5 & \underline{\textbf{80.6}} & 79.5 & 89.7 & 91.3 & 88.7 & 93.8 & 90.4 & 64.3 & 77.4 & 80.1 & 66.6 \\
        & \modelwtpt & 92.2 & 70.4 & 91.4 & 89.0 & - & 80.1 & 92.4 & 92.7 & 89.9 & 94.3 & 92.5 & 65.7 & 77.5 & 82.7 & 69.7 \\
        & \modelwtppunct & \underline{95.6} & \underline{76.1} & \underline{95.3} & \underline{96.7} & - & \underline{\textbf{85.4}} & 95.9 & \underline{95.0} & \underline{\textbf{95.5}} & \underline{\textbf{96.4}} & \underline{95.8} & 75.4 & 83.6 & 91.0 & 71.3 \\
        & \modelourslora & \underline{\textbf{96.0}} & \underline{\textbf{77.6}} & \underline{\textbf{95.4}} & \underline{\textbf{97.3}} & - & \underline{\textbf{85.9}} & \underline{\textbf{96.5}} & \underline{\textbf{95.3}} & \underline{95.5} & \underline{96.1} & \underline{\textbf{95.9}} & \underline{\textbf{80.8}} & \underline{\textbf{85.6}} & \underline{\textbf{92.1}} & \underline{\textbf{73.7}} \\
        \midrule
        \multirow{13}{*}{Ersatz} 
        & \modelspacydpm & 93.3 & 94.3 & - & 94.8 & 93.2 & - & - & - & - & - & - & 67.9 & - & - & - \\
        & \modelersatz & 94.9 & 93.7 & - & 96.0 & 94.2 & - & - & - & - & - & - & 95.2 & - & - & - \\
        & \modelllama & 95.4 & \underline{\textbf{96.3}} & - & 97.7 & 97.0 & - & - & - & - & - & - & 97.0 & - & - & - \\
        & \modelcommandr & 70.1 & 70.7 & - & 71.3 & 80.2 & - & - & - & - & - & - & 1.3 & - & - & - \\
        & \modelours & 93.5 & 92.4 & - & 97.5 & 97.2 & - & - & - & - & - & - & 97.6 & - & - & - \\
        & \modeloursft & \underline{\textbf{98.3}} & 82.7 & - & \underline{\textbf{99.1}} & \underline{\textbf{98.8}} & - & - & - & - & - & - & \underline{\textbf{98.5}} & - & - & - \\
        \cmidrule{2-17}
        & \modelpunkt & 94.0 & - & - & - & 93.7 & - & - & - & - & - & - & - & - & - & - \\
        & \modelpysbd & 45.7 & - & - & - & 55.2 & - & - & - & - & - & - & - & - & - & - \\
        & \modelspacydp & 94.5 & - & - & 94.4 & 94.1 & - & - & - & - & - & - & - & - & - & - \\
        & \modelwtpu & 94.6 & 83.7 & - & 97.5 & 97.5 & - & - & - & - & - & - & 94.1 & - & - & - \\
        & \modelwtpt & 92.8 & 91.0 & - & 96.9 & 97.6 & - & - & - & - & - & - & 94.7 & - & - & - \\
        & \modelwtppunct & \underline{97.7} & \underline{95.9} & - & \underline{\textbf{99.4}} & \underline{\textbf{99.4}} & - & - & - & - & - & - & \underline{97.8} & - & - & - \\
        & \modelourslora & \underline{\textbf{98.1}} & \underline{\textbf{96.1}} & - & \underline{99.3} & 98.8 & - & - & - & - & - & - & \underline{\textbf{98.1}} & - & - & - \\
        \bottomrule
        \end{tabular}
    } 
    \caption{
    Sentence segmentation test F1 scores on languages \texttt{pl}-\texttt{th}.
   }
    \label{tab:all_comparisons_5}
\end{table*}

\begin{table*}[t]
   \centering
   \small
   \scalebox{0.825}{
        \begin{tabular}{llcccccccccc}
        \toprule
        & \textbf{Model} & \texttt{tr} & \texttt{uk} & \texttt{ur} & \texttt{uz} & \texttt{vi} & \texttt{xh} & \texttt{yi} & \texttt{yo} & \texttt{zh} & \texttt{zu} \\

        \midrule
        \multirow{13}{*}{UD} 
        & \modelspacydpm & 97.5 & 93.9 & 0.0 & - & 96.0 & - & - & 79.2 & 0.0 & - \\
        & \modelersatz & 96.8 & - & - & - & - & - & - & - & 89.3 & - \\
        & \modelllama & 97.5 & 94.9 & 97.0 & - & 98.5 & - & - & \underline{\textbf{89.8}} & 95.8 & - \\
        & \modelcommandr & 61.8 & 62.1 & 82.3 & - & 92.4 & - & - & 66.1 & 91.8 & - \\
        & \modelours & 96.3 & 92.7 & 97.7 & - & 90.8 & - & - & 77.0 & 94.5 & - \\
        & \modeloursft & \underline{\textbf{98.6}} & \underline{\textbf{98.3}} & \underline{\textbf{99.3}} & - & \underline{\textbf{99.5}} & - & - & \underline{89.6} & \underline{\textbf{98.6}} & - \\
        \cmidrule{2-12}
        & \modelpunkt & 93.2 & - & - & - & - & - & - & - & - & - \\
        & \modelpysbd & - & - & 99.2 & - & - & - & - & - & 98.9 & - \\
        & \modelspacydp & - & 96.5 & - & - & - & - & - & - & 99.0 & - \\
        & \modelwtpu & 95.9 & 92.0 & 91.7 & - & 88.5 & - & - & 83.5 & 97.9 & - \\
        & \modelwtpt & 95.6 & 92.1 & 95.8 & - & 93.7 & - & - & - & 98.0 & - \\
        & \modelwtppunct & \underline{98.4} & \underline{\textbf{98.6}} & \underline{99.5} & - & \underline{\textbf{99.7}} & - & - & - & \underline{\textbf{99.9}} & - \\
        & \modelourslora & \underline{\textbf{98.5}} & \underline{98.1} & \underline{\textbf{99.5}} & - & \underline{99.4} & - & - & - & \underline{99.3} & - \\
        \midrule
        \multirow{13}{*}{OPUS100} 
        & \modelspacydpm & 93.6 & 89.2 & 29.4 & 63.6 & 90.1 & 64.6 & 4.1 & 27.2 & 0.0 & 25.4 \\
        & \modelersatz & 92.7 & - & - & - & - & - & - & - & 54.7 & - \\
        & \modelllama & 93.8 & 91.1 & 47.6 & 67.4 & 92.9 & 71.7 & 13.6 & 37.5 & 55.7 & 42.5 \\
        & \modelcommandr & 72.4 & 79.8 & 26.6 & 44.8 & 77.2 & 56.1 & 8.3 & 26.6 & 58.7 & 34.9 \\
        & \modelours & 93.2 & 88.3 & 51.3 & 76.3 & 91.1 & 80.3 & 61.1 & \underline{\textbf{67.3}} & 55.6 & 82.1 \\
        & \modeloursft & \underline{\textbf{94.7}} & \underline{\textbf{93.5}} & \underline{\textbf{60.5}} & \underline{\textbf{84.9}} & \underline{\textbf{94.7}} & \underline{\textbf{89.6}} & \underline{\textbf{89.0}} & 52.3 & \underline{\textbf{77.5}} & \underline{\textbf{93.3}} \\
        \cmidrule{2-12}
        & \modelpunkt & 93.4 & - & - & - & - & - & - & - & - & - \\
        & \modelpysbd & - & - & 31.4 & - & - & - & - & - & 69.0 & - \\
        & \modelspacydp & - & 89.8 & - & - & - & - & - & - & 68.2 & - \\
        & \modelwtpu & 92.8 & 88.2 & 53.0 & 76.4 & 90.1 & 77.2 & 73.0 & \underline{\textbf{75.4}} & 80.5 & 72.7 \\
        & \modelwtpt & 93.1 & 89.0 & 50.7 & 78.9 & 90.3 & 80.7 & 73.9 & - & 76.6 & 83.1 \\
        & \modelwtppunct & \underline{95.3} & \underline{94.3} & 66.3 & 85.0 & \underline{94.5} & 89.8 & 80.7 & - & 88.8 & 90.6 \\
        & \modelourslora & \underline{\textbf{95.4}} & \underline{\textbf{94.7}} & \underline{\textbf{72.0}} & \underline{\textbf{87.6}} & \underline{\textbf{94.8}} & \underline{\textbf{90.8}} & \underline{\textbf{86.6}} & - & \underline{\textbf{90.5}} & \underline{\textbf{92.0}} \\
        \midrule
        \multirow{13}{*}{Ersatz} &
        \modelspacydpm & 95.2 & - & - & - & - & - & - & - & 0.0 & - \\
        & \modelersatz & 96.2 & - & - & - & - & - & - & - & 87.4 & - \\
        & \modelllama & 94.3 & - & - & - & - & - & - & - & \underline{\textbf{94.5}} & - \\
        & \modelcommandr & 71.8 & - & - & - & - & - & - & - & 70.5 & - \\
        & \modelours & 93.5 & - & - & - & - & - & - & - & 84.1 & - \\
        & \modeloursft & \underline{\textbf{97.5}} & - & - & - & - & - & - & - & 90.5 & - \\
        \cmidrule{2-12}
        & \modelpunkt & 92.6 & - & - & - & - & - & - & - & - & - \\
        & \modelpysbd & - & - & - & - & - & - & - & - & 92.7 & - \\
        & \modelspacydp & - & - & - & - & - & - & - & - & 95.9 & - \\
        & \modelwtpu & 92.8 & - & - & - & - & - & - & - & 93.7 & - \\
        & \modelwtpt & 93.0 & - & - & - & - & - & - & - & 93.4 & - \\
        & \modelwtppunct & \underline{\textbf{98.3}} & - & - & - & - & - & - & - & \underline{\textbf{97.9}} & - \\
        & \modelourslora & \underline{98.2} & - & - & - & - & - & - & - & 95.0 & - \\
        \bottomrule
        \end{tabular}
    } 
    \caption{
        Sentence segmentation test F1 scores on languages \texttt{tr}-\texttt{zu}.
   }
    \label{tab:all_comparisons_6}
\end{table*}

\end{document}